\documentclass[10pt,twocolumn,letterpaper]{article}

\usepackage{wacv}              % To produce the CAMERA-READY version
\usepackage[margin=0.9in,top=0.95in,bottom=0.80in]{geometry}
\usepackage{graphicx}
\usepackage{amsmath}
\usepackage{amssymb}
\usepackage{appendix}
\usepackage{booktabs}
\usepackage{makecell}
\usepackage{multirow} 

\PassOptionsToPackage{table,xcdraw}{xcolor}  
\usepackage{colortbl}      % those two lines to fix the problem adding supplementary                  
\usepackage{pgfplots}
\pgfplotsset{compat=1.18}
\usepgfplotslibrary{groupplots}
\usepackage[none]{hyphenat}
\usepackage{tikz}
\usepackage{pgfplots}
\pgfplotsset{compat=1.18} % Set this to the version you have installed, 1.18 is just an example
\usepackage{pifont}
\usepackage{dsfont}
\usepackage{algorithm}
\usepackage{algpseudocode}

\usepackage[dvipsnames]{xcolor}

\definecolor{tableyellow}{rgb}{1, 1, 0.7}
\definecolor{tableorange}{rgb}{1, 0.85, 0.7}
\definecolor{tablered}{rgb}{1, 0.7, 0.7}

\definecolor{tabfirst}{rgb}{1, 0.7, 0.7}
\definecolor{tabsecond}{rgb}{1, 0.85, 0.7}
\definecolor{tabthird}{rgb}{1, 1, 0.7}

\definecolor{wacvblue}{rgb}{0.21,0.49,0.74}
\usepackage[pagebackref,breaklinks,colorlinks,allcolors=wacvblue]{hyperref}

\def\wacvPaperID{0341} % *** Enter the WACV Paper ID here
\def\confName{WACV}
\def\confYear{2026}

\title{Inpaint360GS: Efficient Object-Aware 3D Inpainting via Gaussian Splatting for 360° Scenes}

\author{
Shaoxiang Wang$^{1,2}$ \qquad Shihong Zhang$^{3}$ \qquad Christen Millerdurai$^{1}$ \\ 
Rüdiger Westermann$^{3}$ \qquad Didier Stricker$^{1,2}$ \qquad Alain Pagani$^{1}$ \vspace{0.5em}\\
$^{1}$German Research Center for Artificial Intelligence \quad
$^{2}$RPTU \quad
$^{3}$Technical University of Munich 
}

\begin{document}

\twocolumn[{
    \vspace{-5ex}     % 
    \maketitle
    \begin{center}
        \vspace{-4ex}
        \captionsetup{type=figure}
        \includegraphics[width=1.0\linewidth]{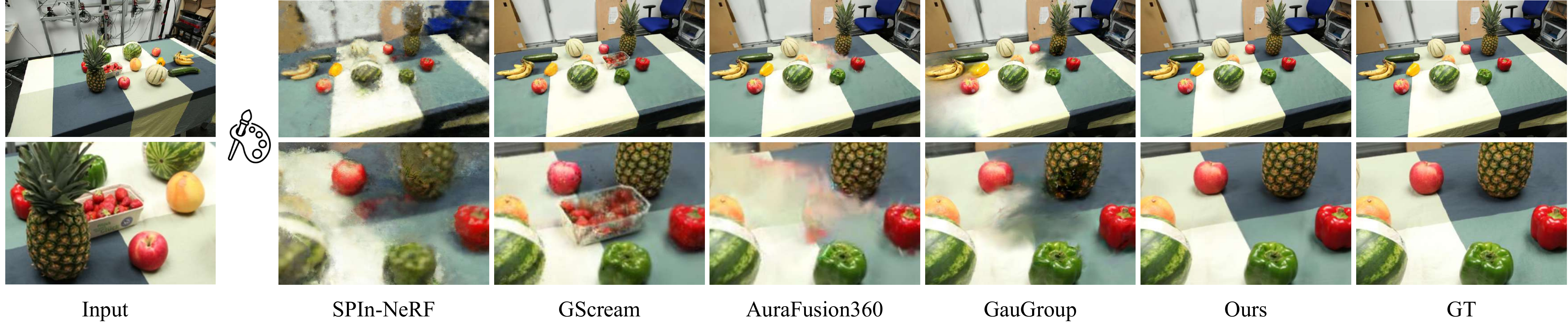}
        \vspace{-4ex}
        \captionof{figure}{We propose a novel object-aware 3D inpainting method, \emph{Inpaint360GS}, which flexibly enables object removal and inpainting in 360° scenes. Our approach effectively handles occlusions in multi-object environments and achieves better geometric and appearance consistency compared to existing state-of-the-art methods, including SPIn-NeRF~\cite{spinnerf}, GScream~\cite{gscream}, AuraFusion360~\cite{aurafusion360}, and GauGroup~\cite{gaussiangrouping}.}
        \vspace{-3ex}\
        \label{fig:teaser}
    \end{center}
}]

\vspace{-4mm}
\begin{abstract}
% \vspace{-1mm}

 Despite recent advances in single-object front-facing inpainting using NeRF and 3D Gaussian Splatting (3DGS), inpainting in complex $360^\circ$ scenes remains largely underexplored. This is primarily due to three key challenges: (i) identifying target objects in the 3D field of 360° environments, (ii) dealing with severe occlusions in multi-object scenes, which makes it hard to define regions to inpaint, and (iii) maintaining consistent and high-quality appearance across views effectively.

 To tackle these challenges, we propose Inpaint360GS, a flexible $360^\circ$ editing framework based on 3DGS that supports multi-object removal and high-fidelity inpainting in 3D space. By distilling 2D segmentation into 3D and leveraging virtual camera views for contextual guidance, our method enables accurate object-level editing and consistent scene completion. We further introduce a new dataset tailored for $360^\circ$ inpainting, addressing the lack of ground truth object-free scenes. Experiments demonstrate that Inpaint360GS outperforms existing baselines and achieves state-of-the-art performance. Project page: \href{https://dfki-av.github.io/inpaint360gs/}{https://dfki-av.github.io/inpaint360gs/}

\end{abstract}

\vspace{-13pt}

\vspace{-2mm}
\section{Introduction}
\vspace{-2mm}
\label{sec:intro}

Recent advances in 3D scene modeling, such as Neural Radiance Fields (NeRFs)~\cite{mildenhall2021nerf} and 3D Gaussian Splatting (3DGS)~\cite{gaussiansplatting}, have enabled realistic view synthesis and high-quality reconstruction. However, vanilla versions of these methods are not designed for scene editing tasks~\cite{gaussianeditor,3dsceneeditor} such as object removal or inpainting, especially in complex $360^\circ$ environments with multiple objects and occlusions. Existing approaches often assume front-facing, single-object setups, and struggle with consistent geometry recovery, object segmentation, and multi-view coherence.

Inpainting on $360^\circ$ 3D scene, this task poses three key challenges: (1) the need for an editable scene representation that supports object segmentation based on flexible prompts (\eg, via VLMs or clicks); (2) defining the underlying never-before-seen (NBS)  regions after object removal, especially under occlusion; and (3) ensuring fast and view-consistent inpainting that preserves structural and virtual continuity across multiple views.
Addressing these challenges requires a scene representation that is both editable and spatially explicit. While several NeRF-based methods~\cite{spinnerf, mvip, nerfiller, innerf360} attempt 3D inpainting, the implicit nature of radiance fields lacks explicit spatial boundaries, limiting object-aware editing. By contrast, 3DGS discretizes scenes into explicit Gaussian elements, supporting localized modification. 
Nevertheless, despite recent advancements in 3DGS-based approaches~\cite{gaussiangrouping, gscream, reffusion}, achieving efficient $360^\circ$ multi-view consistent 3D object inpainting remains an open challenge. Although some methods~\cite{imfine, aurafusion360, huang20253d} consider view consistency, they typically rely on predefined single-object masks and post refinements. These constraints significantly limit their flexibility for interactive multi-object segmentation. Moreover, the long optimization time required by such methods makes rapid scene editing infeasible.

To address these issues, we propose \emph{Inpaint360GS}, a novel framework for multi-object, multi-view consistent inpainting using 3D Gaussian Splatting. We distill 2D segmentation masks into a 3D Gaussian field to assign per-Gaussian object labels. To ensure geometric consistency across views, we leverage the depth information encoded in the Gaussians to guide the inpainting process without requiring explicit depth alignment. This enables fast convergence and high-fidelity results.
Unlike prior methods~\cite{spinnerf,gscream,aurafusion360,gaussiangrouping} that rely solely on given camera poses, our approach exploits the view synthesis capability of the 3D Gaussian field to generate virtual camera views centered around the removed objects. These virtual views provide enriched contextual information to guide the inpainting process.
Finally, to address the lack of datasets, we introduce a new $360^\circ$ benchmark dataset comprising indoor and outdoor scenes with single/multiple objects, along with corresponding object-free ground-truth sequences for quantitative evaluation.

\noindent In summary, our key contributions include:
\begin{itemize} 
\item 
A framework for consistent 2D mask association that integrates 2D segmentation masks into the 3D Gaussian scene representation. While existing works often focus on single-object, our method is explicitly designed for inpainting in 3DGS under multi-object scenarios.

\item 
An efficient depth-guided inpainting method that achieves multi-view completion with consistent structure and texture via virtual camera poses.

\item 
A new benchmark dataset featuring $360^\circ$ indoor and outdoor sequences containing single/multi objects with varying complexity, along with corresponding object-free ground-truth sequences.

\end{itemize}

\section{Related Work}

Efficient and flexible object-level 3D inpainting tasks integrate multiple techniques. To highlight our contributions, we focus the related work discussion on segmentation and inpainting methods that are most relevant to this task.

\noindent \textbf{3D Scene Segmentation}.
Recent advances in segmentation have been led by models such as SAM~\cite{segmentAnything}, HQSAM~\cite{hqsam}, and SEEM~\cite{seem}, which enable zero-shot 2D segmentation. Building on this progress, temporal methods~\cite{deva, heo2023generalized, segmentAndTracking, weng2023mask, li2024univs} propagate masks across video frames to maintain consistency over time.
Meanwhile, fully supervised 3D instance segmentation~\cite{rozenberszki2024unscene3d, sun2023superpoint, rozenberszki2022language} has shown promise results, but remains constrained by limited annotated data and often lacks explicit object-level representations due to the scarcity of densely annotated 3D datasets.

\begin{figure}[htbp]
  \centering
  \captionsetup{skip=0pt}
  \vspace{-2mm}
  \includegraphics[height=2.9cm]{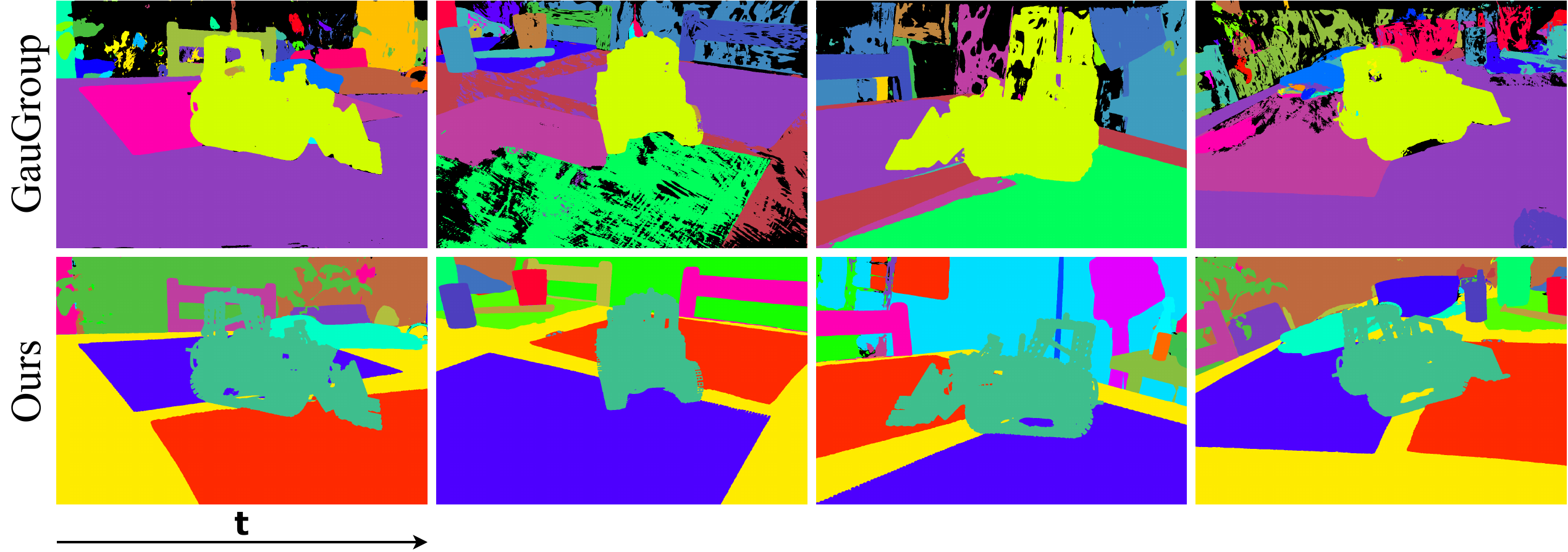}
  \caption{\textbf{Multi-View Segmentation Comparison.} Compared to GauGroup~\cite{gaussiangrouping} our method has more consistent segmentation results across different views.}
  \label{fig:segmentation}
\vspace{-5mm}
\end{figure}

To achieve spatially coherent segmentation in 3D, several methods distill 2D masks~\cite{sam3d,saga,gaga,saminNerf} into radiance fields, while others leverage language embeddings to ground semantics directly in 3D, either in Gaussian Splatting~\cite{langsplat,drsplat25} or through transformer-based visual grounding models such as MiKASA~\cite{chang2024mikasa}.
However, these methods are computationally intensive and unsuitable for interactive editing.
In contrast, approaches like DEVA~\cite{deva} improve multi-object handling by decoupling per-frame segmentation from temporal association, benefiting better scalability to multi-object scene applications such as semantic SLAM~\cite{snislam} and Gaussian-based modeling~\cite{semanticgaussian, gaussiangrouping}. Still, this video-level 2D label propagation often leads to segmentation errors, which degrade downstream tasks like inpainting and editing in GauGroup~\cite{gaussiangrouping}.
To overcome these challenges, our work proposes efficient segmentation association in 3D Gaussian field. By associating raw 2D segmentation outputs and aligning 2D masks in the Gaussian field, we ensure robust multi-view consistency and mitigate the spatial inconsistencies inherent in purely 2D-driven methods as shown in~\cref{fig:segmentation}.

\noindent \textbf{Inpainting}.
Classical inpainting methods, such as pixel diffusion~\cite{bertalmio2000image} and patch-based approaches~\cite{criminisi2004region}, struggle with large or semantically complex regions. Deep learning introduced generative inpainting using context-encoder GANs~\cite{pathak2016context, yu2018generative}, though early results were often blurry. Two-stage methods like EdgeConnect~\cite{nazeri2019edgeconnect} improve structure before texture, and recent diffusion-based models~\cite{lugmayr2022repaint, xie2023smartbrush, zhang2023towards, grechka2024gradpaint} offer higher-quality results at significant computational cost.
Extending inpainting to 3D requires appropriate scene representations. SPIn-NeRF~\cite{spinnerf} pioneered front-facing 3D inpainting via implicit fields. However, the more challenging 360° setting demands multi-view consistency, which is hard to achieve with per-view 2D inpainting. NeRF-based methods~\cite{weder2023removing, nerfiller, yin2023or, innerf360, mvip} attempt to integrate multi-view 2D inpainting with 3D optimization, but often suffer from inconsistency due to diffusion output's diversity and geometry misalignment, limiting them to bounded or small-angle scenes~\cite{lin2024maldnerf}.
Alternatively, Gaussian-based methods provide explicit scene representations that are inherently more suitable for flexible scene editing. Approaches such as InFusion~\cite{infusion}, AuraFusion360~\cite{aurafusion360}, and GScream~\cite{gscream} rely on depth foundation models~\cite{marigold,depthanything}, leading to depth alignment issues.
GauGroup~\cite{gaussiangrouping} injects semantics into Gaussians but remains sensitive to initialization and 2D segmentation quality. Recent works~\cite{aurafusion360, imfine, huang20253d} have further improved multi-view consistency and unseen region detection. Nonetheless, these methods struggle with severe occlusions in complex multi-object scenes, and editing remains costly due to depth scale misalignment and localized texture refinement. They also lack of strategies for selecting informative inpainting views, operating only on training poses.
To overcome these limitations, we define depth directly from the Gaussian field to eliminate scale ambiguity and introduce a conditional virtual view selection strategy, enabling high-fidelity inpainting and efficient convergence in unbounded $360^\circ$ environments.

\vspace{-2mm}
\section{Method} 

We propose an object-aware inpainting framework based on 3DGS.
In~\cref{sec:preliminary}, we review the 3DGS representation.
We introduce 2D mask association across views via a Key Object Management System in Gaussian field~(\cref{sec:labelAss}).
These labels are distilled into 3D~(\cref{sec:distillation}).
After object removal, virtual views are rendered to expose occluded regions~(\cref{sec:activ3d}).
We perform conditional 2D inpainting followed by depth-guided 3D inpainting with hybrid supervision~(\cref{sec:3dInp}).
A new benchmark dataset for $360^\circ$ inpainting is introduced in~\cref{sec:dataset}.

\subsection{Preliminaries}
\label{sec:preliminary}
3D Gaussian Splatting (3DGS) represents a 3D scene field using a set of Gaussians $G = \{g_i\}_{i=1}^{N}$ and employs a differentiable rasterizer~\cite{gaussiansplatting} for efficient rendering, where $N$ is the total number of Gaussians. Each Gaussian $g_i = \{\mathbf{p}_i, \mathbf{s}_i, \mathbf{q}_i, \mathbf{o}_i, \mathbf{c}_i\}$ is defined by its 3D center position $\mathbf{p}_i \in \mathbb{R}^3$, scaling factors $\mathbf{s}_i \in \mathbb{R}^3$, a quaternion $\mathbf{q}_i \in \mathbb{R}^4$ representing 3D orientation and covariance, an opacity value $\mathbf{o}_i \in \mathbb{R}$, and color coefficients $\mathbf{c}_i$ represented using spherical harmonics (SH).

After projecting the 3D Gaussians onto the 2D image plane, 3DGS utilizes the differentiable rasterizer to compute the final pixel color through $\alpha$-blending of depth-ordered Gaussians. The color $\mathbf{C}$ at a pixel is computed as:
% \vspace{-2mm}
\begin{equation}
    \mathbf{C} = \sum_{i \in \mathcal{N}} \mathbf{c}_i \alpha_i T_i,
    \label{eq:color_defination}
\end{equation}
where $\mathcal{N}$ is the set of Gaussians overlapping the pixel, $\alpha_i$ represents the influence of the $i$-th Gaussian, and $T_i$ is the accumulated transmittance defined as $T_i = \prod_{j=1}^{i-1}(1 - \alpha_j)$.

\subsection{2D segmentation mask association via 3D Gaussian}
\label{sec:labelAss}

Our 3D scene is represented using Gaussians, as described in Sec.~\ref{sec:preliminary}. To support object-level editing, each Gaussian must be assigned a unique and consistent object ID across views. A naïve approach projects Gaussians onto 2D masks from models like SAM~\cite{segmentAnything}, but these masks often produce inconsistent labels across viewpoints. GauGroup~\cite{gaussiangrouping} tackles this using DEVA~\cite{deva} to associate object masks across views by treating the image sequence as a video. Specifically, it fails under sparse-view settings. 

To address this issue, we introduce the Key Object Management System, a label association mechanism that ensures consistent object ID assignment for 3D Gaussians. Fig.~\ref{fig:segmentation} shows the resulting ID assignment of our more robust alternative.
This mechanism is analogous to the keyframe overlap check used in SLAM systems~\cite{niceslam, wang2024uni,eslam}, which measures the shared visible content between frames, but here it is adapted to assign view consistent 2D object labels to 3D Gaussian sets. 
The Key Object Management System maintains a Key Object Database, denoted as $\mathcal{D}_{\text{ID}}$, which maps object IDs to their corresponding Gaussian sets.
Suppose there are $Q$ distinct objects in the scene; then, we define the database as $\mathcal{D}_{\text{ID}} = \{ P_1, P_2, \dots, P_Q \}$, where each $P_i$ represents the set of Gaussians belonging to the $i$-th object.
Specifically, $P_i = \{ g^{1}_{i}, g^{2}_{i}, \dots, g^{m}_{i} \}$, where $g_i^k$ denotes the $k$-th Gaussian associated with the $i$-th object and $m$ is the total number of Gaussians for that object.

\begin{figure} [t]
\vspace{-2mm}
  \centering
  \includegraphics[height=3.5cm]{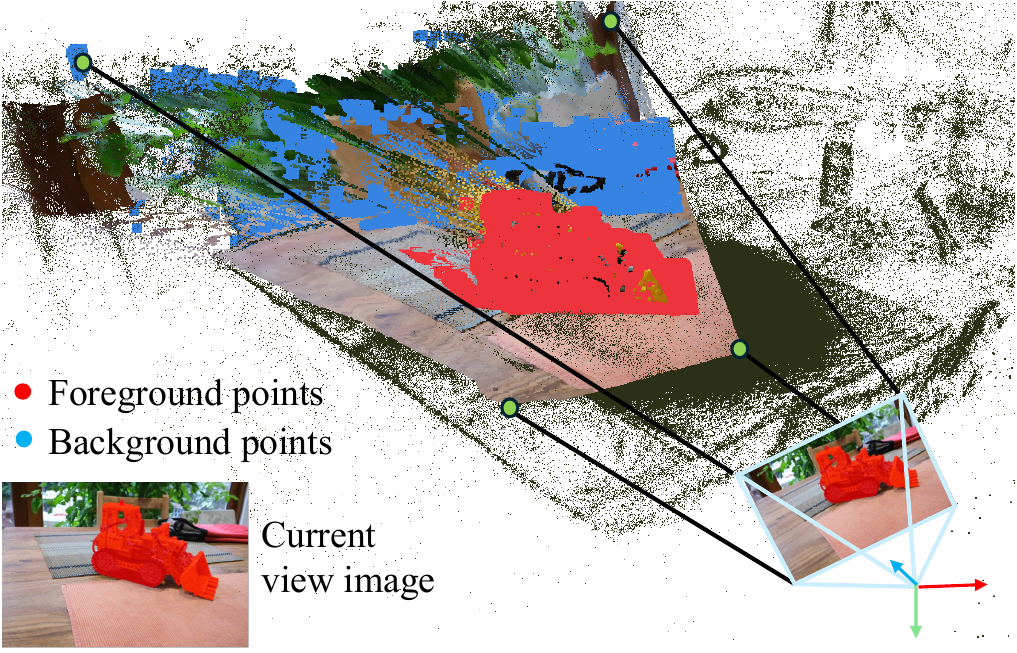}
  \vspace{-2mm}
  \caption{\textbf{Projection of 3D Gaussians onto 2D Segmentation.} 
  K-Means algorithm is employed to effectively distinguish between the foreground (\ie, target object) and background Gaussian points.
  }
  \label{fig:blue_red_mask}
  \vspace{-7mm}
\end{figure}

\begin{figure*}[htbp]
  \vspace{-4mm}
  \centering
  \includegraphics[height=7.75cm]{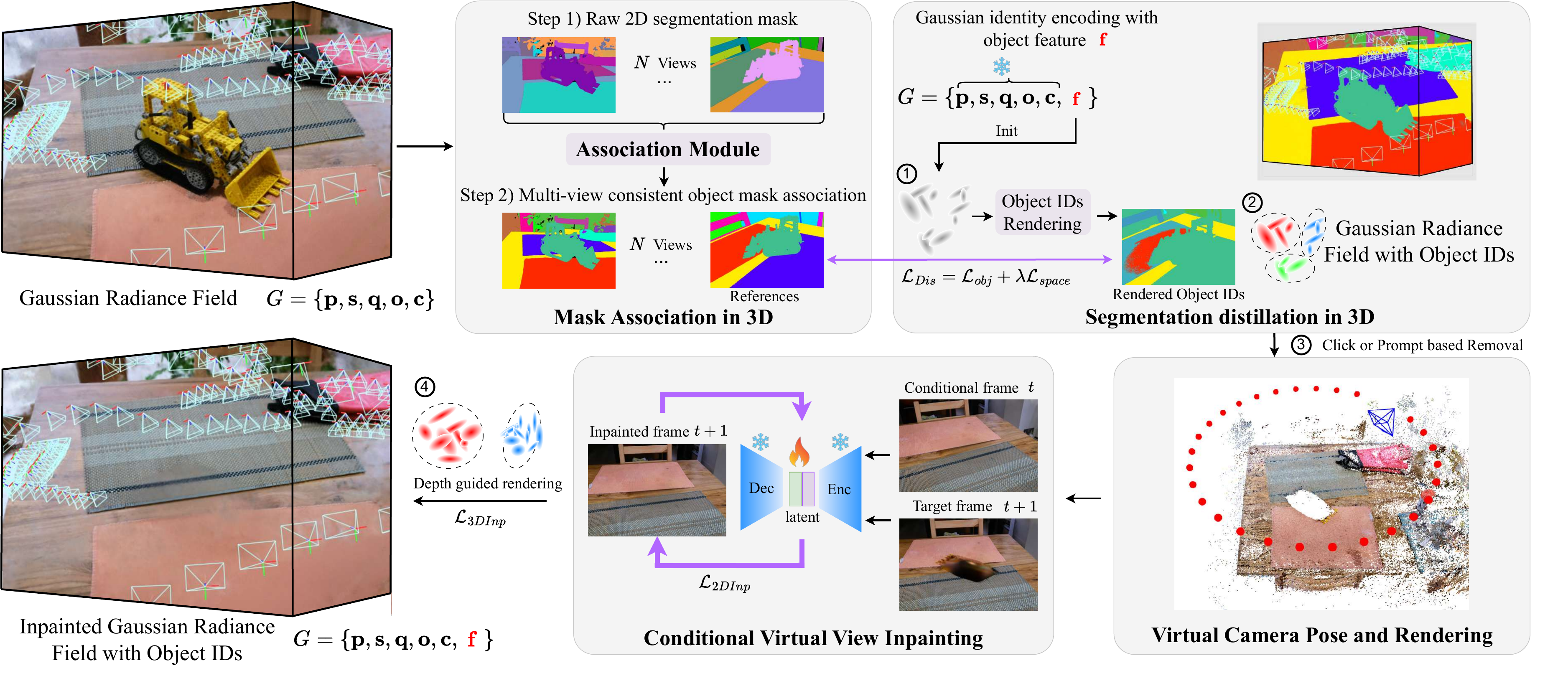}
  \vspace{-6mm}
  \caption{\textbf{Inpaint360GS Architecture Overview.}  Our framework takes a sequence of RGB images to construct a Gaussian Radiance Field (GRF) and extract per-view object masks using a 2D segmentation foundation model. By associating these masks across views within the GRF, we obtain multi-view consistent object masks and embed them into the Gaussian representation, assigning each Gaussian an object ID. This object-aware GRF enables direct 3D object manipulation, such as click-based or prompt-based removal. After removing target objects, we render at novel camera poses to obtain virtual views $\mathcal{V}$. During 2D inpainting, we recursively perform conditional RGB and depth inpainting, which is then used for depth-guided 3D inpainting.
 }
  \label{fig:pipeline}
  \vspace{-4mm}
\end{figure*}

\noindent \textbf{Key Object Database}.
To obtain the $P_i$ set of Gaussians belonging to the $i$-th object, we first project all Gaussians into 2D image coordinates using the corresponding camera poses (see Sec.~\ref{sec:preliminary}).
We then assign the 2D object labels to these Gaussians. 
However, not every projected Gaussian actually belongs to the object. 
As shown in Fig.~\ref{fig:blue_red_mask}, only the \textcolor{red}{red} points correspond to the truck’s foreground, while the \textcolor{blue}{blue} points belong to the background despite overlapping with the truck's segmentation.
To differentiate these, we apply K-Means clustering with $K=2$ in Euclidean space to partition the Gaussians into foreground (\ie, object) and background groups. 
We assign 2D segmentation labels only to the Gaussian cluster closer to the camera, ensuring accurate foreground association. This process is repeated across all training views to establish consistent object-Gaussian correspondences.

\noindent\textbf{Key Object Management System.} 
The Key Object Management System is used to merge and create new 
$P_i$ sets in the Key Object Database to ensure consistent object ID assignments across all frames.
For each view, we first assign temporary object IDs to Gaussians based on the 2D segmentation results.
Then, the Gaussians  $g_i$ associated with each object in the current view are compared with those stored in the Key Object Database $\mathcal{D}_{\text{ID}}$. 
To perform this comparison, we define the Gaussian Set Intersection-over-Union (GS-IoU) metric to quantify the overlap between Gaussian sets from different views.
Specifically, the GS-IoU between the $i$-th proposal and the $j$-th proposal is defined as:
\begin{equation} 
\text{GS-IoU}_{ij} = \frac{| P_i \cap P_j |}{| P_i \cup P_j |} 
\quad 0 \leq \text{GS-IoU}_{ij} \leq 1 ,
\label{eq:gs_iou} \end{equation}
where $P_i$ represents the set of Gaussian indices associated with the 
$i$-th object in the current view, and $P_j \in \mathcal{D}_{\text{ID}}$ denotes the set of indices for the $j$-th object stored in the database.
If the GS-IoU exceeds a threshold $\sigma$, the object is matched to an existing entry in the database, and its Gaussians inherit the corresponding object ID; otherwise, it is treated as a new instance with a new ID. After processing all views sequentially along a continuous camera poses, the Key Object Database contains roughly labeled Gaussians across viewpoints. We emphasize that these rough ID of Gaussians need not be perfect--they mainly serve to associate raw 2D segmentation masks across views, yielding consistent object masks $O$ that are later used as ground truth for the object ID distillation stage.
A concurrent method \cite{gaga} is the most comparable to ours, but it runs about five times slower and yields inferior accuracy. For further corner case (bird-view dense objects scenariom, sparse view case) analysis and visualization results, please refer to Supp.~Sec.~3.

\subsection{Efficient Object ID Distillation in 3D}
\label{sec:distillation}
Directly mapping object IDs from the Key Object Database often yields noisy or incomplete labels, resulting in unreliable point clouds. To address this, we distill the associated object mask from Key Object Database, ensuring consistency across views.

We distill the 2D object masks into the Gaussian field following the approach of GauGroup~\cite{gaussiangrouping}. Each Gaussian point is associated with a randomly initialized feature vector $f$ that represents its object ID embedding.
Next, we apply $\alpha$-blending to obtain a feature map:
\vspace{-0.5mm}
\begin{equation}
\mathbf{F} = \sum_{i \in \mathcal{N}} f_i \alpha_i T_i,
\label{eq:loss_2d}
\end{equation}
\vspace{-0.5mm}
where $\alpha_i$ denotes the influence of the $i$-th Gaussian, and $T_i$ represents the transmittance.
Subsequently, a linear transformation $\Phi(\cdot)$ projects the feature dimension to $Q$, corresponding to the total quantity of distinct objects in the scene. 
The resulting feature vectors are then processed with a \textit{softmax} for identity classification, i.e., $\hat{O} = \text{softmax}(\Phi(F))$, where $\hat{O} \in \mathbb{R}^{H \times W \times Q}$, with $H$ and $W$ representing the height and width of the image respectively.
We compute the 2D classification loss using the multi-class cross-entropy, i.e., $\mathcal{L}_{obj} = \text{CrossEntropy}(\hat{O}, O)$, where $O \in \mathbb{R}^{H \times W \times Q}$ is the associated 2D object mask with $Q$ classes (see Sec.~\ref{sec:labelAss}).
Additionally, we introduce 3D spatial supervision loss to complement the 2D supervision, which significantly accelerates convergence and enables more efficient distillation, particularly around complex object boundaries and fine structures.
For a given Gaussian point with feature vector $f_i$, we consider its $k$-nearest
neighbors $\mathcal{K}(f_i) = \{f_i^1, f_i^2, \dots, f_i^k\}$ in Euclidean space and encourage these neighboring features to be similar. 
We then define the \textit{spatial similarity loss} between $f_i$ and its neighbors as:
\begin{equation}
\mathcal{L}_{space} = 1 - \sum_{i \in k} \frac{f_i \cdot f_k}{\|f_i\| \|f_k\|}.
\label{eq:loss_space}
\end{equation}
The overall loss function for this distillation process is then given by
\begin{equation}
\mathcal{L}_{Dis} = \mathcal{L}_{obj} + \lambda\mathcal{L}_{space},
\label{eq:total_loss}
\end{equation}
where $\lambda$  is a balancing factor that regulates the contribution of the spatial consistency loss to the total loss.

\subsection{Virtual Camera Views for Inpainting}
\label{sec:activ3d}
With each Gaussian assigned a unique object ID, object removal via clicks or prompts becomes straightforward. After removal, only occluded or never-before-seen (NBS) regions(\eg, the base of the object), as most background areas remain visible from other views. Accurately identifying minimal NBS regions preserves valid content and reduces unnecessary inpainting. Prior work~\cite{gaussiangrouping} uses SAM-Tracking (SAMT)~\cite{segmentAndTracking} to detect NBS regions, but it fails under discontinuous frames. Recent methods~\cite{huang20253d,imfine} rely on iterative 3D-to-2D projections and learnable masks, but they are often inaccurate, computationally expensive, and limited to single-object scenarios.

In contrast, our method fully leverages the 3D Gaussian field's capability to synthesize novel views. We apply PCA-based pose alignment to generate a virtual circular trajectory centered on the removed object. Given an optimized 3D Gaussian scene $\mathcal{R}$, we first compute the object center and define a virtual trajectory $\mathcal{P} = \{p_i\}_{i=1}^L$ based on the original camera poses, where $p_i$ denotes the pose at frame $i$ and $L$ is the total number of views. For each pose $p_i$, we render an RGB image $C_i$ and its corresponding depth map $D_i$. 

\vspace{-5mm}
\begin{equation}
\begin{aligned}
\mathcal{V} = \big\{ (C_i, D_i, M_i) \,\big|\, 
& (C_i, D_i) = \texttt{render}(p_i, \mathcal{R}), \\
& M_i = \texttt{SAMT}(C_i) 
\big\}_{i=1}^L
\end{aligned}
\label{eq:virtual_views}
\end{equation}

For multi-object scenes, object occlusion could be addressed by leveraging lightweight object detectors (\eg, YOLOv8~\cite{yolov8}) to identify overlapping instances. Occluding objects around the target are temporarily removed to facilitate reliable NBS region mask $M_i$ extraction using SAMT~\cite{segmentAndTracking}, enabled by the smooth viewpoint transitions. The trajectory radius is adaptively controlled to ensure sufficient coverage of occluded regions without introducing extreme viewpoints. As a result, we obtain a set of virtual views $\mathcal{V} = \{(C_i, D_i, M_i)\}_{i=1}^L$, which serve as input for the inpainting stage.

\subsection{Depth Guided Multi-view Consistent Inpainting}
\label{sec:3dInp}
We address three key challenges in this module: (1) inpainting never-before-seen (NBS) regions on 2D images, (2) initializing inpainted content directly on the 3D scene surface for efficient integration, and (3) optimizing the inpainting process in 3D space.

\noindent \textbf{Recursive Conditional Inpainting.}  
A major challenge in achieving $360^\circ$ coherent rendering of the 3D Gaussian field lies in maintaining multi-view consistency during inpainting. Prior methods~\cite{spinnerf,gaussiangrouping,gscream,aurafusion360} are limited to fixed training camera views. For extreme viewpoints(\eg, oblique angles or views with very small NBS regions) 2D inpainting often results in poor textures and noticeable artifacts.

To overcome this, we leverage a set of continuous virtual frames $\mathcal{V}$. To avoid hallucination artifacts, we adopt Fourier convolutions LaMa~\cite{lama} as the inpainting model to fill the removed regions in the first virtual frame. Starting from the second frame, we use the previously inpainted frame as a conditional reference to guide the inpainting of the current frame. This recursive process ensures that each frame's texture is guided by the previous one, thereby maintaining temporal and visual consistency. Specifically, both the inpainted frame $C_t$ and the target frame $C_{t+1}$ are encoded into a shared latent space via a conditional encoder: $\ell_t, \ell_{t+1} = \text{Encoder}(C_t, C_{t+1})$. The resulting latent features are concatenated and optimized jointly in the feature space. The inpainting loss is defined as:

\vspace{-4mm}
\begin{equation}
\mathcal{L}_{\text{2DInp}} = \left\|(C_{t} - \hat{C}_{t}) \right\|_1 + \left\| M_{t+1} \odot (C_{t+1} - \hat{C}_{t+1}) \right\|_1.
\label{eq:2dinpaint_loss}
\end{equation}

The corresponding mask $M_{t+1}$ indicates the regions to be filled. $\hat{C}_{t},\hat{C}_{t+1}$ are decoded image after every step. After 10 optimization steps, the completed image is obtained by decoding the updated feature: $C_{t+1} = \text{Decoder}(\ell_{t},\ell_{t+1})$. This strategy effectively overcomes the issue of view discontinuity in the training dataset. Moreover, the recursive conditional guidance enforces temporal continuity of texture information, fully leveraging the capability of novel view synthesis in 3D.

\noindent \textbf{Depth-Guided Gaussian Initialization.}  
Initializing the 3D point cloud is critical for successful Gaussian Splatting reconstruction. While Infusion~\cite{infusion} relies on a depth completion model, AuraFusion360~\cite{aurafusion360} and GScream~\cite{gscream} adopt Marigold~\cite{marigold} for zero-shot depth estimation followed by scale alignment via diffusion models. However, these methods introduce additional dependencies and substantially increase training time. 

Instead, we leverage the intrinsic properties of the Gaussian field to define the depth as:
\begin{equation}
\mathbf{D} = \sum_{i \in \mathcal{N}} z_i \alpha_i T_i,
\label{eq:depth_defination}
\end{equation}
where $z_i$ is the $z$-coordinate of the $i$-th Gaussian in the camera coordinate system, $\alpha_i$ denotes the influence of $i$-th Gaussian, and $T_i$ is the accumulated transmittance. Since missing depth regions typically exhibit low texture complexity than color image, they can be effectively inpainted using models like LaMa~\cite{lama}. Given the inpainted depth $D_{inp}$ and the corresponding color image $C_{inp}$, we fuse them with the inpainting mask $M$ to obtain a point cloud for the NBS region, which is then used to initialize the Gaussians.

\begin{figure}[tb]
  \centering
  \includegraphics[height=1.7cm]{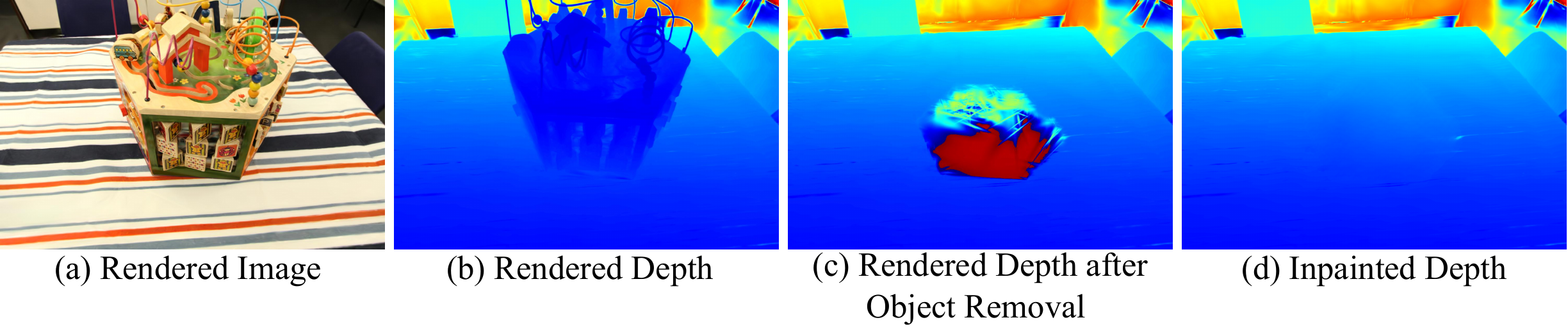}
  \vspace{-7mm}
  \caption{\textbf{Depth Completion.} 
  Leveraging the inherent structure of the scene, our method performs depth inpainting without requiring explicit depth alignment.
  }
  \label{fig:depth_inpainting}
  \vspace{-4mm}
\end{figure}

\noindent \textbf{3D Inpainting.} During the 3D scene inpainting phase, an intuitive idea is to make the Gaussians in the remaining (non-masked) regions non-trainable, and optimize only those within the masked areas.  
However, empirical observations indicate that this strategy tends to produce noisy textures and unstable boundary transitions.To address this, we propose a \emph{3D hybrid supervision scheme} that combines localized and global objectives. Specifically, we supervise masked regions using $\mathcal{L}_{1}$ and $\mathcal{L}_{\textrm{LPIPS}}$ losses, while enforcing global structural consistency with SSIM computed over the entire image:

\vspace{-4mm}
\begin{align}
\mathcal{L}_{\text{3DInp}} =\; & (1 - \lambda_{1}) \left\| M \odot (C_{inp} - \hat{C}) \right\|_1 
 \notag \\
& + \lambda_{1} \mathcal{L}_{\textrm{D-SSIM}}(C_{inp},\hat{C}) + \lambda_{2} \mathcal{L}_{\textrm{LPIPS}}(C_{inp},\hat{C},M).
\label{eq:3dinpaint_loss}
\end{align}
\vspace{-5mm}

Here, \( M \) denotes the binary inpainting mask, \( \hat{C} \) the rendered image, and \( C_{\text{inp}} \) the inpainted result used as reference. Unlike SSIM, which is sensitive to localized inconsistencies when computed within small masks, applying it over the full image stabilizes optimization and improves boundary smoothness.

% \subsection{Dataset for $360^\circ$ Inpainting}
\subsection{\texorpdfstring{Dataset for $360^\circ$ Inpainting}{Dataset for 360° Inpainting}}
\label{sec:dataset}

Existing radiance field datasets are unsuitable for $360^\circ$ inpainting due to several limitations. Datasets like NeRF2NeRF~\cite{nerf2nerf}, MipNeRF360~\cite{mipnerf360}, and LERF~\cite{lerf2023} lack object-free(without object) scenes, making quantitative evaluation infeasible. While SPIn-NeRF~\cite{spinnerf} offers object-free ground truth, it is limited to front-facing views and indoor scenes, with photometric inconsistencies caused by varying camera settings. Other datasets~\cite{aurafusion360, imfine} lack multi-object scenarios and suffer from test-view leakage in point clouds, further undermining the validity of quantitative evaluations.
To address these issues, we introduce a new $360^\circ$ inpainting dataset with object-inclusive and object-free sequences. It contains 11 scenes: 7 single-object and 4 multi-object settings with occlusions, covering diverse indoor and outdoor environments. Camera parameters (exposure, white balance, ISO) are fixed to eliminate photometric variation. To ensure fair evaluation, test-view point clouds are excluded from training. See Supp.~Sec.~1 for details.

\section{Experiments and Results}
\label{sec:experiment}

\begin{table*}[ht]
\vspace{-5mm}
% \normalsize
\small

\centering
\renewcommand{\arraystretch}{1.3}
\setlength{\tabcolsep}{3pt}  
\begin{tabular}{ l |  c @{\hspace{10pt}} c @{\hspace{10pt}} c @{\hspace{10pt}} c @{\hspace{10pt}} c @{\hspace{10pt}} c @{\hspace{10pt}} c @{\hspace{10pt}}  c }
\hline 
 Methods              & PSNR $\uparrow$ &masked PSNR $\uparrow$ & SSIM $\uparrow$ & masked SSIM$\uparrow$ & LPIPS $\downarrow$ & masked LPIPS $\downarrow$ & FID $\downarrow$   \\
\hline 
 
 SPIn-NeRF \cite{spinnerf}          & 19.71  & 34.53    & 0.5000  & 0.9854    & 0.5002  &0.0140 & 229.95  \\
 GScream \cite{gscream}             & 20.95  & 28.47    & 0.7380  & 0.9819    & 0.2715  & 0.0161  & 206.25  \\
 AuraFusion360 \cite{aurafusion360}    & 23.15  & 35.78    & 0.7923  & 0.9872    & 0.1915  & 0.0097  & 47.71  \\
 GauGroup \cite{gaussiangrouping}   & 23.20  & 35.73    & 0.7928  & 0.9862    & 0.1770  & 0.0102  & 65.87  \\
 Inpaint360GS (Ours)                & \textbf{24.40}  & \textbf{36.29}    & \textbf{0.8370}  & \textbf{0.9886}    & \textbf{0.1300}  & \textbf{0.0078}  & \textbf{35.93}  \\
\hline                     

\end{tabular}
\vspace{-2mm}
\caption{\textbf{Quantitative comparison of 360° inpainting methods on the Inpaint360GS dataset.}}
\label{tab:quantitativeEva}
\vspace{-2mm}
\end{table*}

\begin{figure*}[htbp]
  \centering
  \captionsetup{skip=0pt}
   \vspace{-0mm}
  \includegraphics[height=5.1cm]{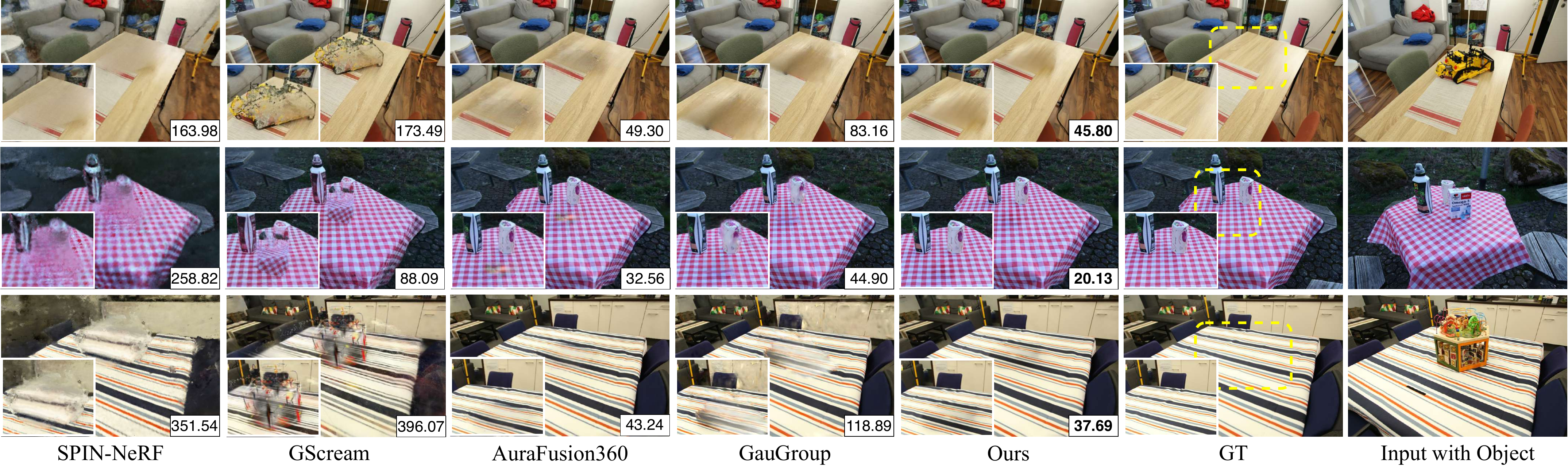}
  \caption{\textbf{Inpainting Result Comparison on our Inpaint360GS dataset.} We compare our method with the single-view inpainting approach GScream~\cite{gscream} and the multi-view inpainting methods SPIn-NeRF~\cite{spinnerf},AuraFusion360~\cite{aurafusion360} and GauGroup~\cite{gaussiangrouping}. The metric FID is reported at the right corner. Our approach achieves superior inpainting performance across various scenarios. Please zoom in for details. For per-scene multi-view results, please refer to Supp.~Sec.~4.}
  \label{fig:inpainting_result_inpaint360gs}
\vspace{-4mm}
\end{figure*}

\subsection{Experimental setup.}
\noindent \textbf{Datasets.}  
We evaluate Inpaint360GS across multiple benchmarks:
(1) Inpaint360GS (ours): A new dataset with 11 scenes (7 single-object, 4 multi-object). All experiments are conducted at $1/4$ resolution, and evaluations are performed on object-free test images.  
(2) Additional benchmarks: To demonstrate scalability, we test on three extra scenes collected from Mip-NeRF 360~\cite{mipnerf360}, Instruct-NeRF2NeRF~\cite{nerf2nerf}, and LERF~\cite{lerf2023}.

\noindent \textbf{Metrics.}  
We evaluate visual quality using PSNR, SSIM~\cite{ssim}, LPIPS~\cite{lpips}, and Frechet Inception Distance (FID)~\cite{fid}. All metrics are computed on both full images and NBS region in the Inpaint360GS test set. For external datasets lacking object-free ground truth, we provide qualitative comparisons. 

\noindent \textbf{Baselines and Implementation.} 
We compare our methods with four recent baseline methods: SPIn-NeRF~\cite{spinnerf},  GScream~\cite{gscream}, AuraFusion360~\cite{aurafusion360} and GauGroup~\cite{gaussiangrouping}. We retrain and test the model using their open-source code. All experiments are conducted on a single NVIDIA H100 GPU. For more implementation details, please refer to Supp.~Sec.~2.

\subsection{Evaluation against State-of-the-Art Methods}

\begin{figure*}[htbp]
  \centering
  \captionsetup{skip=0pt}
   \vspace{-4mm}
  \includegraphics[height=7.7cm]{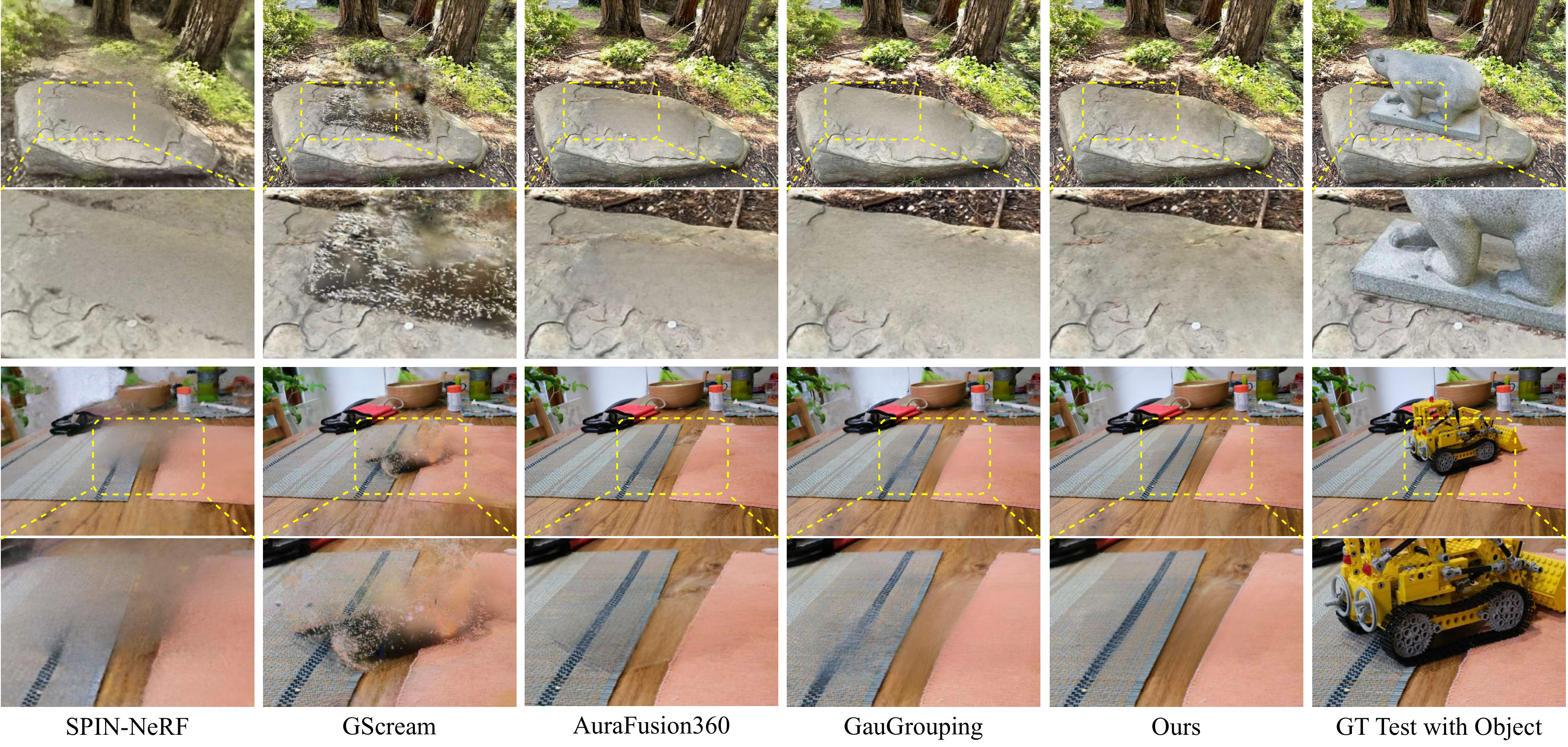}
  \caption{\textbf{Inpainting Result Comparison on Instruct-NeRF2NeRF~\cite{nerf2nerf} and Mip-NeRF 360~\cite{mipnerf360}.} Our method produces visually plausible 3D inpainted textures with smooth and coherent boundaries.}
  \label{fig:inpainting_result_kitchen_bear}
\vspace{-4mm}
\end{figure*}

\noindent \textbf{Qualitative comparisons.} 
 Results on the Inpaint360GS dataset are shown in ~\cref{fig:teaser} and~\cref{fig:inpainting_result_inpaint360gs}. Our method demonstrates superior texture quality and achieves the best FID score, which highlights the effectiveness of our pipeline design. The virtual camera poses enable accurate identification of NBS regions, while the conditional virtual view inpainting ensures consistent texture generation across multiple views. In~\cref{fig:inpainting_result_kitchen_bear}, we present inpainting results on the \texttt{bear} and \texttt{kitchen} scenes. The rightmost column provides a reference image containing the target object. Compared with other baselines, our method achieves noticeably smoother boundaries and more plausible texture synthesis. We attribute this to our conditional inpainting guided by virtual camera poses.

\noindent \textbf{Quantitative Evaluation.} 
In~\cref{tab:quantitativeEva}, we report PSNR, SSIM, LPIPS, and FID metrics on the Inpaint360GS dataset for both masked regions and full images. Our method consistently outperforms all baselines across all metrics. Front-facing inpainting baselines like SPIn-NeRF~\cite{spinnerf} and GScream~\cite{gscream} are fundamentally limited in 360° inpainting due to their lack of multi-view awareness. Although AuraFusion360~\cite{aurafusion360} targets 360° scenes, it struggles in complex multi-object scenarios, where depth misalignments arise from unreliable NBS region identification as illustrated in~\cref{fig:teaser}. GauGroup~\cite{gaussiangrouping} supports multi-view inputs, but inconsistent object IDs hinder reliable object removal. In contrast, our method demonstrates robust performance in both single-object and multi-object scenarios. This robustness is primarily attributed to the consistent object IDs maintained across views, which enable reliable cross-view reasoning. Additionally, the use of virtual camera poses allows accurate localization of the NBS regions, while the conditional virtual view inpainting effectively enforces multi-view consistency.

In addition,~\cref{tab:runtimeMemory} summarizes the runtime and memory consumption of all methods, evaluated on an NVIDIA H100 GPU using the \texttt{kitchen} scene from Mip-NeRF 360~\cite{mipnerf360} and the \texttt{bear} scene from Instruct-NeRF2NeRF~\cite{nerf2nerf}. 
For ours and GauGroup, the vanilla model includes object ID information. The reported inpainting time accounts for both 2D and 3D inpainting stages.
In terms of efficiency, our method exhibits two major advantages. First, it maintains consistent object identities across views, which facilitates flexible scene editing. Compared to GauGroup, our approach achieves higher rendering quality while using a model that is 30\% more compact. Second, the inpainting stage is $5\text{-}10\times$ faster than existing SOTA methods, enabling interactive usage. This efficiency is primarily attributed to accurate depth estimation, which removes the need for explicit alignment, and significantly accelerates the 3D inpainting process. Detailed runtime analysis can be found in Supp.~Sec.~3.

\begin{table}[ht]
\vspace{0mm}
% \footnotesize
% \scriptsize
\tiny
\centering
\renewcommand{\arraystretch}{1.3}
\setlength{\tabcolsep}{3pt}  
\begin{tabular}{ c   l |  c @{\hspace{5pt}} c @{\hspace{5pt}} c @{\hspace{5pt}} c @{\hspace{5pt}}  c }
\hline 
\multirow{2}{*}{Scene} & \multirow{2}{*}{\begin{tabular}{c} Method \end{tabular}}& \multirow{2}{*}{Object ID}  &Vanilla model  &Inpainting time$\downarrow$ & Total Time$\downarrow$ &  Storage$\downarrow$ \\ 
                       &                                                         &            & training$\downarrow$/Mins & Mins           & Mins                   & MB                  \\        
\hline 
\multirow{5}{*}{\makecell{\rotatebox[origin=c]{90}{bear\cite{nerf2nerf}}}} 
 & SPIn-NeRF \cite{spinnerf}          &\textcolor{red}{\ding{55}}   & 79 & 196 & 275     & 336   \\
 & GScream \cite{gscream}             &\textcolor{red}{\ding{55}}   & -- & 52  & 52      & \textbf{73.2}  \\
 & AuraFusion \cite{aurafusion360}    &\textcolor{red}{\ding{55}}   & 25 & 26  & 51      & 448.9 \\
\cline{3-7}    
 & GauGroup \cite{gaussiangrouping}   &\textcolor{green}\checkmark  & 55   & 20 & 75    & 774.8 \\
 & Ours                               &\textcolor{green}\checkmark  & \textbf{21.5} & \textbf{2.5} & \textbf{24}      & 477.5 \\
\hline                     

\multirow{5}{*}{\rotatebox{90}{kitchen\cite{mipnerf360}}} 
& SPIn-NeRF \cite{spinnerf}           &\textcolor{red}{\ding{55}}   & 59 & 148 & 207     & 336   \\
& GScream \cite{gscream}              &\textcolor{red}{\ding{55}}   & -- & 30  & 30      & \textbf{67.9}  \\
& AuraFusion \cite{aurafusion360}     &\textcolor{red}{\ding{55}}   & 20 & 43  & 63      & 183.5 \\
\cline{3-7}        
& GauGroup \cite{gaussiangrouping}    &\textcolor{green}\checkmark  & 27 & 13 & 40   & 897.4 \\
& Ours                                &\textcolor{green}\checkmark  & \textbf{12} & \textbf{3}  & \textbf{15}   & 663.5 \\
\hline 
\end{tabular}
\vspace{-2mm}
\caption{\textbf{Runtime and Model Size Comparison.} All unnecessary intermediate outputs are disabled to ensure fair comparison across methods.}
\label{tab:runtimeMemory}
\vspace{-1mm}
\end{table}

\subsection{Design Choice and Ablation Study}

\begin{table}[htbp]
\vspace{-2mm}
\footnotesize
\centering
\renewcommand{\arraystretch}{1.3}
\begin{tabular}{c @{\hspace{10pt}} lccccc} % First column changed from 'l' to 'c'
\hline &Method                 & PSNR    &  SSIM   & LPIPS  & FID   \\
\hline 
& a) w/o obj. association         & 23.31   & 0.7921  & 0.1932 & 66.75 \\
& b) w/o depth guidance           & 24.15   & 0.8199  & 0.1256 & 35.75 \\
& c) w/o virtual camera pose       & 24.18   & 0.7987  & 0.1574 & 38.74 \\
& d) w/o cond. inpainting         & 24.23   & 0.8156  & 0.1420 & 37.57 \\
& e) w/o 3D hyb. supervison       & 24.01   & 0.7997  & 0.1398 & 38.42 \\
& f) Ours                         & \textbf{24.40}   & \textbf{0.8370}  & \textbf{0.1300} & \textbf{33.93} \\
\hline 
\end{tabular}  
\vspace{-2mm}
\caption{\textbf{Ablation on Inpaint360GS dataset.} }
\label{tab:ablation}
\vspace{-7mm}
\end{table}

\noindent \textbf{Effectiveness of Object Mask Association.}
We compare our object mask association strategy with that of DEVA~\cite{deva}, which is adopted by GauGroup~\cite{gaussiangrouping}. As shown in~\cref{tab:ablation} a), using DEVA-generated masks for scene reconstruction results in noticeably degraded performance. In~\cref{fig:segmentation}, we provide qualitative evidence of the robustness and cross-view consistency of our method. Moreover, we validate the reliability of our mask association under challenging scenarios, including densely packed objects, bird's-eye viewpoints, and sparse input configurations. Additional visualizations and analyses are provided in Supp.~Sec.~3.

\noindent \textbf{Effectiveness of Depth Guidance.}
Depth guidance substantially contributes to the efficiency of the 3D inpainting process. As reported in~\cref{tab:ablation} b), removing depth guidance leads to a noticeable decline in reconstruction quality. This highlights the importance of accurate geometric priors in accelerating convergence and enhancing final performance.

\begin{figure}[htbp]
  \centering
  \captionsetup{skip=0pt}
  \vspace{-2mm}
  \includegraphics[height=3cm]{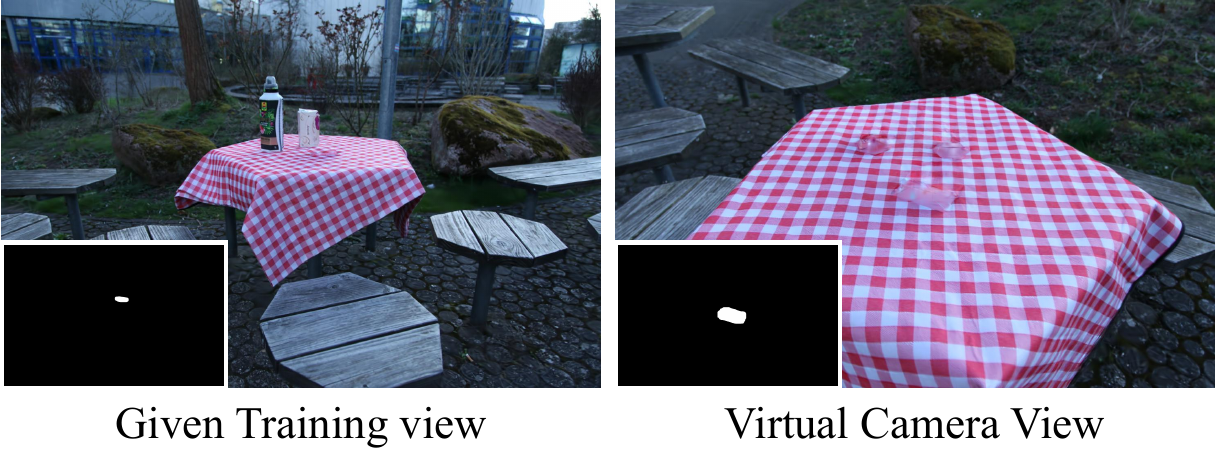}
  \caption{\textbf{Ablation on virtual camera view.} Compared to the original training views, virtual camera views provide better visibility for NBS regions by overcoming the limitations of extreme viewing angles and occlusions.}
  \label{fig:ablation_virtual_camera}
\vspace{-4mm}
\end{figure}

\noindent \textbf{Effectiveness of Virtual Camera Pose.}
Virtual camera poses help mitigate the challenges introduced by extreme viewpoints in the training data. In~\cref{fig:ablation_virtual_camera}, we demonstrate the detected NBS region. Moreover, our method leverages flexible object identity to perform occlusion-aware inpainting. Specifically, we detect occluded instances using object detection and temporarily remove them before inpainting. After the inpainting process, the temporarily removed objects are reinserted into the scene. This strategy allows the system to better exploit contextual information from the surrounding environment.~\cref{tab:ablation} c) shows the ablation on it.

\begin{figure}[htbp]
  \centering
  \captionsetup{skip=0pt}
  \vspace{-1mm}
  \includegraphics[height=3.7cm]{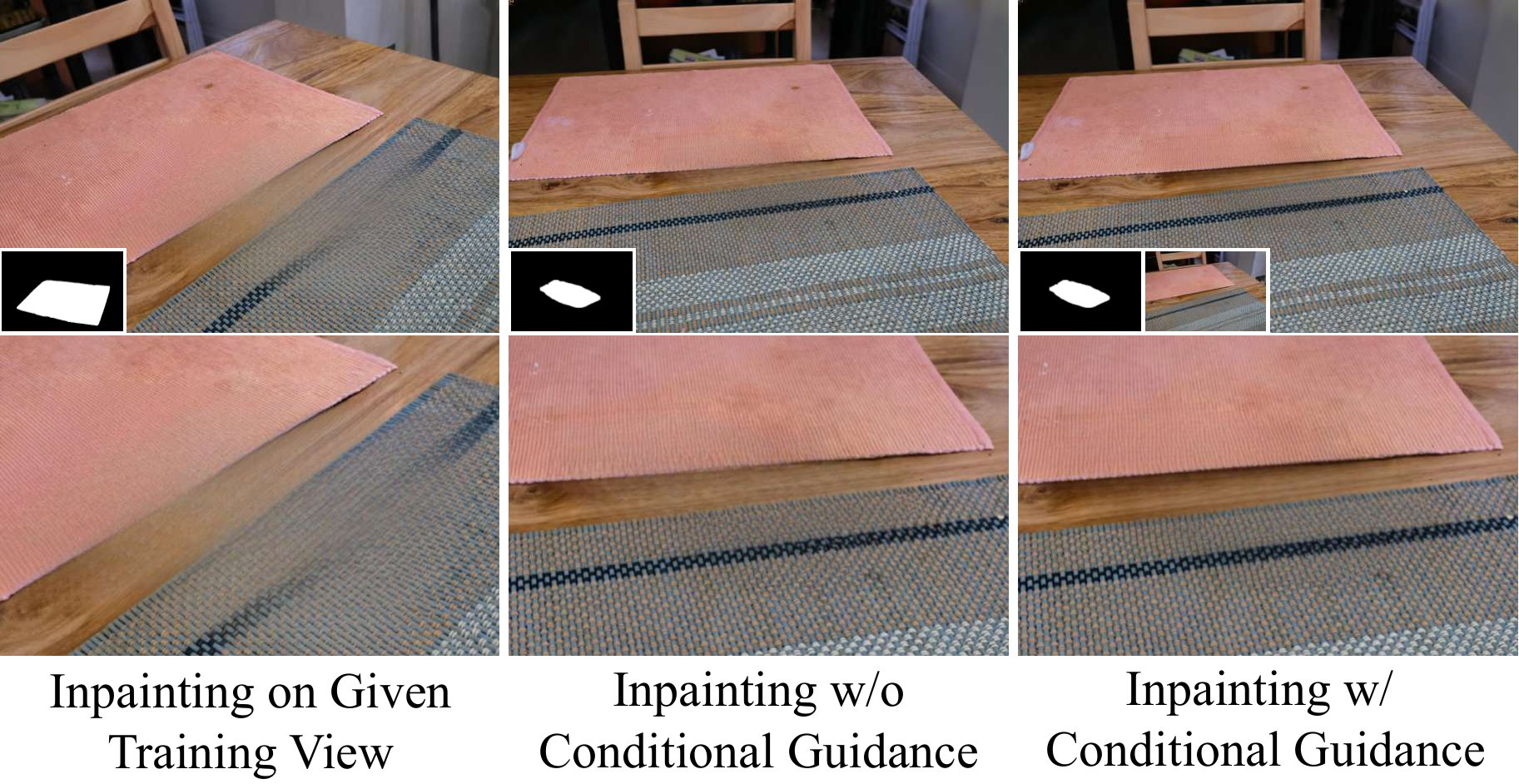}
  \caption{\textbf{Ablation on conditional inpainting.}}
  \label{fig:ablation_conditional_inp}
\vspace{-2mm}
\end{figure}

\noindent \textbf{Effectiveness of Conditional Inpainting.}
We adopt a conditional inpainting strategy in which each 2D inpainting step is guided recursively by the previously rendered frame. The use of continuous camera poses facilitates the propagation of consistent texture context across views. As shown in~\cref{tab:ablation} d) and~\cref{fig:ablation_conditional_inp}, removing this strategy leads to a noticeable decline in inpainting quality.

\begin{figure}[htbp]
  \centering
  \captionsetup{skip=0pt}
   \vspace{-1mm}
  \includegraphics[height=2.07cm]{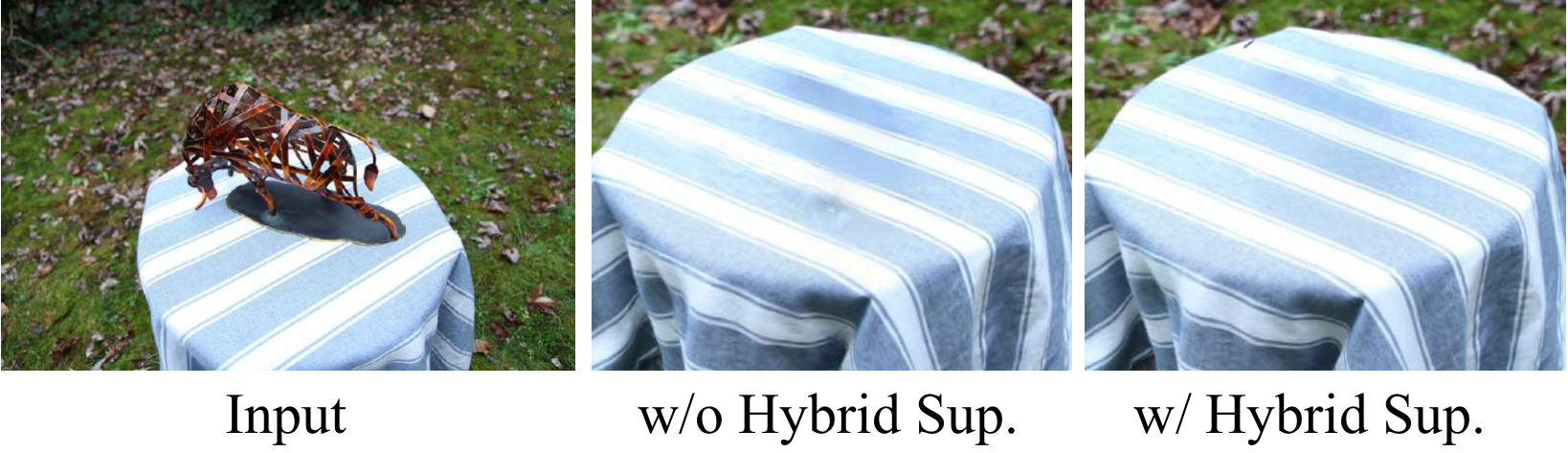}
  \caption{\textbf{Ablation on 3D Hybrid Supervision.}}
  \label{fig:ablation_hybrid_sup}
\vspace{-7mm}
\end{figure}

\noindent \textbf{Effectiveness of 3D hybrid supervision.} In~\cref{tab:ablation} e) and~\cref{fig:ablation_hybrid_sup}, we show that employing the proposed 3D hybrid supervision significantly improves inpainting quality compared to the naive masking-only strategy.

\vspace{-0mm}
\section{Conclusion}
\vspace{-1mm}

Inpaint360GS is a novel object-aware inpainting framework based on 3D Gaussian Splatting in $360^\circ$ scenes. 
By distilling 2D segmentation masks into 3D space and leveraging a virtual-view, depth-guided inpainting strategy, our method enables faster convergence while ensuring structural and photometric consistency.
We also introduce a benchmark dataset for $360^\circ$ inpainting with object-inclusive and object-free data, and extensive experiments show that Inpaint360GS significantly outperforms SOTA methods.
Despite these promising results, our approach occasionally exhibits residual shadow artifacts cast by the removed objects and struggles with inpainting irregular complex textures, which remain to be explored in future work.

\noindent\textbf{Acknowledgements:} 
This work has been partially supported by the EU projects CORTEX2 (GA No. 101070192) and LUMINOUS (GA No. 101135724), as well as the project ARROW by the German Research Foundation (DFG, GA No. 564809505).

\clearpage
{
    \small
    \bibliographystyle{ieeenat_fullname}
    \bibliography{main}
}


\clearpage

\clearpage
\appendix
\setcounter{page}{1}
\maketitlesupplementary
\setcounter{section}{0}
\setcounter{equation}{0}
\renewcommand{\thesection}{\Alph{section}}

%%%%%%%%% ABSTRACT

\begin{abstract}
In the supplemental material, we provide additional details about the following:
\begin{itemize}
    \item Dataset Details.
    (Section~\ref{sec:dataset_details})
    \item Implementation Details. (Section~\ref{sec:imp_detail})
    \item Additional Ablation Study and Experiment Analysis. (Section~\ref{sec:additional_analysis_ablation})
    \item Per-Scene Breakdown of the Results. (Section~\ref{sec:per_scene_resuls})
\end{itemize}

\end{abstract}

\begin{figure*}[!htbp]
  \centering
  \includegraphics[height=9.5cm]{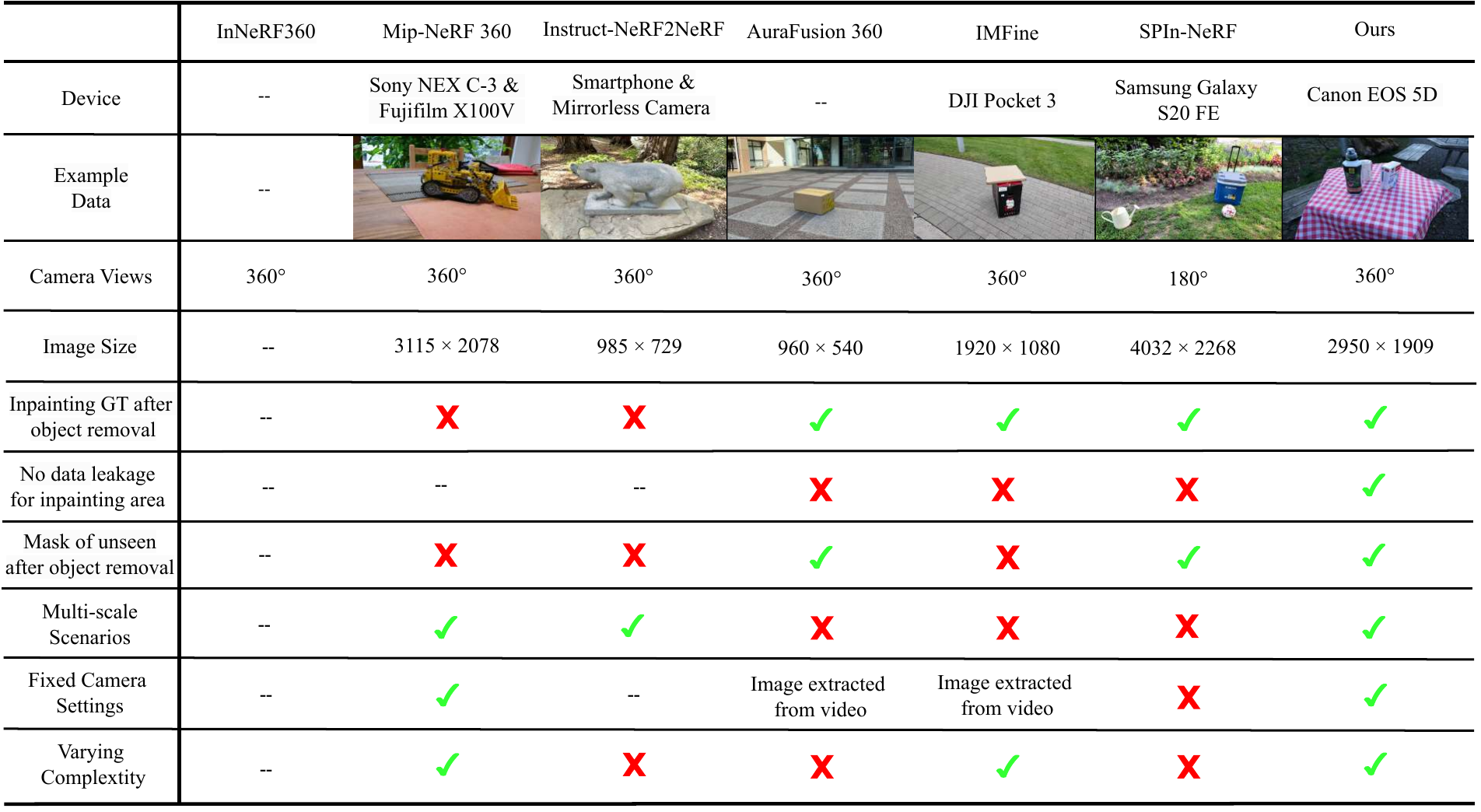}
  \caption{\textbf{Dataset Comparison.} We compare our new dataset with existing datasets commonly used for inpainting tasks, including unpublished InNeRF360~\cite{innerf360}, Mip-NeRF 360~\cite{mipnerf360}, AuraFusion 360~\cite{aurafusion360}, IMFine~\cite{imfine}, SPIn-NeRF~\cite{spinnerf}, and Instruction-NeRF2NeRF~\cite{nerf2nerf}. Our dataset is designed for well-structured 360° inpainting scenarios, including challenging multiple occluded objects, no data leakage in the inpainting regions of the point cloud, and consistent camera settings within each scene.} 
  \label{fig:dataset_detail}
\vspace{-0mm}
\end{figure*}

\section{Dataset Details}
\label{sec:dataset_details}
We provide a comprehensive analysis of datasets employed in our study, highlighting the limitations of existing datasets and motivating the introduction of a novel dataset specifically designed for 3D 360° inpainting evaluation.

\noindent \textbf{Mip-NeRF 360~\cite{mipnerf360}.} This dataset comprises professionally captured 360° imagery obtained with high-end cameras. It features exceptional image quality, carefully curated scenes, and precisely calibrated camera parameters. However, Mip-NeRF 360 lacks ground truth for after object removal scenarios, thereby precluding its use for quantitative assessment of 3D inpainting performance.

\noindent \textbf{Instruction-NeRF2NeRF~\cite{nerf2nerf}.} This dataset provides complete 360° views and encompasses a wide variety of scenes. Nonetheless, similar to Mip-NeRF 360, it does not include ground truth for post-removal conditions. In addition, its relatively lower image quality and limited resolution, while sufficient for current methodologies, may not meet the demands of future advances in 3D inpainting.

\noindent \textbf{AuraFusion 360~\cite{aurafusion360}.} This dataset includes only a single object per scene and lacks challenging multi-object, complex environments. Our dataset addresses this limitation by incorporating scenes with multiple occluded objects. In Gaussian-based inpainting, where accurate point cloud initialization is critical, we ensure fair evaluation by excluding any inpainting-view-specific points. In contrast, AuraFusion360 suffers from data leakage, as its sparse point cloud includes points visible only in inpainting views. Moreover, frames extracted from video often lack sufficient quality.

\noindent \textbf{IMFine~\cite{imfine}.} Similar to AuraFusion 360, the IMFine dataset also suffers from data leakage from test set. In addition, it does not provide masks for regions that become visible only after object removal. This lack of ground-truth masking makes it impossible to distinguish between masked and background areas, thereby preventing meaningful quantitative evaluation such as FID calculation on the inpainted regions. The frames are extracted from the video as well.

\noindent \textbf{SPIn-NeRF~\cite{spinnerf}.} SPIn-NeRF provides ground truth for inpainting following object removal, addressing a crucial limitation of Mip-NeRF360. However, its scope is restricted to front-facing views, limiting its applicability to full 360° inpainting tasks. Additionally, the dataset primarily consists of small, enclosed environments, thereby constraining its utility to a narrow range of inpainting applications. Furthermore, inconsistent camera parameters (such as ISO, exposure, and white balance) between the original and post-removal captures introduce unintended variations in scene appearance. This discrepancy compromises the reliability of the ground-truth data, rendering quantitative evaluation meaningless, as the observed differences may stem from photometric inconsistencies rather than actual inpainting errors.

\noindent \textbf{Our Dataset.} To overcome the aforementioned limitations, we introduce a new high-quality dataset specifically designed for 360° inpainting with quantitative evaluation. This dataset is acquired using diverse imaging devices across scenes of varying scales and incorporates multiple difficulty levels within the same scene to better accommodate future developments in 3D inpainting.

To ensure diversity in scene scales and realistic application scenarios, we employ a combination of DSLR cameras, and drones for data collection. Large-scale outdoor scenes are captured using a DJI Mini 2 drone, which is equipped with a 24 mm f/2.8 lens with a fixed focus range. For smaller outdoor scenes, we utilize a Canon 5D with an 24-105 mm zoom lens, fixed at its widest focal length (24 mm). This choice minimizes focal length variations, thereby reducing geometric distortions, perspective inconsistencies, and optical aberrations, facilitating subsequent processing. 

For each scene, we manually configure white balance, ISO, shutter speed, aperture, and focus based on a reference image, and keep these settings fixed throughout the capture process to ensure photometric consistency across frames. For indoor scenes, we utilize large diffuse light sources and LED spotlights to mitigate strong cast shadows. In outdoor environments, we capture scenes under overcast conditions. Overcast conditions produce soft shadows that minimally affect scene illumination.

Each scene consists of 100-200 images, during which target objects are manually moved to facilitate dynamic scene acquisition. The dataset consists of two main parts. The first part includes all objects in the scene. The second part serves as the ground truth for inpainting, where targeted objects are removed to introduce novel viewpoints, enabling quantitative evaluation of inpainting performance. To ensure that both the training and test inpainting datasets share a consistent coordinate system, we process them jointly using the publicly available COLMAP~\cite{colmap} software to obtain camera poses and a sparse point cloud. Within each scene, cameras share a single set of intrinsic parameters, and we adopt a pinhole camera model for undistortion. Importantly, to prevent data leakage, we remove point cloud regions corresponding to the test-time occluded region, a crucial step that has often been overlooked in prior works~\cite{aurafusion360,imfine}. 

Regarding the mask of the object, we use SAM~\cite{hqsam} method and our proposed mask association to connect with each other to get unified object ID. With the selected object ID, we can get the object mask per image. In addition, in order to evaluate the unseen area after object removal and background respectively. We also prepared the mask of the unseen region after object removal for this dataset.

\section{Implementation Details}
\label{sec:imp_detail}
\noindent \textbf{Gaussian Field Initialization.}  
We initialize our scene using the default settings from the original 3D Gaussian Splatting framework.  
Notably, we operate in evaluation mode, where only \(7/8\) of the training data is used for training, while the remaining \(1/8\) interval-sampled data is reserved for evaluation.

\noindent \textbf{Mask Association.}  
To obtain raw 2D segmentation masks, we employ the 2D segmentation foundation model HQSAM~\cite{hqsam}. The model is used with their default parameter configurations. During the association stage, we set a predefined GS-IoU threshold \( \sigma = 0.2 \) for matching objects in the Key Object Database. To improve association accuracy per view, each image is divided into $16 \times 16$ patches, and mask matching is performed at the patch level. The maximum number of object categories allowed in the classification process is 256.

\noindent \textbf{Object Feature Distillation.}  
To distill object features from the 2D associated object masks into the 3D Gaussian Field, we randomly initialize each Gaussian with a 16-dimensional feature vector \( f_i \) to represent its identity. For neighbor aggregation, we apply a k-nearest neighbor (KNN) strategy with \( k = 5 \). Additionally, a linear transformation \( \Phi(\cdot) \)
 projects the feature dimension to \( Q \), where \( Q \) represents the quantity of object categories obtained during mask association, with a maximum of 256. In the overall loss function, we set the weighting factor \( \lambda = 0.0005 \). The optimization process is conducted over 2000 iteration steps.

\noindent \textbf{Virtual Camera Views.}  
For the virtual camera views $\mathcal{V} = {(I_j, D_j, M_j)}_{j=1}^L$, we utilize 90$\%$ of the known training camera poses and the object center in world coordinates to initialize the virtual camera centers. These centers are distributed along a circular trajectory whose radius is adaptively determined based on the area of the NBS region mask. Notably, a smaller camera path radius brings the virtual camera closer to the object, which typically results in a larger NBS region mask. A too-large inpainting region may lead to failure cases for the 2D inpainter. Specifically, we empirically the mask area to lie within 1$\%$ to 50$\%$ of the full image area to ensure effective inpainting. 

\noindent \textbf{Object Removal and 2D Inpainting.}  
For object removal, we leverage SAM-Tracking~\cite{segmentAndTracking} to enable both prompt-based and click-based interactive segmentation. Once an object is identified, all Gaussian points corresponding to the object, including those computed using the Delaunay convex hull, are removed from the scene.

During the 2D inpainting stage, the input is the rendered scene where removed objects create empty regions. We use SAM-Tracking to generate the corresponding inpainting masks. They are from virtual camera views $\mathcal{V}$. The rendered image after removal object, corresponding mask and last inpainted image are fed into the LaMa inpainting model~\cite{lama} to reconstruct missing regions. The encoder and decoder of LaMa are frozen, while latent representation $(\ell_{t},\ell_{t+1})$ is trainable here. A similar approach is applied for depth inpainting, ensuring structural consistency across views.

During the 2D inpainting stage, the input consists of rendered images with missing regions caused by object removal. Inpainting masks are generated using SAM-Tracking. The masked image, corresponding inpainting mask, and the previously inpainted image are fed into the LaMa inpainting model~\cite{lama} to reconstruct the missing content. While the encoder and decoder of LaMa are frozen, the latent representations $(\ell_t, \ell_{t+1})$ extracted from rendered images remain trainable. Optimization steps we set 10 here. A similar procedure is applied for depth inpainting to ensure structural consistency across views. The above inpainting process is executed on virtual camera views $\mathcal{V}$.

\noindent \textbf{3D Inpainting.}  
We initialize the NBS region of the Gaussian field using depth-color fusion from the first inpainted color and depth images of the virtual camera view. During the 3D inpainting stage, we set the loss weights to \( \lambda_{1} = 0.2 \) and \( \lambda_{2} = 0.005 \), and perform optimization for 2000 iterations.

% \begin{algorithm*}[!t]
\begin{algorithm}[ht]
    % \footnotesize % or \scriptsize for even smaller text
    \scriptsize
    \caption{\textit{Inpaint360GS}
    }
		\label{alg:optimization}
		\begin{algorithmic}
                \State RGB images              \Comment{Input}  
			\State $p \gets$ SfM Points	\Comment{Sparse point position and camera pose in 3D}
                \State $p, s, \alpha, c\gets$ OptimizedAttributes() \Comment{Position, covariances, opacities,  colors through 3DGS~\cite{gaussiansplatting}}
                \State $m = (m_1, m_2, \ldots, m_K ) \gets$ Zero-shot 2D Segmentation	\Comment{\textbf{SAM's masks} at Various $K$ Views}
                \State $(O_1, O_2, \ldots, O_K ) \gets$ Mask association through Key Object Management\Comment{Multi-view \textbf{consistent associated masks} in 3D}
			\State $f \gets \text{identity vector}$
	\Comment{Initialize identity vector for each Gaussian}
    		\State $(p, s, \alpha, c, \textcolor{red}{f})\gets$ FreezeParam() \Comment{Freeze all parameters except identity vector $\textcolor{red}{f}$ }
			\While{not converged}
			\State $V, C, O \gets$ SampleTrainingView()	\Comment{Camera view, image and mask}
			\State $\hat{C}, \hat{D}, \hat{O} \gets$ Rasterize($p$, $s$, $\alpha$, $c$, $\textcolor{red}{f}$,$V$)	\Comment{Rendered image, rendered depth and identity mask}
			\textcolor{red}{\State $\mathcal{L}_{Dis} \gets \mathcal{L}_{\text{obj}}(O, \hat{O}) + \lambda\mathcal{L}_{\text{space}}(\textbf{$f$, $f^1, f^2, \dots, f^k$})$} \Comment{Distillation Loss function}
			\State $\textcolor{red}{f}$ $\gets$ Adam($\nabla \mathcal{L}_{Dis}$) \Comment{Backprop \& Step}
			\EndWhile
                \State $\mathcal{V} = \{(C_i, D_i, M_i)\}_{i=1}^L$             \Comment{Virtual camera view after object removal}
                \State $C_{inp}, D_{inp}$ $\gets$ ConditionLaMa($\mathcal{V}$) \Comment{Inpainted color and depth}
                \State $\textcolor{red}{\mathcal{R}_{inp}}$ $\gets$ $C_{inp}, D_{inp}$      \Comment{Initialize Gaussian field $\textcolor{red}{\mathcal{R}_{inp}}$ for NBS region}
                \While{not converged}
                \State  \textcolor{red}{$\mathcal{L}_{\text{3DInp}} \gets (1 - \lambda_{1})\mathcal{L}_{1}(C_{\text{inp}},\hat{C},M) + \lambda_{1} \mathcal{L}_{\textrm{D-SSIM}}(C_{\text{inp}},\hat{C})$} 
                \State  \textcolor{red}{$+ \lambda_{2} \mathcal{L}_{\textrm{LPIPS}}(C_{\text{inp}},\hat{C},M)$}                                              \Comment{3D inpainting loss function}
                \State $\textcolor{red}{\mathcal{R}_{inp}}$ $\gets$ Adam($\nabla \mathcal{L}_{3DInp}$) \Comment{Backprop \& Step}
                \EndWhile
		\end{algorithmic}
            \label{alg:gaussian_alg}
 % \end{algorithm*}
 \end{algorithm}
 
\subsection{Proof of the Validity of Depth Definition}

\begin{figure*}[htbp]
  \centering
  \includegraphics[height=3.7cm]{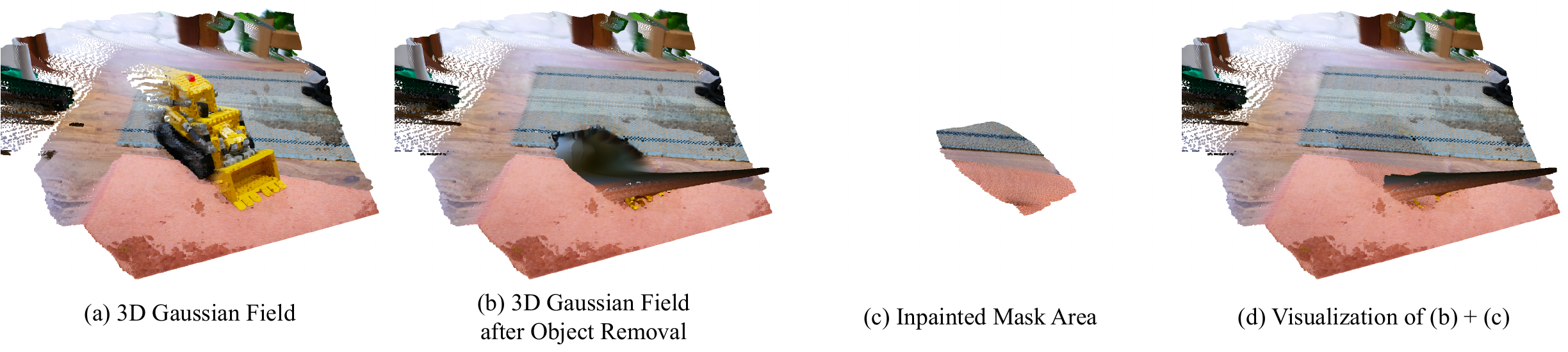}
  \caption{\textbf{Validity of Depth Definition.} While (a) and (b) represent the point clouds generated from the Gaussian field under the given camera pose before and after object removal, respectively, (c) is constructed via color-depth fusion between the inpainted image and the depth defined in~\cref{eq:sup_depth_defination}. The point cloud in (c) can be effectively used as initialization for the 3D inpainting stage.}
  \label{fig:depth_validation}
\end{figure*}

A typical neural point-based approach (\eg,~\cite{kopanas2022}) computes the color $C$ of a pixel by blending $\mathcal{N}$ ordered points overlapping the pixel:
\begin{equation}
	\label{eq:sup_color_defination}
	C = \sum_{i \in \mathcal{N}}
	c_{i}\alpha_i
	\prod_{j=1}^{i-1}(1-\alpha_{j}) = \sum_{i \in \mathcal{N}}
	c_{i}\alpha_{i}T_{i} = \sum_{i \in \mathcal{N}}c_{i}w_{i},
\end{equation}
where $\mathbf{c}_i$ is the color of each point and $\alpha_i$ is given by evaluating a 2D Gaussian with covariance $\Sigma$ ~\cite{yifan2019} multiplied with a learned per-point opacity. $T_{i} = \prod_{j=1}^{i-1}(1-\alpha_{j})$ is transmittance after passing ${i}$ gaussian point.

Similarly, depth is defined as
\begin{equation}
	\label{eq:sup_depth_defination}
	D = \sum_{i \in \mathcal{N}}
	z_{i}\alpha_{i}
	\prod_{j=1}^{i-1}(1-\alpha_{j}) = \sum_{i \in \mathcal{N}}z_{i}w_{i}
\end{equation}
where $z_i$ is z-coordinate in the camera coordinate system.

Our goal is to prove the weight along a current sampling ray $r$ as:
$$ 
w_i = 1 - T_i = w_{i-1} + T_{i-1}\alpha_{i} 
$$

% \vspace{-20mm}

$$
\begin{aligned}
w_i &= w_{i-1} + T_{i-1}\alpha_{i} \\
    &= w_{i-2} + T_{i-2}\alpha_{i-1} + T_{i-1}\alpha_{i}\\
    &...\\
    &= T_{0}\alpha_{1} + T_{0}\alpha_{1} + ... + T_{n-1}\alpha_{n} \\
    &= (T_{0} - T_{1}) + (T_{1} - T_{2}) + ... + (T_{n-1} - T_{n}) \\
    &= T_{0} - T_{n} = 1 -T_{n}
\end{aligned}
$$

To validate the effectiveness of our depth definition, we present visualizations in~\cref{fig:depth_validation}. Subfigures (a) and (b) show point clouds rendered from the 3D Gaussian field before and after object removal, respectively. In (c), we visualize the fused point cloud generated by combining the inpainted RGB image and the estimated depth defined in~\cref{eq:sup_depth_defination}. Notably, unlike (a) and (b), which are derived directly from the Gaussian field, (c) is obtained through depth-color fusion. When using (c) as the initialization for the 3D inpainting stage on (b), the resulting reconstruction (d) demonstrates strong geometric consistency, validating our initialization strategy. This approach avoids the depth alignment issues present in~\cite{gscream,aurafusion360,infusion}.

\section{Additional Ablation Study and Experiment Analysis}
\label{sec:additional_analysis_ablation}

\noindent{\textbf{Detailed Time Analysis of pipeline:} We report the runtime breakdown of different stages in our pipeline for the \texttt{bear} and \texttt{kitchen} scenes, corresponding to Tab.~2 in the main paper. The “Pure 3DGS” time refers to the training time required to learn the Gaussian field without any editing components. Adding the time for Mask Association and Distillation yields the total time for the “Vanilla Gaussian” baseline in Tab.~2. The “Inpainting” time includes both 2D and 3D inpainting steps.

\begin{figure}[b]
  \centering
  \captionsetup{skip=0pt}
  \includegraphics[height=3cm]{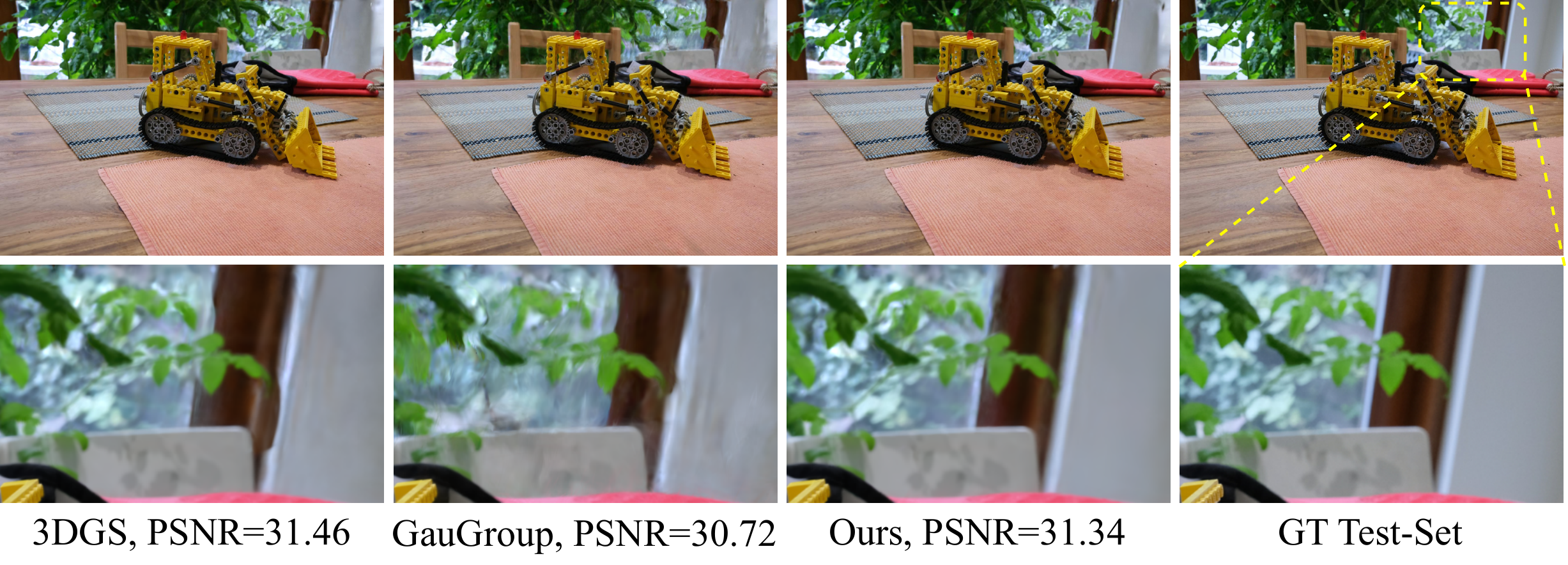}
  \vspace{-0mm}
  \caption{\textbf{RGB Rendering Comparison.} While GauGroup\cite{gaussiangrouping} sacrifices rendering quality in color fidelity to incorporate object IDs, our method achieves comparable PSNR$[\text{dB} \uparrow]$ to the naive 3DGS\cite{gaussiansplatting} method. Please zoom in for details.}
  \label{fig:render_color}
\vspace{-0mm}
\end{figure}

\begin{table}[htbp]
\scriptsize  % \tiny
\centering
\renewcommand{\arraystretch}{1.3}
\setlength{\tabcolsep}{3pt}  
\begin{tabular}{c | l  c @{\hspace{5pt}} c @{\hspace{5pt}} c @{\hspace{5pt}} c }
\hline 
\textbf{Tab.~\textcolor{red}{3}} \hspace{0em}  & Pure 3DGS & Mask Association & Distillation & Inpainting & Total \\  
\hline 
 \texttt{bear}    & 17 mins   & 2.5 mins   & 2 mins     & 2.5 mins  & 24 mins \\
 \texttt{kitchen} & 8 mins    & 3 mins     & 1 mins     & 3 mins    & 15 mins \\
\hline                     

\end{tabular}
\caption{\textbf{Detailed Runtime and Model Size Comparison.}}
\label{tab:runtimeMemory_supp}
\end{table}

\noindent \textbf{Analysis of the Effectiveness of Consistent Object ID Mask on Rendering.}  
Compare our two-stage method with the one-stage semantic Gaussian method, GauGroup~\cite{gaussiangrouping}. Our approach achieves superior global consistency, not only for foreground objects but also for the background.  
\Cref{fig:render_color} compares the rendering quality of our method with that of GauGroup~\cite{gaussiangrouping}. Our approach achieves superior rendering quality due to more consistent multi-view segmentation masks and a training strategy that independently optimizes the 3D Gaussian Splatting (3DGS) and the integration of semantic masks. Due to the incorporation of object masks, the background geometry is further refined compared to vanilla 3DGS~\cite{gaussiansplatting}.

\noindent \textbf{Analysis of Object Removal Accuracy.}  
In \cref{fig:mipnerf_kitchen_removal}, we compare the performance of our method in target object removal. The results demonstrate that our approach achieves more precise object removal and produces a more accurate inpainting-ready base. This indicates that our method can more effectively assign consistent spatial Gaussian representations, leading to better convergence without misclassified surrounding artifacts. To quantitatively assess the accuracy of object mask identification, we introduce the Average Mask Coverage Ratio (AMCR), defined as:

\begin{equation}
\text{AMCR} = \frac{1}{N} \sum_{k=1}^{N} \left( \frac{\| M_k \|_1}{H \times W} \times 100\% \right)
\label{eq:amcr}
\end{equation}

It quantifies the proportion of empty regions in the image after object removal, averaged over the $N$ training images. For each image, the binary mask $M_k \in {[0,1]}^{H \times W}$ indicates pixels to be inpainted, with 1 denoting removed regions. A lower AMCR value implies more accurate object segmentation and less redundant inpainting area, which typically leads to better reconstruction performance.

\begin{figure}[ht]
  \centering
  \captionsetup{skip=0pt}
    \vspace{-0mm}
  \includegraphics[height=8cm]{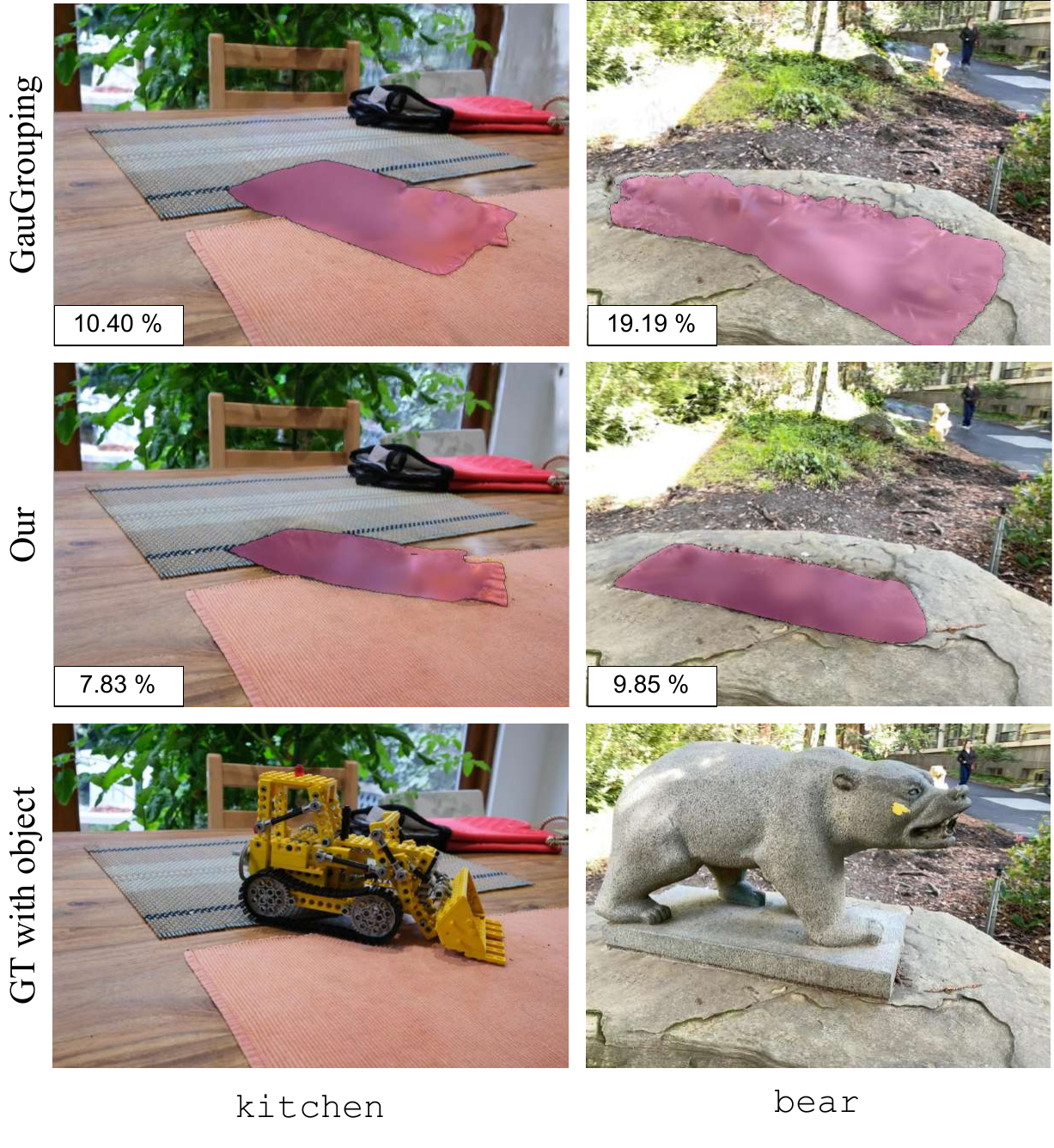}
  \caption{\textbf{Object Removal Comparison.} Our method accurately removes the target object, demonstrating superior 3D segmentation compared to GauGroup~\cite{gaussiangrouping}. A more precise removal leads to better inpainting results. We report the Average Mask Coverage Ratio(AMCR) $[\% \downarrow]$, indicating the proportion of empty regions in the image, lower values reflect better segmentation effectiveness.}
  \label{fig:mipnerf_kitchen_removal}
\vspace{-0mm}
\end{figure}

\noindent \textbf{Ablation on Loss Term.} 
In \cref{fig:ablation_space_loss}, we validate the effectiveness of the spatial similarity loss function described for object ID distillation. 
The results demonstrate that incorporating this loss significantly improves artifact removal and preserves complex object boundaries during object removal.
\begin{figure}[t]
  \centering
  \captionsetup{skip=0pt}
      \vspace{-0mm}
  \includegraphics[height=2.7cm]{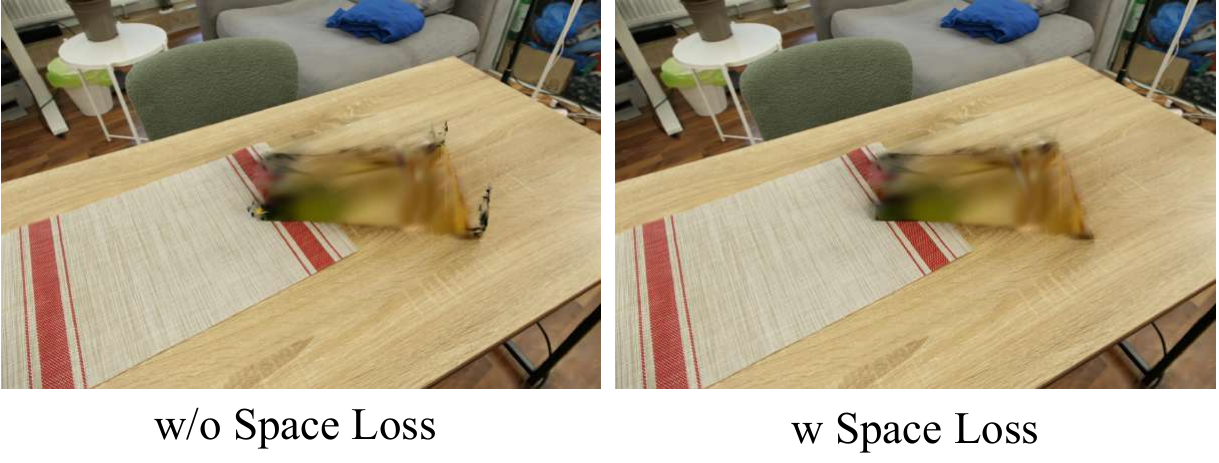}
\caption{\textbf{Ablation on Spatial Similarity Loss.} Without the spatial similarity loss, object removal on complex structures leaves significant artifacts.}
  \label{fig:ablation_space_loss}
\vspace{0mm}
\end{figure}

\noindent \textbf{Ablation on Depth-guided Inpainting.} 
In \cref{fig:ablation_depth}, we demonstrate that incorporating a depth prior dramatically accelerates convergence, achieving a reasonably good result within only $200$ steps. 
\begin{figure}[ht]
  \centering
  \captionsetup{skip=0pt}
      \vspace{-0mm}
  \includegraphics[height=3cm]{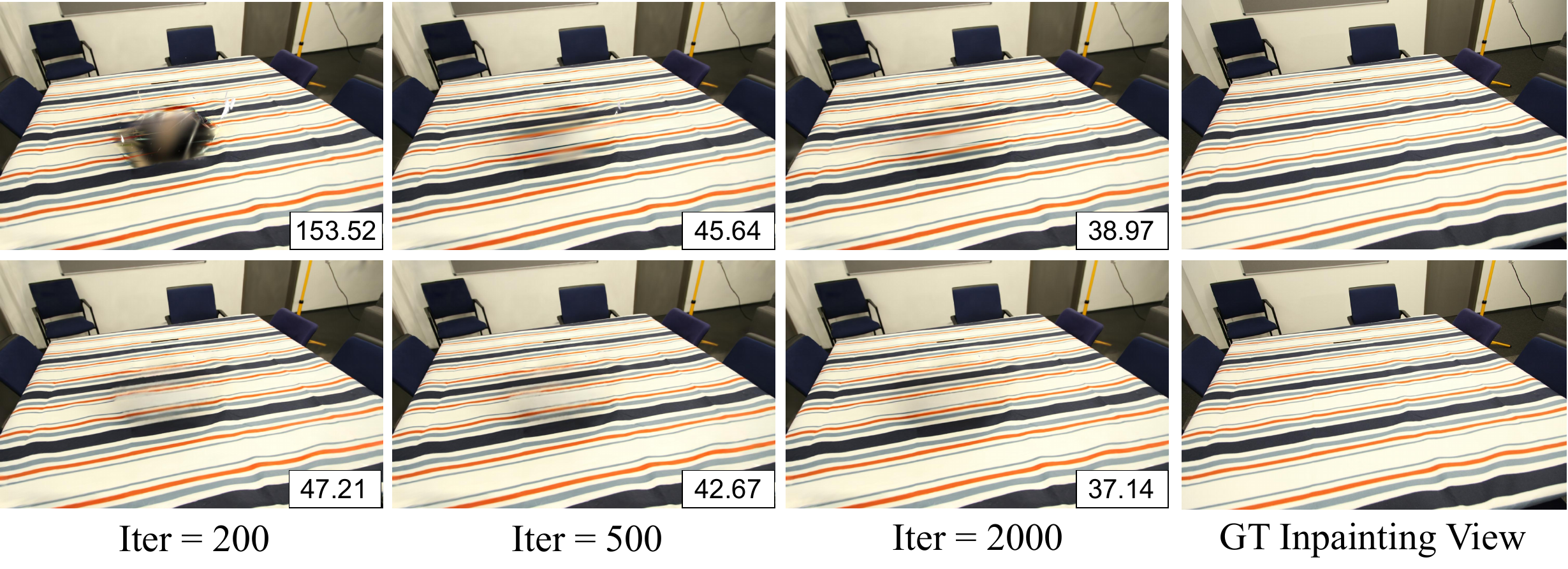}
  \caption{\textbf{Ablation on Depth-guided Inpainting.} We report the FID score [$\downarrow$] here. With depth-guided inpainting we can achieve faster convergence and better quality.} 
  \label{fig:ablation_depth}
\vspace{0mm}
\end{figure}

\noindent \textbf{Ablation on 2D Segmentation Foundation Model Selection.}
As shown in \cref{fig:ablation_sam_vs_hqsam_bear}, while Gaga~\cite{gaga} adopts SAM~\cite{segmentAnything} and utilizes 20$\%$ of the Gaussians within the mapped region to distinguish foreground and background, our method employs HQSAM~\cite{hqsam} combined with K-means clustering for this task. Driven by a more compact loss function, our approach achieves a 5$\times$ speed-up in overall efficiency, enabling the potential for interactive applications.

\begin{figure}[ht]
  \centering
  \includegraphics[height=4.6cm]{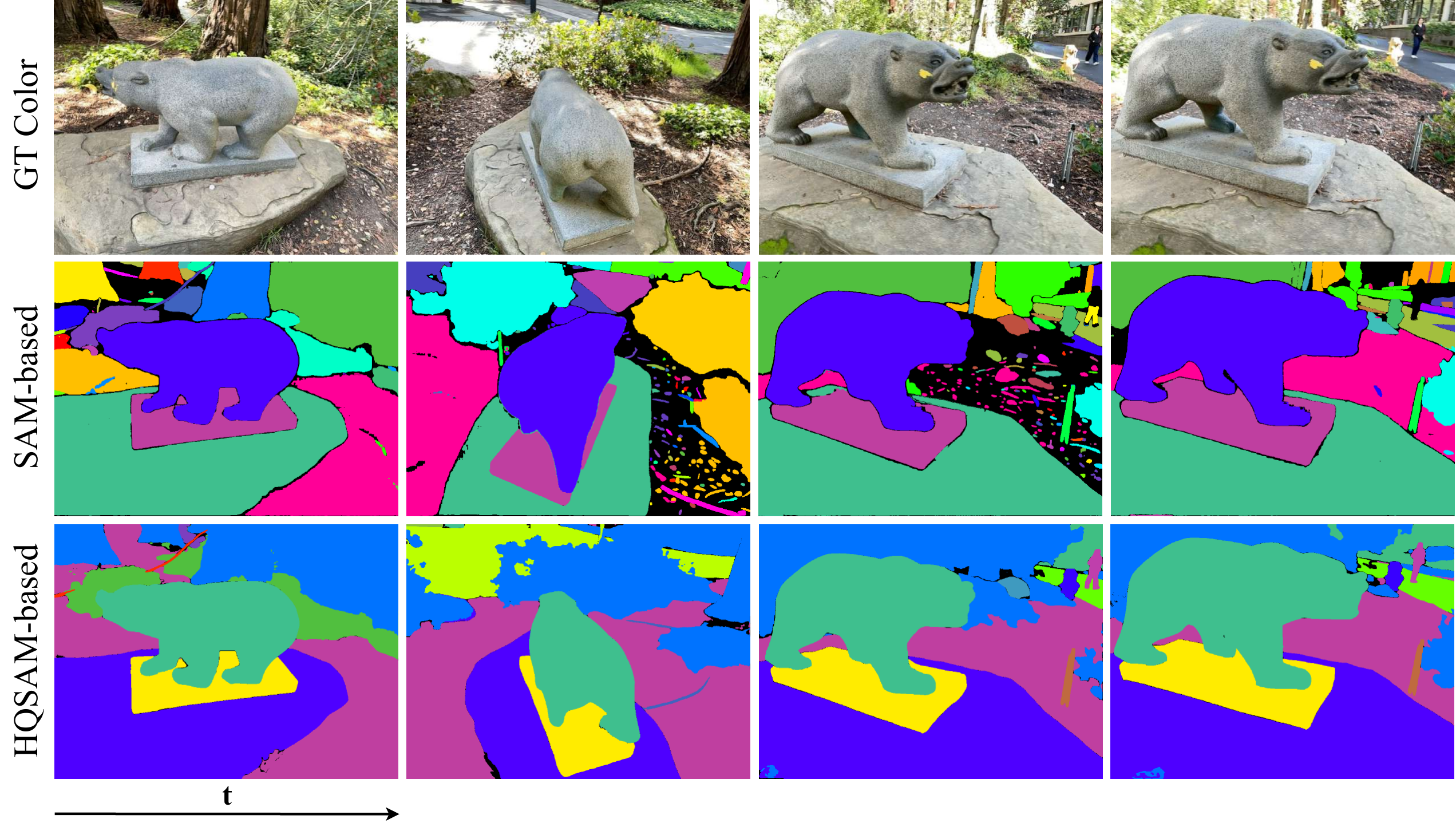}
  \caption{\textbf{Ablation on 2D Segmentation Foundation Models between SAM~\cite{segmentAnything} and HQSAM~\cite{hqsam} on Instruct-NeRF2NeRF~\cite{nerf2nerf} dataset.} 
 }
  \label{fig:ablation_sam_vs_hqsam_bear}
\vspace{-0mm}
\end{figure}

\noindent \textbf{Analysis of Mask Association on Corner Case (Validation of K-Means $K=2$ stability).}
In~\cref{fig:ablation_coner_case}, we visualize the rendered objects after distillation on the LERF~\cite{lerf2023} dataset. This particular scene is challenging due to its high object density and the presence of extreme bird's-eye view angles. Such conditions pose significant difficulties for foreground-background separation using our K-means-based binary clustering. As shown, the DEVA-based GauGroup~\cite{gaussiangrouping} produces noisy and inconsistent reconstructions under these settings. Such as \texttt{red chair} on the table, its thin leg can not be segmented correctly.  In contrast, our method exhibits robust performance across different viewpoints. Effective multi-view scene segmentation in such cases is crucial for accurate object removal in subsequent stages.

Nevertheless, the method also struggles when applying K-Means clustering with $K=2$. For certain objects, such as the \texttt{old camera} and the \texttt{gray pumpkin}, the algorithm incorrectly assigns them to the same object ID while attempting to separate foreground from background. Empirically, inspired by the strategy adopted in Gaga~\cite{gaga}, we add an additional parameter that retains only $50\%$ of the points in the foreground for this specific scene. This adjustment enables correct segmentation, suggesting that more effective methods or parameter choices remain to be explored.

\begin{figure}[ht]
  \centering
  \includegraphics[height=5cm]{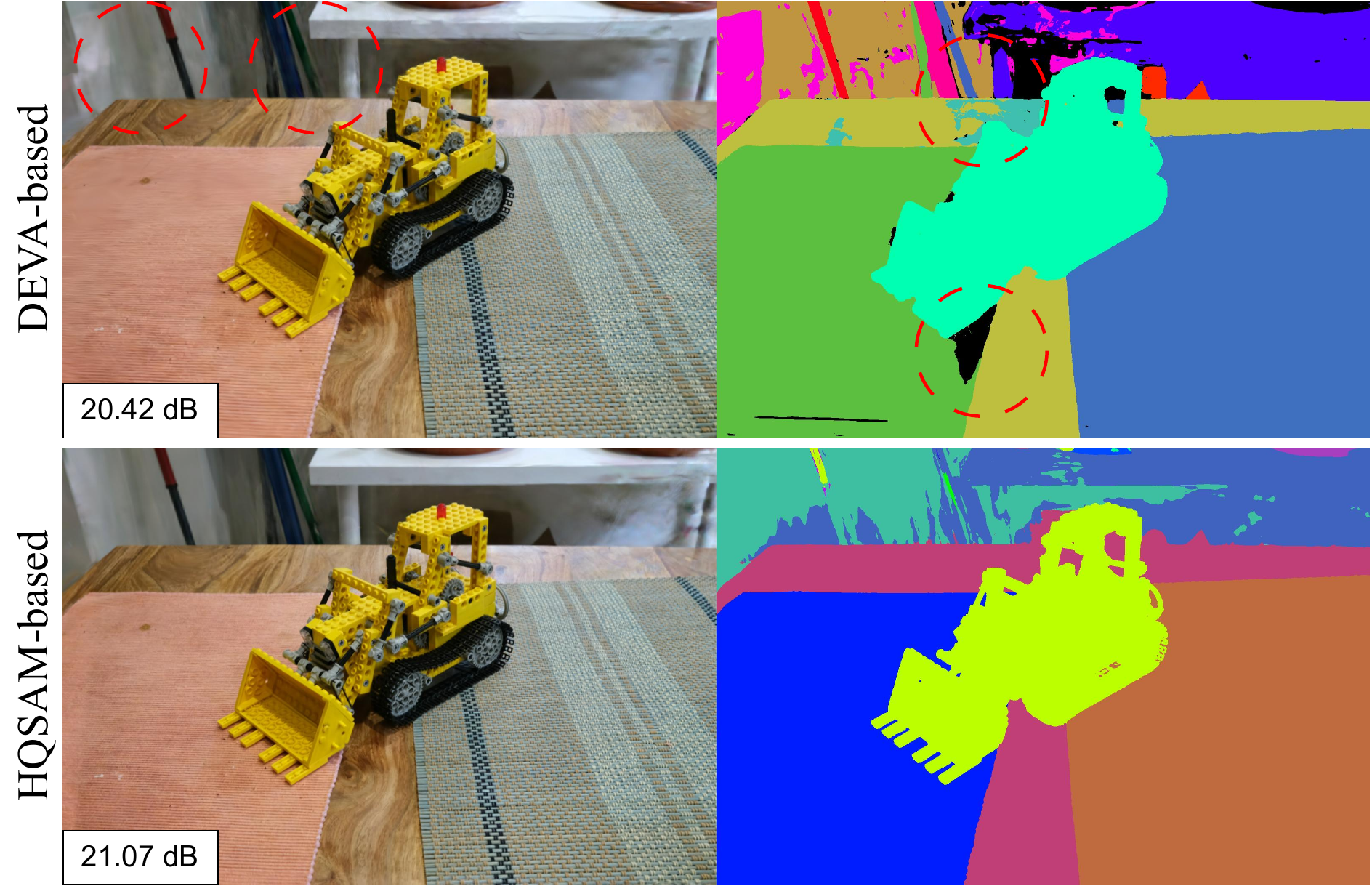}
  \caption{\textbf{Performance on Sparse View Inputs.} Our two-stage method can achieve a constantly better rendering quality(\eg, background) and segmentation result.  
 }
  \label{fig:ablation_coner_case_sparse_view}
\vspace{-0mm}
\end{figure}

\noindent \textbf{Analysis of Mask Association on Corner Case (Sparse View).}
To validate the effectiveness of our mask association under sparse-view settings, we selected 1/8 of the images (35 out of 279) from the \texttt{kitchen} scene of MipNeRF360. We compare our method against GauGroup, which is based on DEVA. The results show that our approach remains robust. We attribute this to the fact that our method performs mask association directly in the 3D point cloud, whereas DEVA treats the problem as a video signal, which introduces significant challenges. As illustrated in ~\cref{fig:ablation_coner_case_sparse_view}, our method achieves superior rendering quality and, moreover, provides more consistent and unified segmentation results.

\begin{figure}[ht]
  \centering
  \includegraphics[height=4.6cm]{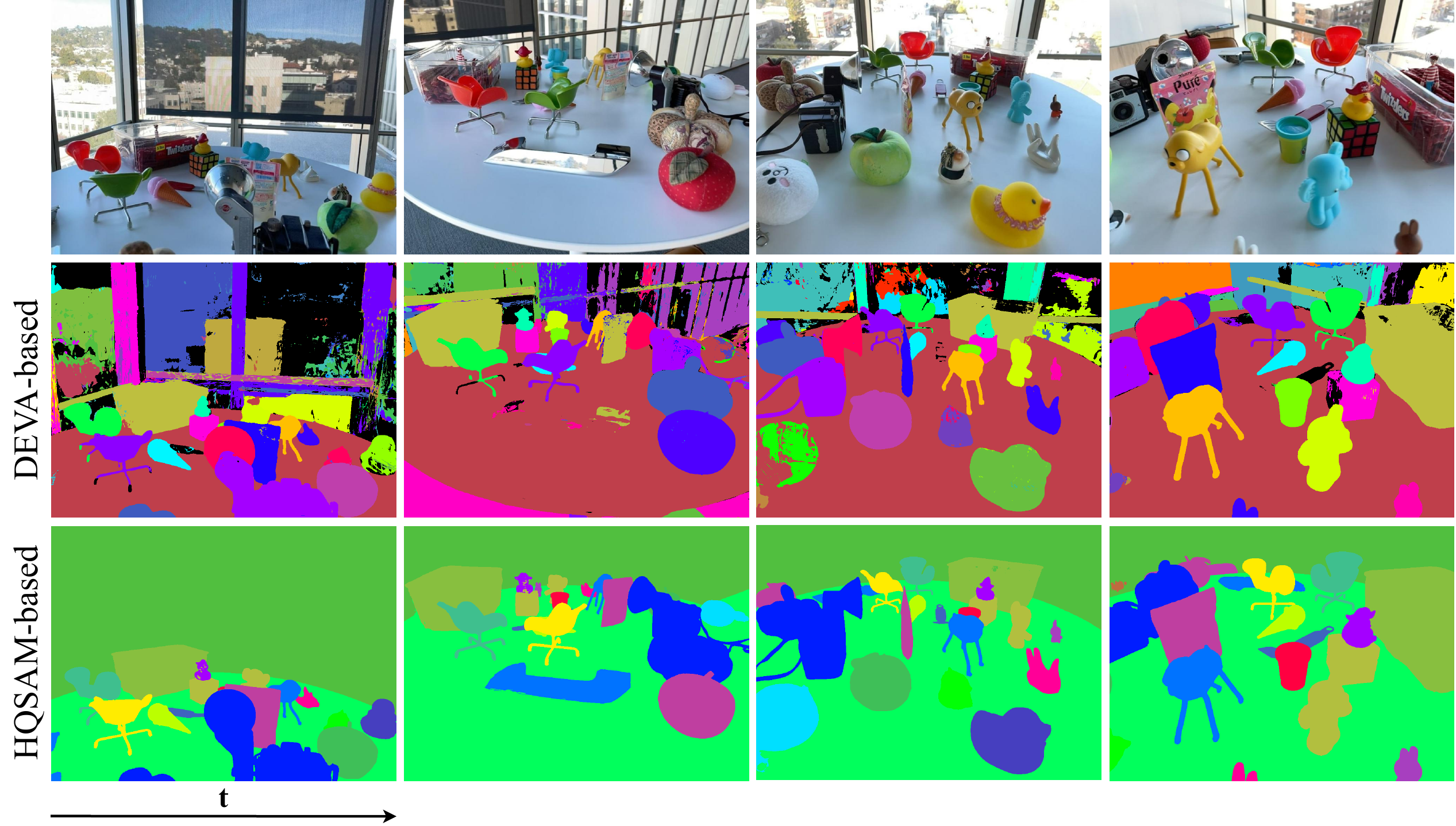}
  \caption{\textbf{Performance on Corner Case in the LERF~\cite{lerf2023} Dataset.} 
 }
  \label{fig:ablation_coner_case}
\vspace{-0mm}
\end{figure}

\noindent \textbf{Ablation on 2D Inpainting Model.}
Many recent methods introduce diffusion models for 2D inpainting~\cite{leftrefill,diffussionInpainting}. However, these models often produce visually plausible but semantically uncontrollable textures. Achieving view-consistent textures across multiple perspectives becomes particularly challenging. As a result, approaches like AuraFusion360~\cite{aurafusion360} and ImFusion~\cite{imfine} require extensive per-scene finetuning to enforce multi-view consistency.
In~\cref{fig:ablation_2DInpainter}, we compare LaMa~\cite{lama} and LeftRefill~\cite{leftrefill}. While diffusion-based methods show high-quality results, our choice of LaMa offers a more efficient alternative, aligning with our emphasis on practical and scalable scene reconstruction. Exploring diffusion models with lightweight finetuning remains a promising future direction.
\begin{figure}[tb]
  \centering
  \includegraphics[height=3.5cm]{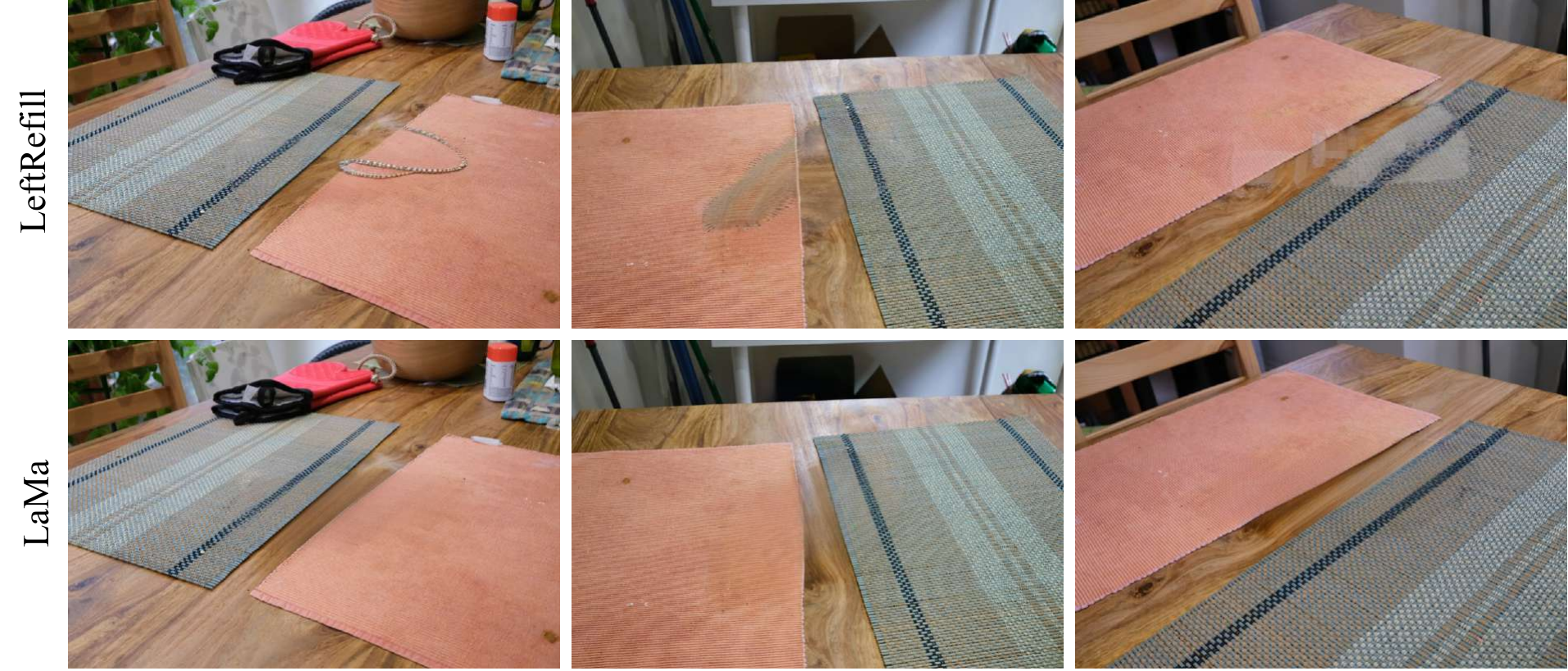}
  \caption{\textbf{Ablation on 2D Inpainting Model.} 
 }
  \label{fig:ablation_2DInpainter}
\vspace{-4mm}
\end{figure}

\noindent \textbf{Ablation on Number of Virtual Camera Views.}
In~\cref{tab:ablation_number_of_virtual_views}, we investigate how the number of virtual camera views affects the inpainting performance in terms of FID. We report results under two settings: (1) using only the constrained training views, and (2) using virtual views without conditional previous-frame guidance. The performance curves show that our model converges when approximately 30 virtual views are used, demonstrating the effectiveness and sufficiency of our view sampling strategy.

\begin{figure*}[htbp]
  \centering
  \includegraphics[height=3.5cm]{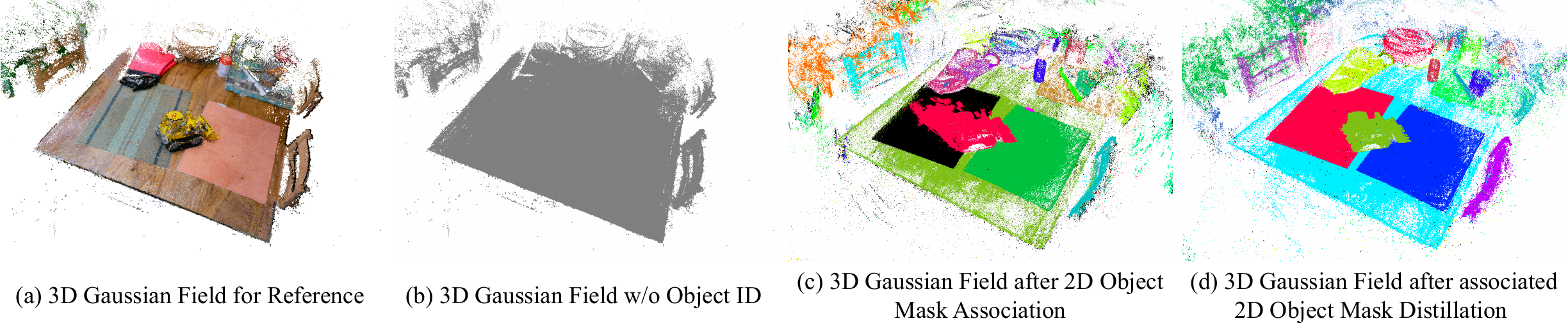}
  \caption{\textbf{Visualization of Point Cloud with/without Object IDs Information} on \texttt{kitchen} scene~\cite{mipnerf360}. After obtaining the pure Gaussian field through a standard 3D reconstruction process (a), we leverage mask association to generate (c), a raw and noisy point cloud with initial identity labels. Through identity distillation, we finally obtain (d), where consistent 2D identities are embedded into the 3D Gaussian field.} 
  \label{fig:mask_association_ply}
\vspace{-0mm}
\end{figure*}

\begin{table}[ht]
\centering
\begin{tikzpicture}
    \begin{axis}[
        width=8cm,
        height=5cm,
        xlabel={Number of Virtual Camera Views},
        ylabel={FID},
        symbolic x coords={1, 3, 10, 15, 20, 30, 40, 50, 60, 70, 80, 90,100},
        xtick=data,
        grid=both,
        legend pos=north west,
        ymin=30, ymax=80,
        title style={font=\scriptsize},
        xlabel style={font=\scriptsize},
        ylabel style={yshift=-4mm, font=\scriptsize},
        legend style={font=\fontsize{5}{6}\selectfont, inner sep=0.1pt, outer sep=0.1pt},
        tick label style={font=\tiny},
    ]
    
    \addplot[color=green, mark=triangle*] coordinates {
        (1, 75.32)
        (3, 51.32)
        (10,46.75)
        (20,38.79)
        (30,35.93)
        (40,34.77)
        (50,34.86)
        (60,35.25)
        (70,34.51)
        (80,35.27)
        (90,35.71)
        (100,35.59)
    };
    \addlegendentry{Virtual Camera View}

    \addplot[color=black, mark=square*] coordinates {
        (30, 42.23)
    };
    \addlegendentry{Virtual Camera View (w/o cond. inp.)}

    \addplot[color=red, mark=*] coordinates {
        (60, 39.51)
    };
    \addlegendentry{Known Training Views}

    \end{axis}
\end{tikzpicture}
\caption{Impact of the Number of Virtual Camera Views on FID.}
\label{tab:ablation_number_of_virtual_views}
\end{table}

\noindent \textbf{Hyperparameter Selection for the Perceptual Loss.} In \cref{tab:lpips_psnr}, we demonstrate the impact of varying LPIPS (perceptual loss) weights on the FID of our reconstructed views.
\begin{table}[ht]
\centering
\begin{tikzpicture}
    \begin{axis}[
        width=8cm,
        height=5cm,
        xlabel={Weight of LPIPS},
        ylabel={FID},
        symbolic x coords={0,0.0001,0.0005,0.001,0.005,0.01,0.1,0.5,1.0},
        xtick=data,
        grid=both,
        legend pos=north west,
        ymin=30, ymax=60,
        title style={font=\scriptsize},
        xlabel style={font=\scriptsize},
        ylabel style={yshift=-4mm, font=\scriptsize},
        legend style={font=\fontsize{5}{6}\selectfont, inner sep=0.1pt, outer sep=0.1pt},
        tick label style={font=\tiny},
    ]
    
    \addplot[color=red, mark=triangle*] coordinates {
        (0,31.8598122)
        (0.0001,31.3892382)
        (0.0005,30.2979380)
        (0.001,31.5503051)
        (0.005,31.7782560)
        (0.01,31.55291093)
        (0.1,31.3247146)
        (0.5,33.32)
        (1.0,35.9869566)
    };
    \addlegendentry{bag}

    \addplot[color=green, mark=square*] coordinates {
        (0,33.44)
        (0.0001,33.2934328)
        (0.0005,31.5004732)
        (0.001,32.9742582)
        (0.005,32.6094461)
        (0.01,33.3548234)
        (0.1,44.4118076)
        (0.5,50.37919)
        (1.0,54.1161291)
    };
    \addlegendentry{toys}

    \addplot[color=blue, mark=o] coordinates {
        (0,37.52)
        (0.0001,37.1404143)
        (0.0005,37.6573919)
        (0.001,38.8632408)
        (0.005,36.9231354)
        (0.01,38.9406797)
        (0.1,38.1166222)
        (0.5,37.680723)
        (1.0,36.6788298)
    };
    \addlegendentry{cube}

    \addplot[color=black, mark=*] coordinates {
        (0,45.9448896)
        (0.0001,45.6065446)
        (0.0005,45.3287147)
        (0.001,46.4322905)
        (0.005,44.3785523)
        (0.01,44.8914710)
        (0.1,44.2554392)
        (0.5,42.98098)
        (1.0,43.504636)
    };
    \addlegendentry{truck}

    \end{axis}
\end{tikzpicture}
\caption{Impact of Weight of LPIPS on FID.}
\label{tab:lpips_psnr}
\end{table}

\noindent \textbf{Analysis of the Gaussian ID Distillation Process.}  
In~\cref{fig:mask_association_ply}, we visualize the process of distilling object IDs into the Gaussian field. Our pipeline begins with a reconstructed pure Gaussian field (a). We first initialize per-Gaussian object ID features to obtain (b). After performing mask association, we obtain (c), a Gaussian field with raw object identity labels. However, the mask association process is primarily used to generate view-consistent segmentation masks across frames. Finally, through our 3D distillation process, the refined and consistent object identities are embedded into the Gaussian field, as shown in (d).

\begin{figure}[tb]
  \centering
  \includegraphics[height=3.7cm]{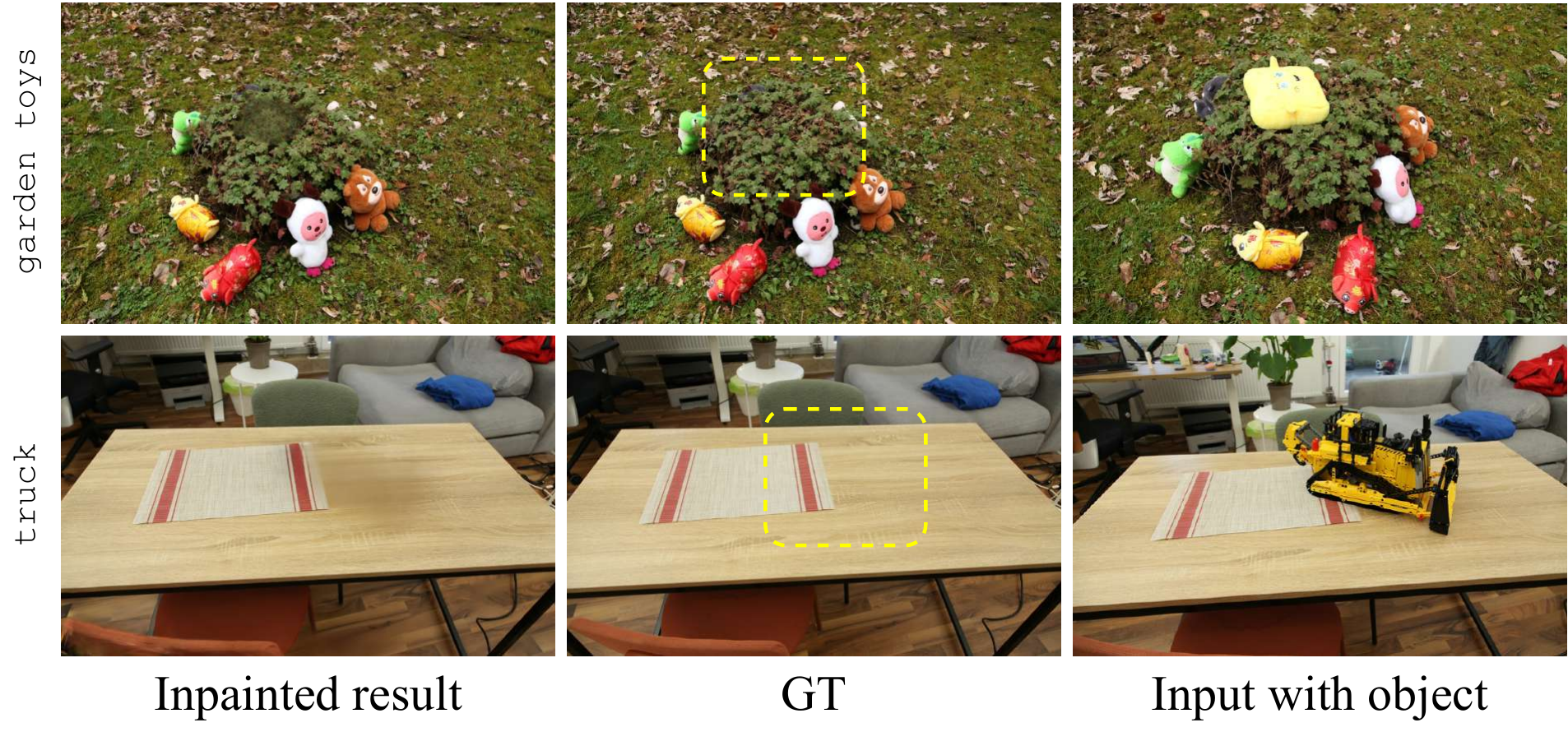}
  \caption{\textbf{Inpainting Failure Case.} 
 }
  \label{fig:inpainting_failure_case}
\vspace{-0mm}
\end{figure}

\noindent \textbf{Discussion on limitation and feature work.} 
The main limitation of our work lies in the inpainting stage after object removal, as illustrated in ~\cref{fig:inpainting_failure_case}.
First, our method is not able to properly handle shadows cast by removed objects. Second, to balance computational efficiency with the need for producing reasonable and controllable results, we employ LaMa as the 2D inpainter. However, this choice limits the inpainting quality in scenes with complex textures, where LaMa often fails to reconstruct fine-grained details. Diffusion-based methods,\eg, AuraFusion360~\cite{aurafusion360}, can generate complex textures from a single view but struggle to ensure consistency across multiple views, and their refinement typically requires long inference times.

\section{Per-Scene Breakdown of the Results.}
\label{sec:per_scene_resuls}
In~\cref{fig:multiview_kitchen} to~\cref{fig:multiview_truck}, we provide detailed multi-view comparisons across different scenes, and the per-scene quantitative results in~\cref{tab:quantitativePerSceneEva} further confirm the robustness and consistency of our method. However, floaters can still be observed in NBS regions from certain viewpoints (see our video), which mainly stem from inconsistencies in the inpainting results across views. This highlights the need for developing more consistent and efficient inpainters in future work.

When comparing to baselines, we observe that GScream~\cite{gscream} achieves stronger performance than SPIn-NeRF~\cite{spinnerf} on full-image metrics, but performs worse in the masked regions because it cannot reliably remove target objects. In contrast, SPIn-NeRF demonstrates better handling of object removal, which results in improved performance on masked-area evaluations.

In addition, to further validate its applicability in forward-facing scenarios, we compare our approach with GauGroup on SPIn-NeRF~\cite{spinnerf} dataset. Our method still delivers superior results, highlighting its scalability and generalization ability beyond $360^\circ$ settings.

\begin{table*}[ht]
\vspace{0mm}
% \normalsize
% \footnotesize
\scriptsize
% *\tiny
\centering
\renewcommand{\arraystretch}{1.3}
\setlength{\tabcolsep}{3pt}  
\begin{tabular}{c l |  c @{\hspace{10pt}} c @{\hspace{10pt}} c @{\hspace{10pt}} c @{\hspace{10pt}} c @{\hspace{10pt}} c @{\hspace{10pt}} c @{\hspace{10pt}}  c }
\hline 

Scene & Methods              & PSNR $\uparrow$ &masked PSNR $\uparrow$ & SSIM $\uparrow$ & masked SSIM$\uparrow$ & LPIPS $\downarrow$ & masked LPIPS $\downarrow$ & FID $\downarrow$   \\
\hline  
\multirow{5}{*}{\makecell{\rotatebox[origin=c]{90}{\texttt{fruits}}}} 
 & SPIn-NeRF \cite{spinnerf}          & 11.15 &34.21 &0.4617 &0.9963 &0.6253 &0.0056 &367.19 \\
 & GScream \cite{gscream}             & \cellcolor{tabthird}23.39 & 31.03 & \cellcolor{tabthird}0.8506 & 0.9934 & 0.2559 & 0.0091 & 80.17   \\
 & AuraFusion \cite{aurafusion360}    & \cellcolor{tabsecond}23.93 & \cellcolor{tabsecond}37.38 & \cellcolor{tabsecond}0.8617 & \cellcolor{tabsecond}0.9975 & \cellcolor{tabthird}0.2450 & \cellcolor{tabsecond}0.0042 & \cellcolor{tabthird}61.94   \\
 & GauGroup \cite{gaussiangrouping}   & 22.55 & \cellcolor{tabthird}35.92 & 0.8485 & \cellcolor{tabthird}0.9973 & \cellcolor{tabsecond}0.2365 & \cellcolor{tabthird}0.0043 & \cellcolor{tabsecond}60.13 \\
 & Inpaint360GS (Ours)                & \cellcolor{tabfirst}27.38	&\cellcolor{tabfirst}44.54	&\cellcolor{tabfirst}0.9014	&\cellcolor{tabfirst}0.9993	&\cellcolor{tabfirst}0.1657	&\cellcolor{tabfirst}0.0011	&\cellcolor{tabfirst}30.33  \\
\hline 
\multirow{5}{*}{\makecell{\rotatebox[origin=c]{90}{\texttt{doppelherz}}}} 
 & SPIn-NeRF \cite{spinnerf}          &21.24 &41.66 &0.5421 &0.9986 &0.5227 &0.0031 &258.82   \\
 & GScream \cite{gscream}             & 24.56 & 38.73 & 0.8108 & 0.9958 & 0.1849 & 0.0030 & 88.09   \\
 & AuraFusion \cite{aurafusion360}    & \cellcolor{tabsecond}27.81 & \cellcolor{tabsecond}44.39 & \cellcolor{tabthird}0.8545 & \cellcolor{tabthird}0.9989 & \cellcolor{tabthird}0.1379 & \cellcolor{tabthird}0.0014 & \cellcolor{tabsecond}32.56   \\
 & GauGroup \cite{gaussiangrouping}   & \cellcolor{tabthird}27.40 & \cellcolor{tabthird}43.69 & \cellcolor{tabsecond}0.8787 & \cellcolor{tabsecond}0.9991 & \cellcolor{tabsecond}0.1096 & \cellcolor{tabsecond}0.0013 & \cellcolor{tabthird}44.90   \\
 & Inpaint360GS (Ours)                &\cellcolor{tabfirst}29.2	& \cellcolor{tabfirst}46	&\cellcolor{tabfirst}0.9129	&\cellcolor{tabfirst}0.9994	&\cellcolor{tabfirst}0.0789	&\cellcolor{tabfirst}0.0009	&\cellcolor{tabfirst}20.13  \\
\hline  
\multirow{5}{*}{\makecell{\rotatebox[origin=c]{90}{\texttt{toys}}}} 
 & SPIn-NeRF \cite{spinnerf}          & \cellcolor{tabthird}25.97 &\cellcolor{tabthird}39.79 &0.6558 &\cellcolor{tabthird}0.9919 &0.3785 &0.0086 &119.03 \\
 & GScream \cite{gscream}             & 25.27 & 31.68 & 0.8164 & 0.9865 & 0.1860 & 0.0138 & 376.61   \\
 & AuraFusion \cite{aurafusion360}    & \cellcolor{tabsecond}27.05 & \cellcolor{tabsecond}39.94 & \cellcolor{tabsecond}0.8011 & \cellcolor{tabsecond}0.9917 & \cellcolor{tabthird}0.1996 & \cellcolor{tabthird}0.0073 & \cellcolor{tabsecond}41.03   \\
 & GauGroup \cite{gaussiangrouping}   & 24.08 & 34.90 & \cellcolor{tabthird}0.7683 & 0.9886 & \cellcolor{tabsecond}0.1796 & \cellcolor{tabsecond}0.0065 & \cellcolor{tabthird}64.97  \\
 & Inpaint360GS (Ours)                & \cellcolor{tabfirst}28.14	&\cellcolor{tabfirst}40.58	&\cellcolor{tabfirst}0.8707	&\cellcolor{tabfirst}0.9928	&\cellcolor{tabfirst}0.0995	&\cellcolor{tabfirst}0.0053	&\cellcolor{tabfirst}33.29  \\
\hline  
\multirow{5}{*}{\makecell{\rotatebox[origin=c]{90}{\texttt{garden toys}}}} 
 & SPIn-NeRF \cite{spinnerf}          & 21.79 &33.57 &0.5730 &0.9855 &0.3778 &0.0134 &116.17  \\
 & GScream \cite{gscream}             & 21.01 & 28.60 & 0.7066 & 0.9841 & 0.2358 & 0.0130 & 130.18   \\
 & AuraFusion \cite{aurafusion360}    & \cellcolor{tabthird}21.34 & \cellcolor{tabthird}30.49 & \cellcolor{tabthird}0.7147 & \cellcolor{tabthird}0.9834 & \cellcolor{tabthird}0.2372 & \cellcolor{tabthird}0.0134 & \cellcolor{tabthird}64.41   \\
 & GauGroup \cite{gaussiangrouping}   & \cellcolor{tabsecond}22.41 & \cellcolor{tabsecond}33.56 & \cellcolor{tabsecond}0.7585 & \cellcolor{tabsecond}0.9850 & \cellcolor{tabsecond}0.1590 & \cellcolor{tabsecond}0.0103 & \cellcolor{tabsecond}48.70 \\
 & Inpaint360GS (Ours)                & \cellcolor{tabfirst}23.68	&\cellcolor{tabfirst}33.71	&\cellcolor{tabfirst}0.8094	&\cellcolor{tabfirst}0.9857	&\cellcolor{tabfirst}0.1228	&\cellcolor{tabfirst}0.0098	&\cellcolor{tabfirst}30.58  \\
\hline  
 \multirow{5}{*}{\makecell{\rotatebox[origin=c]{90}{\texttt{bag}}}} 
 & SPIn-NeRF \cite{spinnerf}          &23.08 &\cellcolor{tabthird}34.39 &0.5278 &0.9872 &0.4728 &0.0076 &124.15 \\
 & GScream \cite{gscream}             & 24.84 & 32.52 & 0.7913 & 0.9827 & 0.2264 & 0.0124 & 187.60   \\
 & AuraFusion \cite{aurafusion360}    & \cellcolor{tabsecond}26.46 & 34.22 & \cellcolor{tabthird}0.8211 & \cellcolor{tabthird}0.9861 & \cellcolor{tabthird}0.2056 & \cellcolor{tabthird}0.011 & \cellcolor{tabthird}55.12   \\
 & GauGroup \cite{gaussiangrouping}   & \cellcolor{tabthird}26.28 & \cellcolor{tabsecond}35.04 & \cellcolor{tabsecond}0.827 & \cellcolor{tabsecond}0.9874 & \cellcolor{tabsecond}0.1586 & \cellcolor{tabsecond}0.0062 & \cellcolor{tabsecond}33.74 \\
 & Inpaint360GS (Ours)                & \cellcolor{tabfirst}27.97 &\cellcolor{tabfirst}37.45	&\cellcolor{tabfirst}0.8627	&\cellcolor{tabfirst}0.9887	&\cellcolor{tabfirst}0.1263 &\cellcolor{tabfirst}0.0056	&\cellcolor{tabfirst}31.41\\
\hline               
\multirow{5}{*}{\makecell{\rotatebox[origin=c]{90}{\texttt{car}}}} 
 & SPIn-NeRF \cite{spinnerf}          & 19.15 & 22.12 & 0.3901 & 0.9456 & 0.5541 & 0.0485 & 334.78 \\
 & GScream \cite{gscream}             & \cellcolor{tabthird}19.35 & 23.02 & 0.7015 & 0.9474 & 0.2741 & 0.0413 & 324.76   \\
 & AuraFusion \cite{aurafusion360}    & \cellcolor{tabfirst}21.22 & \cellcolor{tabsecond}26.01 & \cellcolor{tabfirst}0.7718 & \cellcolor{tabfirst}0.9524 & \cellcolor{tabfirst}0.1769 & \cellcolor{tabfirst}0.0283 & \cellcolor{tabfirst}67.82  \\
 & GauGroup \cite{gaussiangrouping}   & 18.43 & \cellcolor{tabthird}24.65 & \cellcolor{tabthird}0.6516 & \cellcolor{tabthird}0.9468 & \cellcolor{tabthird}0.2609 & \cellcolor{tabthird}0.0388 & \cellcolor{tabthird}157.23  \\
 & Inpaint360GS (Ours)                & \cellcolor{tabsecond}20.71 & \cellcolor{tabfirst}27.96 & \cellcolor{tabsecond}0.7309 & \cellcolor{tabsecond}0.9475	&\cellcolor{tabsecond}0.1943	& \cellcolor{tabsecond}0.0357	 & \cellcolor{tabsecond}88.95\\
\hline  
\multirow{5}{*}{\makecell{\rotatebox[origin=c]{90}{\texttt{red cone}}}} 
 & SPIn-NeRF \cite{spinnerf}          &18.71 &32.04 &0.3572 &0.9929 &0.5177 &0.0094 &127.88 \\  
 & GScream \cite{gscream}             &19.31 & 30.53 & 0.6970 & 0.9866 & 0.2528 & 0.0121 & 84.36   \\
 & AuraFusion \cite{aurafusion360}    & \cellcolor{tabthird}20.55 & \cellcolor{tabthird}36.14 & \cellcolor{tabthird}0.7526 & \cellcolor{tabsecond}0.9927 & \cellcolor{tabthird}0.1967 & \cellcolor{tabthird}0.0077 & \cellcolor{tabthird}31.02  \\
 & GauGroup \cite{gaussiangrouping}   & \cellcolor{tabsecond}21.14 & \cellcolor{tabsecond}37.44 & \cellcolor{tabsecond}0.7744 & \cellcolor{tabthird}0.9914 & \cellcolor{tabsecond}0.1346 & \cellcolor{tabsecond}0.0053 & \cellcolor{tabfirst}19.97  \\
 & Inpaint360GS (Ours)                & \cellcolor{tabfirst}21.45	&\cellcolor{tabfirst}38.83	&\cellcolor{tabfirst}0.7973	&\cellcolor{tabfirst}0.9933	&\cellcolor{tabfirst}0.1201	&\cellcolor{tabfirst}0.0051	&\cellcolor{tabsecond}21.42  \\
\hline  
\multirow{5}{*}{\makecell{\rotatebox[origin=c]{90}{\texttt{yellow cone}}}} 
 & SPIn-NeRF \cite{spinnerf}          &17.92 &36.09 &0.3130 &0.9893 &0.6374 &0.0087 &379.17 \\  
 & GScream \cite{gscream}             & 24.77 & 33.21 & 0.8124 & 0.9880 & 0.1775 & 0.0089 & 140.88   \\
 & AuraFusion \cite{aurafusion360}    & \cellcolor{tabthird}25.90 & \cellcolor{tabthird}39.06 & \cellcolor{tabthird}0.8195 & \cellcolor{tabthird}0.9912 & \cellcolor{tabthird}0.1590 & \cellcolor{tabthird}0.0049 & \cellcolor{tabthird}35.78   \\
 & GauGroup \cite{gaussiangrouping}   & \cellcolor{tabsecond}26.32 & \cellcolor{tabsecond}39.99 & \cellcolor{tabsecond}0.8480 & \cellcolor{tabsecond}0.9921 & \cellcolor{tabsecond}0.1171 & \cellcolor{tabsecond}0.0035 & \cellcolor{tabsecond}28.78  \\
 & Inpaint360GS (Ours)                & \cellcolor{tabfirst}26.33 &\cellcolor{tabfirst}42.51	&\cellcolor{tabfirst}0.8642	&\cellcolor{tabfirst}0.9926	&\cellcolor{tabfirst}0.0935	&\cellcolor{tabfirst}0.0039	&\cellcolor{tabfirst}21.38  \\
\hline  
\multirow{5}{*}{\makecell{\rotatebox[origin=c]{90}{\texttt{cube}}}} 
 & SPIn-NeRF \cite{spinnerf}          &17.52 &27.32 &0.6621 &0.9708 &0.4315 &0.0279 &351.46   \\
 & GScream \cite{gscream}             & 15.32 & 22.09 & 0.6596 & 0.9703 & 0.4321 & 0.0290 & 396.07   \\
 & AuraFusion \cite{aurafusion360}    & \cellcolor{tabsecond}22.48 & \cellcolor{tabsecond}27.82 & \cellcolor{tabsecond}0.8645 &\cellcolor{tabsecond} 0.9807 &\cellcolor{tabsecond} 0.1506 & \cellcolor{tabsecond}0.0118 & \cellcolor{tabsecond} 43.24  \\
 & GauGroup \cite{gaussiangrouping}   & \cellcolor{tabthird}20.10 & \cellcolor{tabthird}27.51& \cellcolor{tabthird}0.8127 & \cellcolor{tabthird}0.9749 & \cellcolor{tabthird}0.2071 & \cellcolor{tabthird}0.0197 & \cellcolor{tabthird}118.93   \\
 & Inpaint360GS (Ours)                &\cellcolor{tabfirst}22.52	&\cellcolor{tabfirst}28.58	&\cellcolor{tabfirst}0.8879	&\cellcolor{tabfirst}0.9874	&\cellcolor{tabfirst}0.1079	&\cellcolor{tabfirst}0.0083	&\cellcolor{tabfirst}37.14  \\
\hline  
\multirow{5}{*}{\makecell{\rotatebox[origin=c]{90}{\texttt{redbull}}}} 
 & SPIn-NeRF \cite{spinnerf}          &20.98 &41.00 &0.4699 &0.9973 &0.4691 &0.0052 &186.81 \\
 & GScream \cite{gscream}             & 19.24 & 26.42 & 0.6218 & 0.9923 & 0.3637 & 0.0087 & 286.52   \\
 & AuraFusion \cite{aurafusion360}    & \cellcolor{tabsecond}23.22 & \cellcolor{tabthird}40.80 & \cellcolor{tabthird}0.7258 & \cellcolor{tabsecond}0.9982 & \cellcolor{tabthird}0.2178 & \cellcolor{tabsecond}0.0021 & \cellcolor{tabsecond}47.57   \\
 & GauGroup \cite{gaussiangrouping}   & \cellcolor{tabthird}23.06 & \cellcolor{tabsecond}41.36 & \cellcolor{tabsecond}0.7409 & \cellcolor{tabthird}0.9981 & \cellcolor{tabsecond}0.1870 & \cellcolor{tabthird}0.0025 & \cellcolor{tabthird}63.98  \\
 & Inpaint360GS (Ours)                &\cellcolor{tabfirst}23.55	&\cellcolor{tabfirst}42.62	&\cellcolor{tabfirst}0.7655	&0\cellcolor{tabfirst}.9988	&\cellcolor{tabfirst}0.1573	&\cellcolor{tabfirst}0.0014	&\cellcolor{tabfirst}34.94  \\
\hline  
\multirow{5}{*}{\makecell{\rotatebox[origin=c]{90}{\texttt{truck}}}} 
 & SPIn-NeRF \cite{spinnerf}          &24.23  & 30.54.99 &0.7604 &\cellcolor{tabsecond}0.9919 &0.3626 &0.0101 &164.01 \\
 & GScream \cite{gscream}             & 21.93 & 27.16 & 0.8458 & 0.9813 & 0.1838 & 0.0181 & 173.49  \\
 & AuraFusion \cite{aurafusion360}    & \cellcolor{tabsecond}25.51 & \cellcolor{tabsecond}31.43 & \cellcolor{tabthird}0.8763 & \cellcolor{tabthird}0.9903 & \cellcolor{tabthird}0.1675 & \cellcolor{tabsecond}0.0081 & \cellcolor{tabsecond}49.30 \\
 & GauGroup \cite{gaussiangrouping}   & \cellcolor{tabthird}23.70 & \cellcolor{tabthird}28.39 & \cellcolor{tabsecond}0.8829 & \cellcolor{tabthird}0.9898 & \cellcolor{tabsecond}0.1465 & \cellcolor{tabthird}0.0115 & \cellcolor{tabthird}83.21  \\
 & Inpaint360GS (Ours)                &\cellcolor{tabfirst}25.62	&\cellcolor{tabfirst}33.99 & \cellcolor{tabfirst}0.9172	&\cellcolor{tabfirst}0.9923	&\cellcolor{tabfirst}0.0975	&\cellcolor{tabfirst}0.0080	&\cellcolor{tabfirst}45.63  \\
\hline  
\multirow{5}{*}{\makecell{\rotatebox[origin=c]{90}{\texttt{avg.}}}} 
 & SPIn-NeRF \cite{spinnerf}          & 19.71  & 34.53    & 0.5000  & 0.9854    & 0.5002  &0.0140 & 229.95  \\
 & GScream \cite{gscream}             & 20.95  & 28.47    & 0.7380  & 0.9819    & 0.2715  & 0.0161  & 206.25  \\
 & AuraFusion360 \cite{aurafusion360}    & \cellcolor{tabthird}23.15  & \cellcolor{tabsecond}35.78    & \cellcolor{tabthird}0.7923  & \cellcolor{tabsecond}0.9872    & \cellcolor{tabthird}0.1915  & \cellcolor{tabsecond}0.0097  & \cellcolor{tabsecond}47.71  \\
 & GauGroup \cite{gaussiangrouping}   & \cellcolor{tabsecond}23.20  & \cellcolor{tabthird}35.73    & \cellcolor{tabsecond}0.7928  & \cellcolor{tabthird}0.9862    & \cellcolor{tabsecond}0.1770  & \cellcolor{tabthird}0.0102  & \cellcolor{tabthird}65.87  \\
 & Inpaint360GS (Ours)                & \cellcolor{tabfirst}24.40  & \cellcolor{tabfirst}36.29    & \cellcolor{tabfirst}0.8370  & \cellcolor{tabfirst}0.9886    & \cellcolor{tabfirst}0.1300  & \cellcolor{tabfirst}0.0078  & \cellcolor{tabfirst}35.93  \\
\hline

\end{tabular}
\vspace{-3mm}
\caption{\textbf{Per scene quantitative comparison on the Inpaint360GS dataset.}}
\label{tab:quantitativePerSceneEva}
\vspace{-1mm}
\end{table*}

\begin{figure*}[tb]
  \centering
  \includegraphics[height=16.5cm]{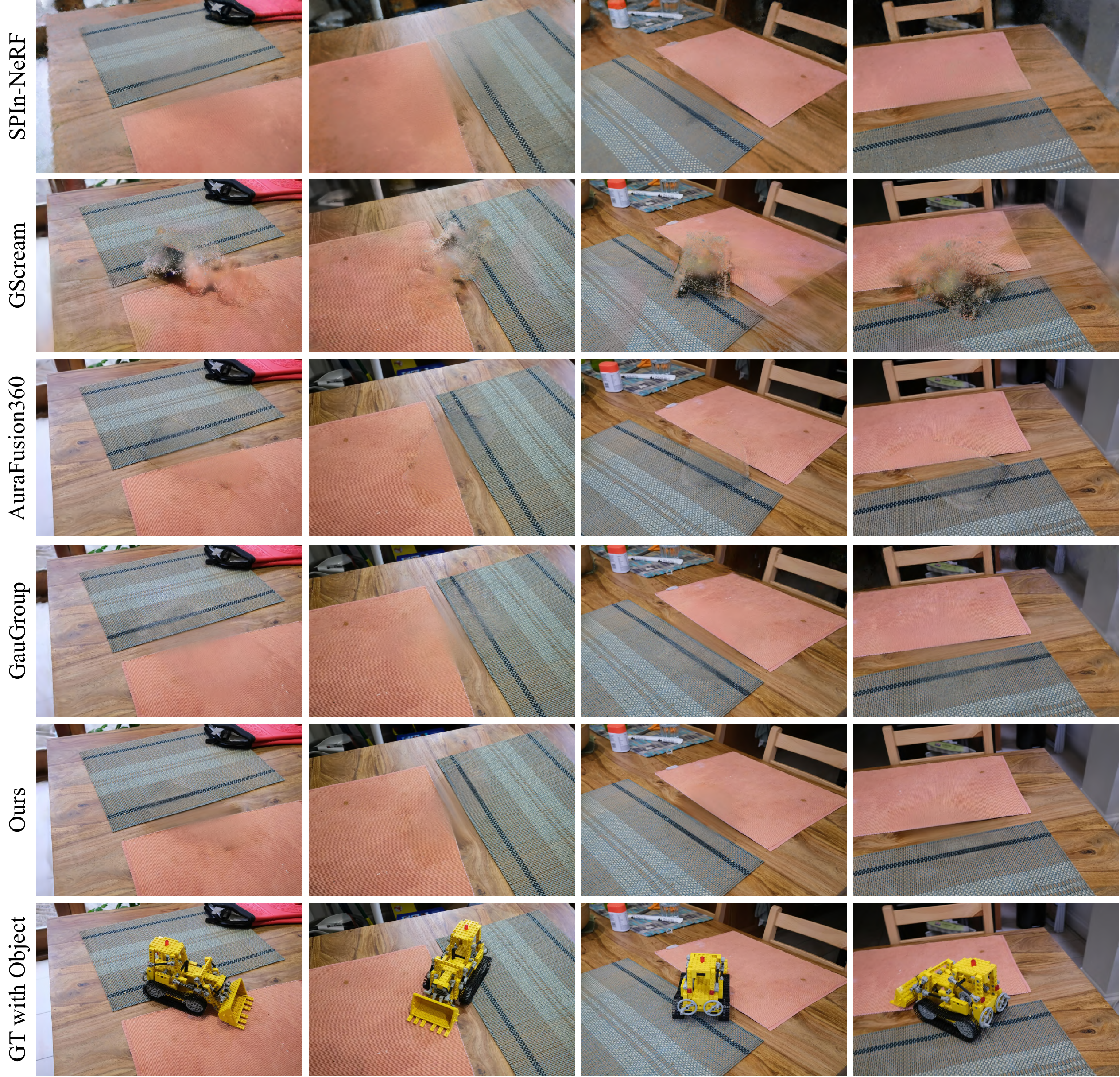}
  \caption{\textbf{Multi-view comparison on Mip-NeRF 360~\cite{mipnerf360}}~\texttt{kitchen}. We evaluate SPIn-NeRF~\cite{spinnerf}, GScream~\cite{gscream}, AuraFusion360~\cite{aurafusion360}, GauGroup~\cite{gaussiangrouping} and our method, with object-inclusive ground truth images provided for each corresponding view. Each column represents a distinct viewpoint, and four representative angles are selected to comprehensively demonstrate the performance across the full set of views. Our method achieves superior multi-view consistency with detailed texture and smooth boundary compared to the baseline approaches.
 }
  \label{fig:multiview_kitchen}
\vspace{-0mm}
\end{figure*}

\begin{figure*}[tb]
  \centering
  \includegraphics[height=18.7cm]{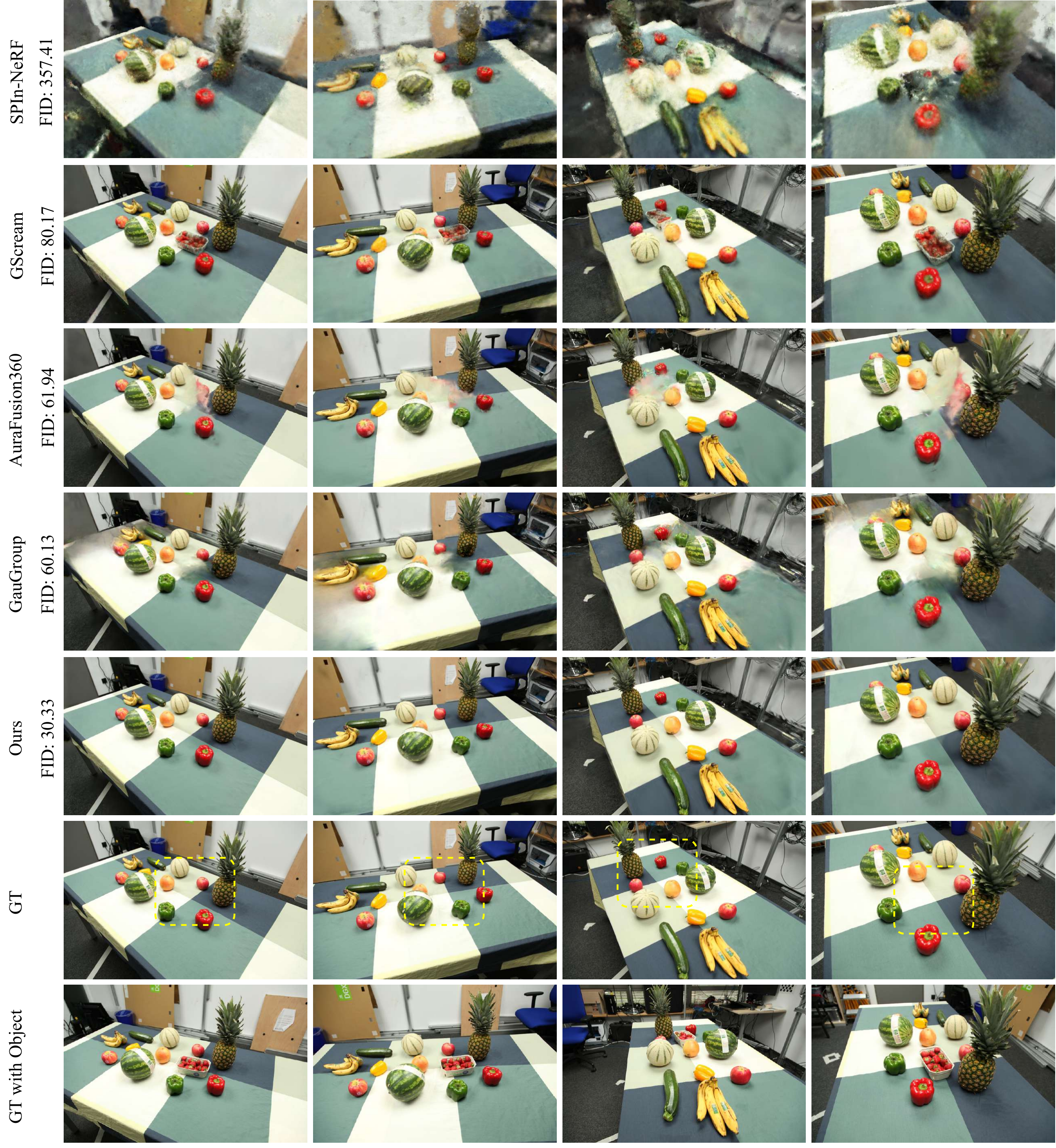}
  \caption{\textbf{Multi-view comparison on Inpaint360GS}~\texttt{fruits}. We evaluate SPIn-NeRF~\cite{spinnerf}, GScream~\cite{gscream}, AuraFusion360~\cite{aurafusion360}, GauGroup~\cite{gaussiangrouping} and our method, with object-inclusive ground truth images provided for each corresponding view.  In the scene with multiple objects, our method demonstrates a clear advantage. This can be attributed to our precise object ID assignment within the Gaussian field, which is further integrated into the virtual camera view. As a result, our method is able to identify more accurate never-been-seen (NBS) regions. We attribute the above performance gains to these key design choices.
 }
  \label{fig:multiview_fruits}
\vspace{-0mm}
\end{figure*}

\begin{figure*}[tb]
  \centering
  \includegraphics[height=18.7cm]{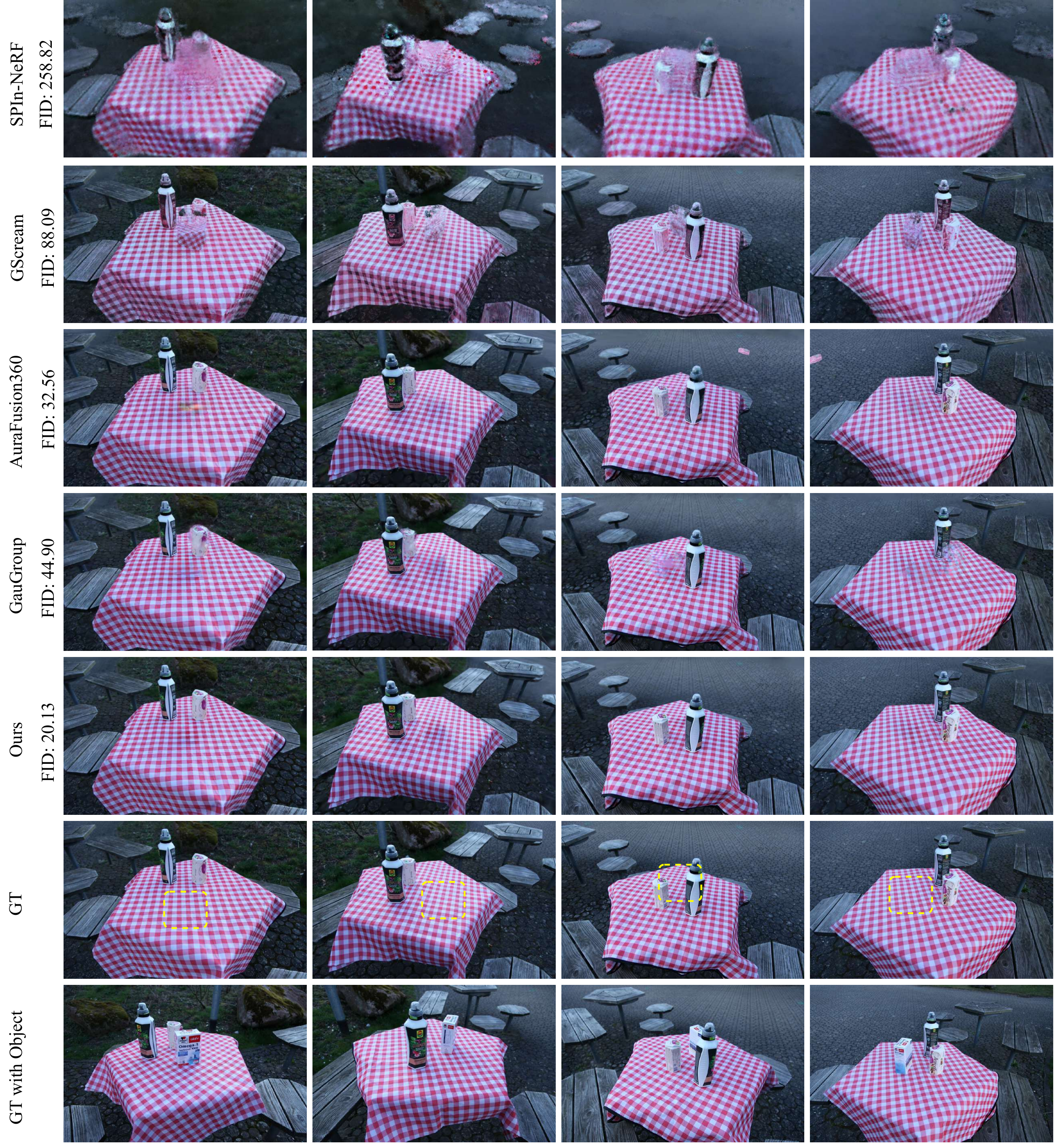}
  \caption{\textbf{Multi-view comparison on Inpaint360GS}~\texttt{doppelherz}. We evaluate SPIn-NeRF~\cite{spinnerf}, GScream~\cite{gscream}, AuraFusion360~\cite{aurafusion360}, GauGroup~\cite{gaussiangrouping} and our method, with object-inclusive ground truth images provided for each corresponding view. The scene poses significant challenges due to distant viewpoints and multiple objects, making NBS region detection unreliable. While AuraFusion360 suffers from floating textures due to poor depth alignment, our method remains robust, benefiting from the structured virtual camera trajectory that facilitates consistent and accurate NBS region identification. Our approach first removes occluding objects and then performs inpainting, enabling efficient utilization of scene information for faithful reconstruction.}
  \label{fig:multiview_doppelherz}
\vspace{-0mm}
\end{figure*}

\begin{figure*}[tb]
  \centering
  \includegraphics[height=18.7cm]{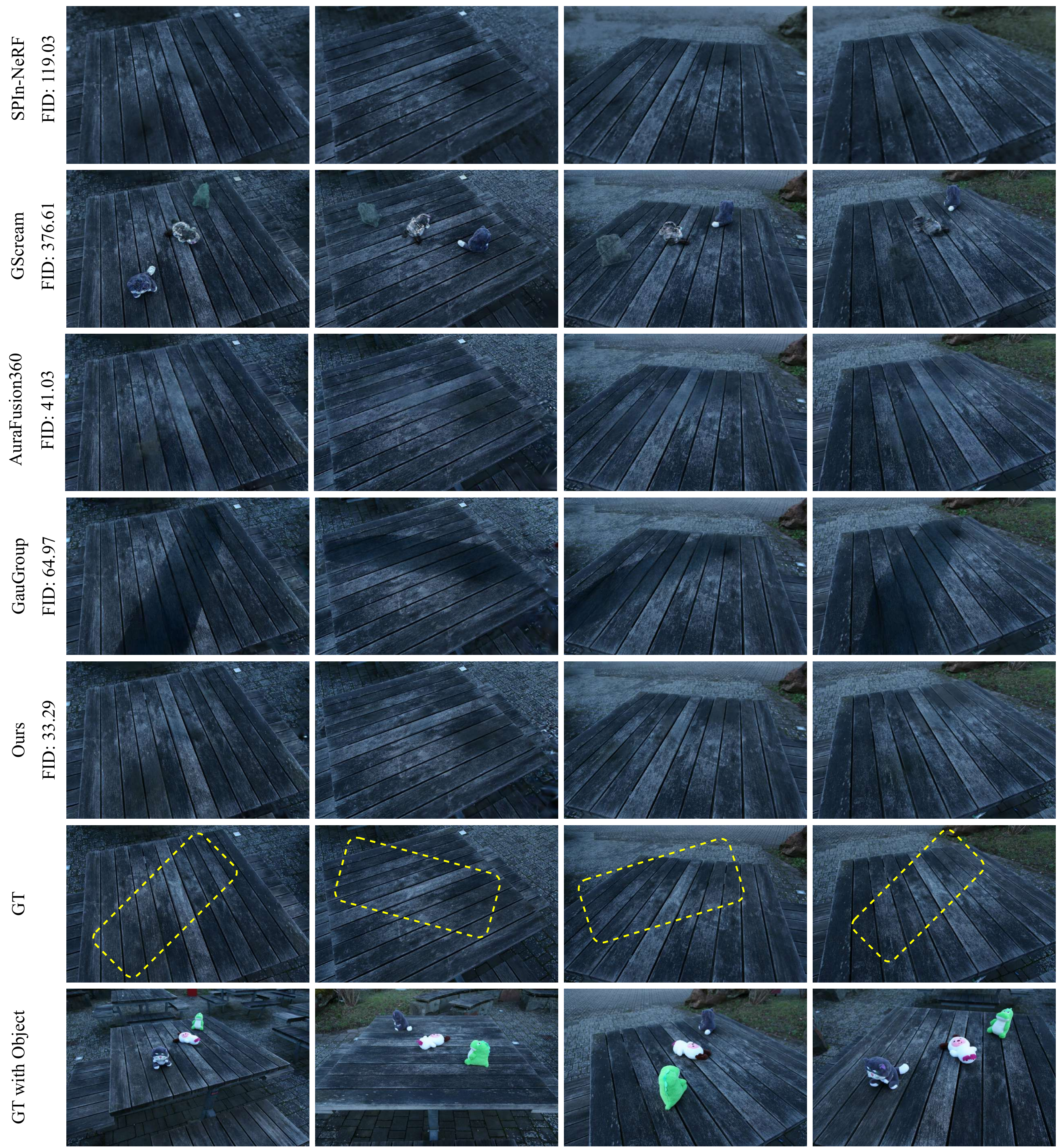}
  \caption{\textbf{Multi-view comparison on Inpaint360GS}~\texttt{toys}. We evaluate SPIn-NeRF~\cite{spinnerf}, GScream~\cite{gscream}, AuraFusion360~\cite{aurafusion360}, GauGroup~\cite{gaussiangrouping} and our method, with object-inclusive ground truth images provided for each corresponding view. This scene, though containing multiple objects, is relatively simple due to the sparse layout and lack of occlusion. Both AuraFusion360 and SPIn-NeRF demonstrate visually pleasing results under this setting. Nonetheless, our method achieves more consistent appearance across views.
 }
  \label{fig:multiview_toys}
\vspace{-0mm}
\end{figure*}

\begin{figure*}[tb]
  \centering
  \includegraphics[height=18.7cm]{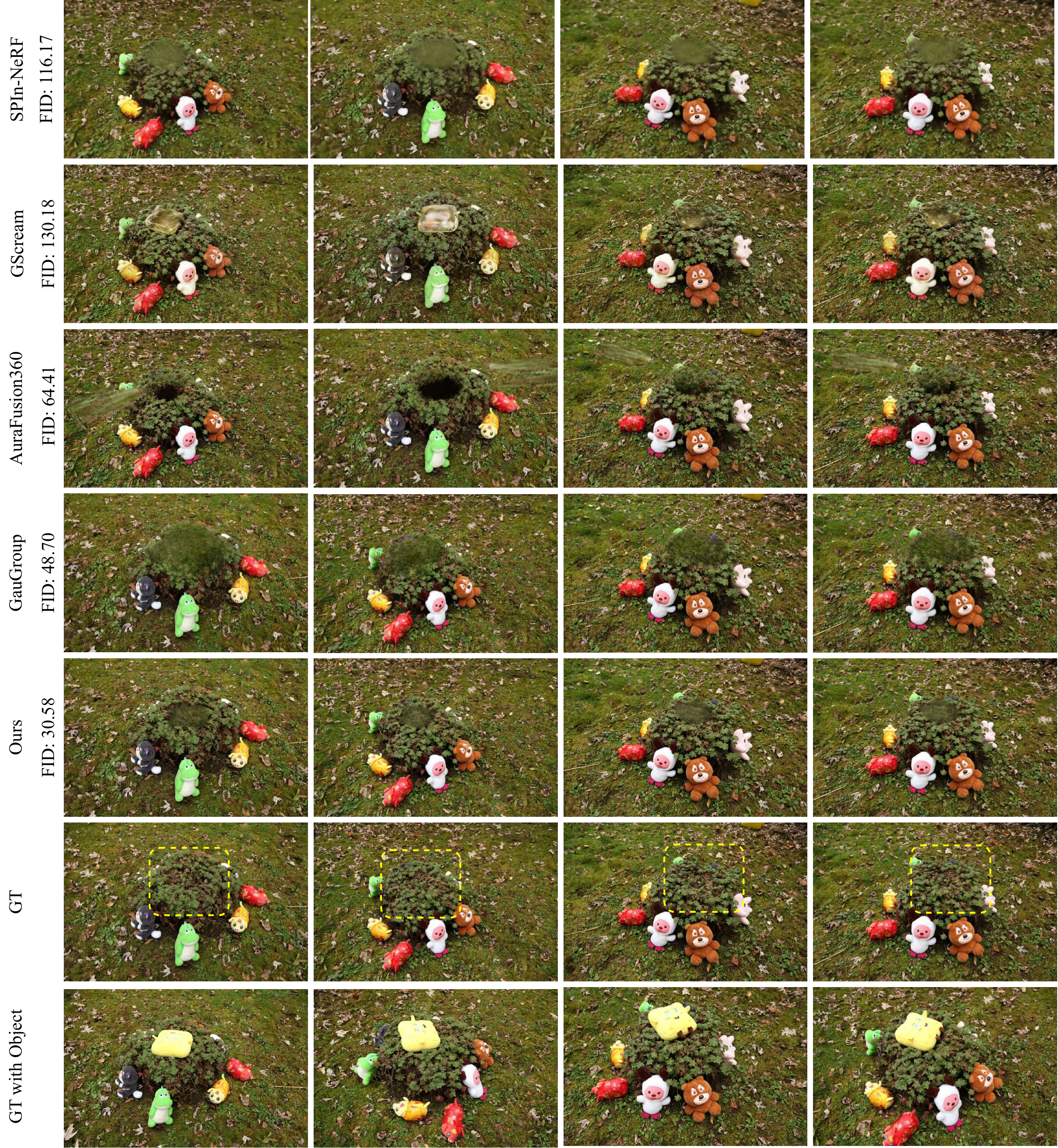}
  \caption{\textbf{Multi-view comparison on Inpaint360GS}~\texttt{garden toys}. We evaluate SPIn-NeRF~\cite{spinnerf}, GScream~\cite{gscream}, AuraFusion360~\cite{aurafusion360}, GauGroup~\cite{gaussiangrouping} and our method, with object-inclusive ground truth images provided for each corresponding view.This scene is particularly challenging due to the unpredictable NBS region and the stochastic nature of the leaf textures. Our chosen 2D inpainting model (LaMa), while efficient, lacks the generative capacity of diffusion-based models to synthesize such fine-grained details. Nevertheless, our method achieves the best overall visual quality among all baselines, despite lacking highly detailed textures.
 }
  \label{fig:multiview_garden_toys}
\vspace{-0mm}
\end{figure*}

\begin{figure*}[tb]
  \centering
  \includegraphics[height=18.7cm]{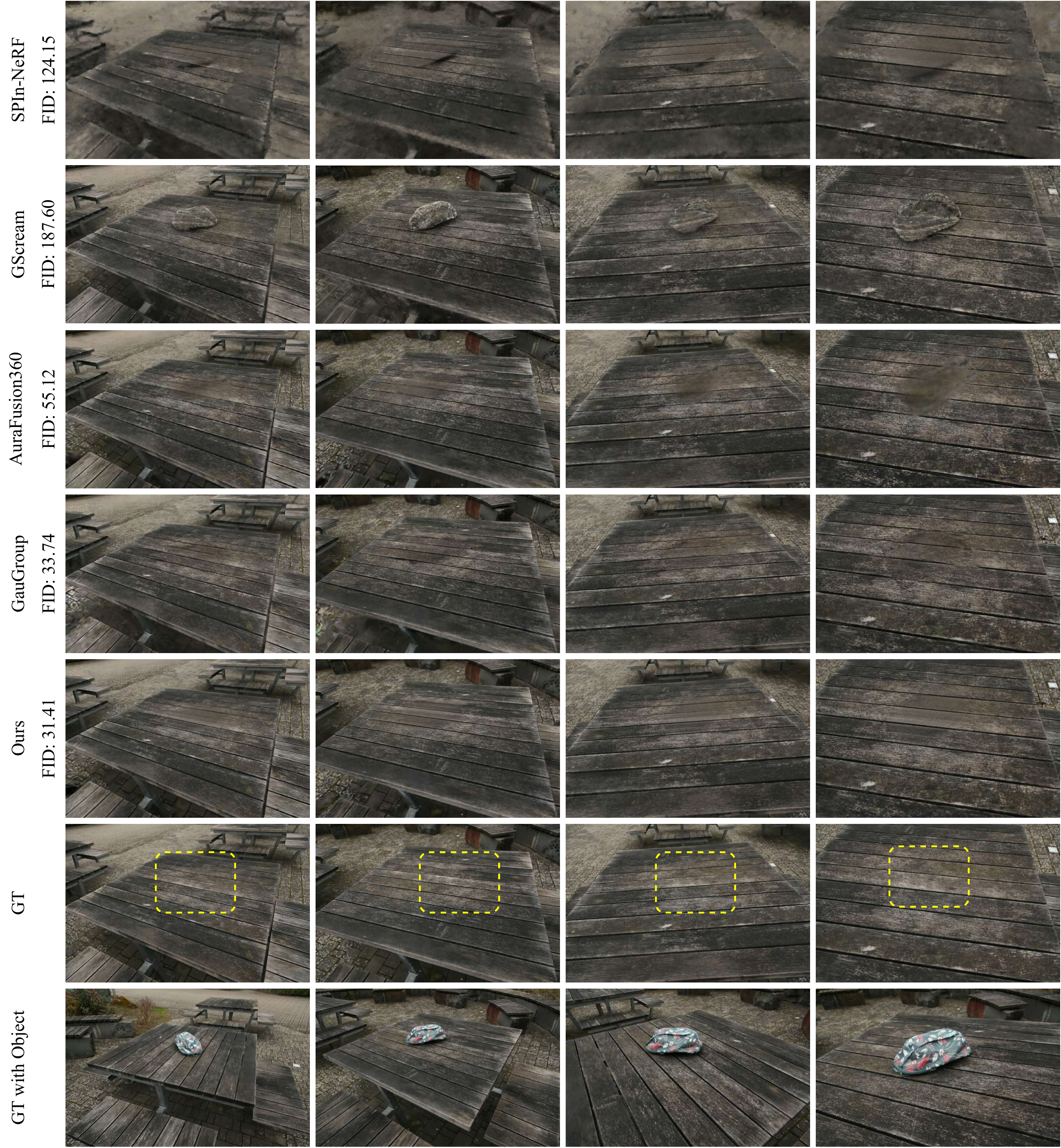}
  \caption{\textbf{Multi-view comparison on Inpaint360GS}~\texttt{bag}. We evaluate SPIn-NeRF~\cite{spinnerf}, GScream~\cite{gscream}, AuraFusion360~\cite{aurafusion360}, GauGroup~\cite{gaussiangrouping} and our method, with object-inclusive ground truth images provided for each corresponding view.  Our method achieves the best FID score, produces noticeably smoother edges, and is approximately 5 $\times$ faster than the 3D inpainting stage of the second-best method GauGroup~\cite{gaussiangrouping}.
 }
  \label{fig:multiview_bag}
\vspace{-0mm}
\end{figure*}

\begin{figure*}[tb]
  \centering
  \includegraphics[height=18.7cm]{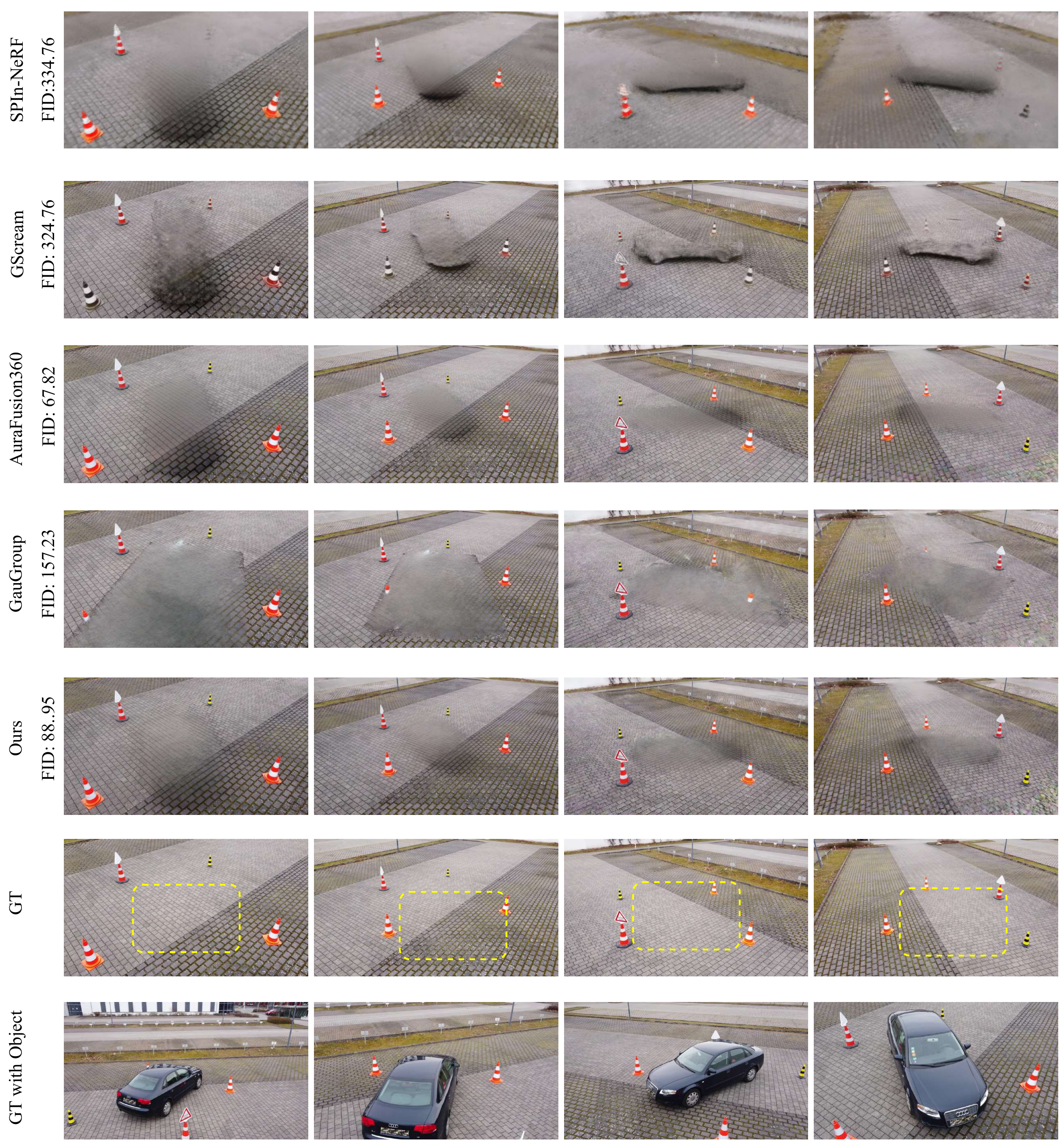}
  \caption{\textbf{Multi-view comparison on Inpaint360GS}~\texttt{car}. We evaluate SPIn-NeRF~\cite{spinnerf}, GScream~\cite{gscream}, AuraFusion360~\cite{aurafusion360}, GauGroup~\cite{gaussiangrouping} and our method, with object-inclusive ground truth images provided for each corresponding view.This scene is particularly challenging due to the complex and texture-less ground surface, which makes it difficult to infer plausible textures. AuraFusion360 achieves strong FID performance due to its single-view guidance combined with extensive post-refinement. However, its optimization time is approximately 20$\times$ longer than ours. In contrast, our method achieves competitive results with a significantly more efficient pipeline.
 }
  \label{fig:multiview_car}
\vspace{-0mm}
\end{figure*}

\begin{figure*}[tb]
  \centering
  \includegraphics[height=18.7cm]{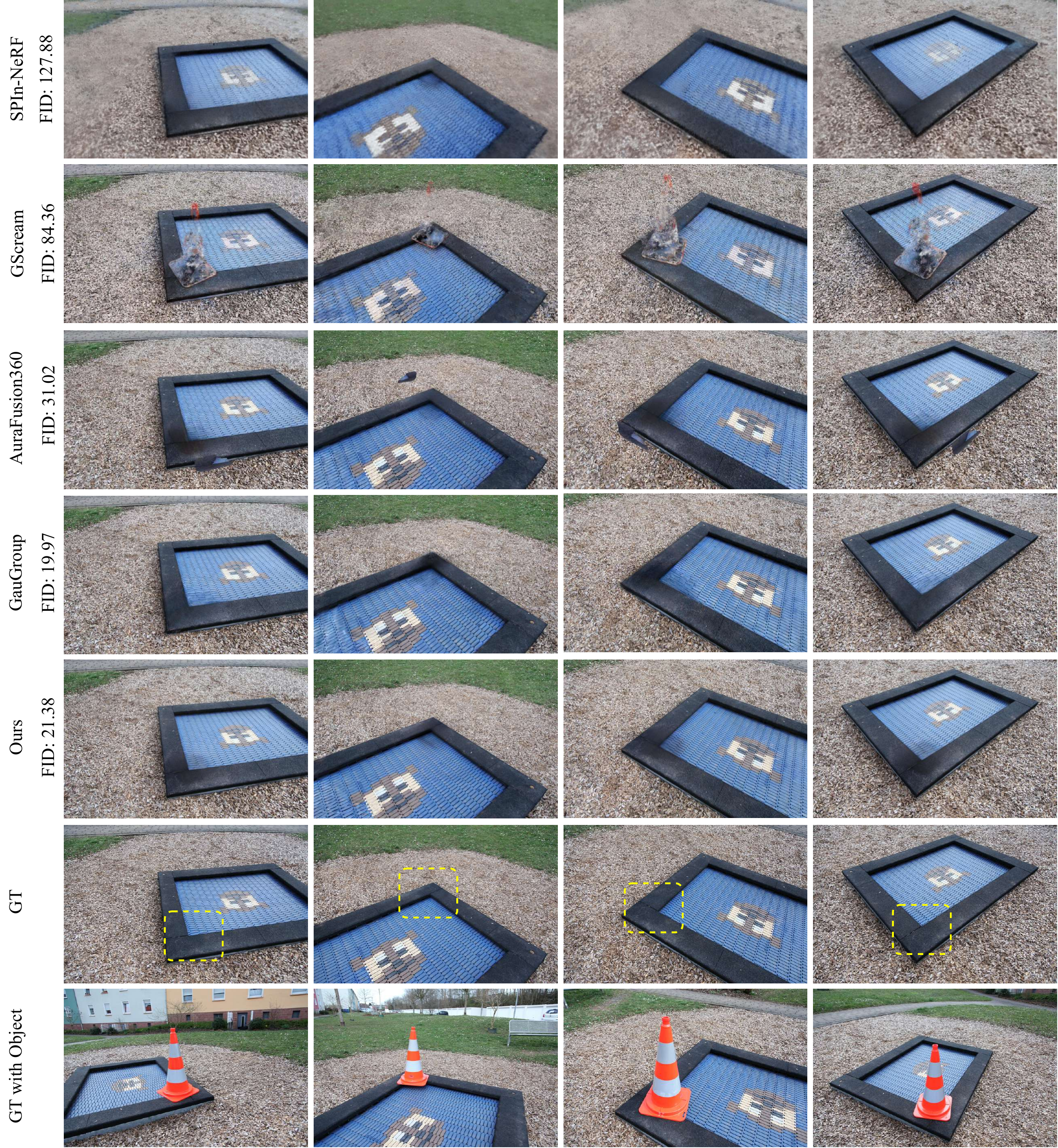}
  \caption{\textbf{Multi-view comparison on Inpaint360GS}~\texttt{red cone}. We evaluate SPIn-NeRF~\cite{spinnerf}, GScream~\cite{gscream}, AuraFusion360~\cite{aurafusion360}, GauGroup~\cite{gaussiangrouping} and our method, with object-inclusive ground truth images provided for each corresponding view. This scene presents a challenging case due to significant depth variations and complex textures, making accurate inpainting difficult. GauGroup achieves the best visual quality, while our method performs comparably, producing plausible results with effective depth reasoning.
 }
  \label{fig:multiview_cone_red}
\vspace{-0mm}
\end{figure*}

\begin{figure*}[tb]
  \centering
  \includegraphics[height=18.7cm]{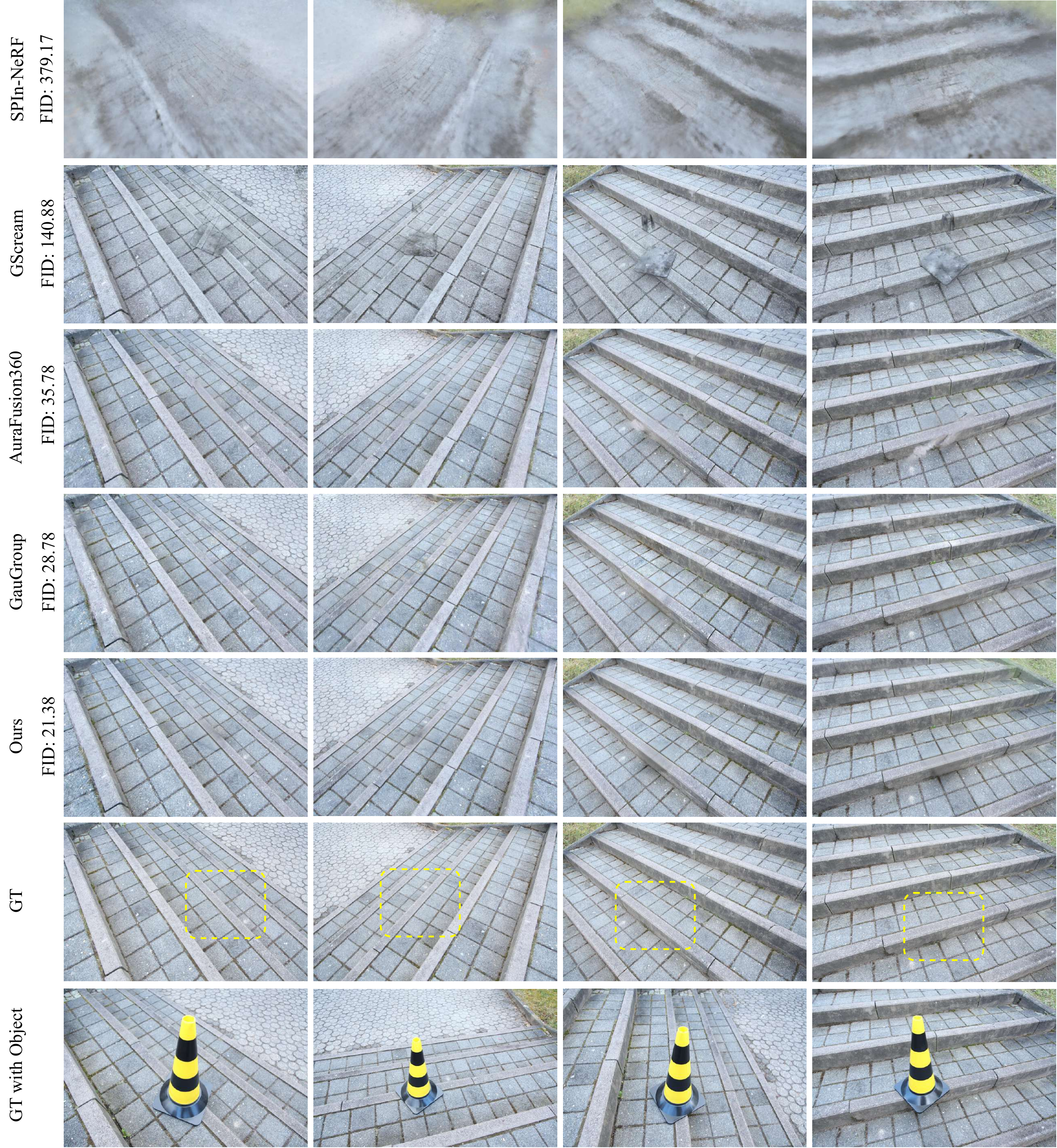}
  \caption{\textbf{Multi-view comparison on Inpaint360GS}~\texttt{yellow cone}. We evaluate SPIn-NeRF~\cite{spinnerf}, GScream~\cite{gscream}, AuraFusion360~\cite{aurafusion360}, GauGroup~\cite{gaussiangrouping} and our method, with object-inclusive ground truth images provided for each corresponding view. This scene includes a staircase, posing a challenge for depth estimation. Our method converges efficiently and maintains strong performance. Notably, GauGroup~\cite{gaussiangrouping} achieves the second-best results but requires 5$\times$ longer optimization time.}
  \label{fig:multiview_cone_yellow}
\vspace{-0mm}
\end{figure*}

\begin{figure*}[tb]
  \centering
  \includegraphics[height=18.7cm]{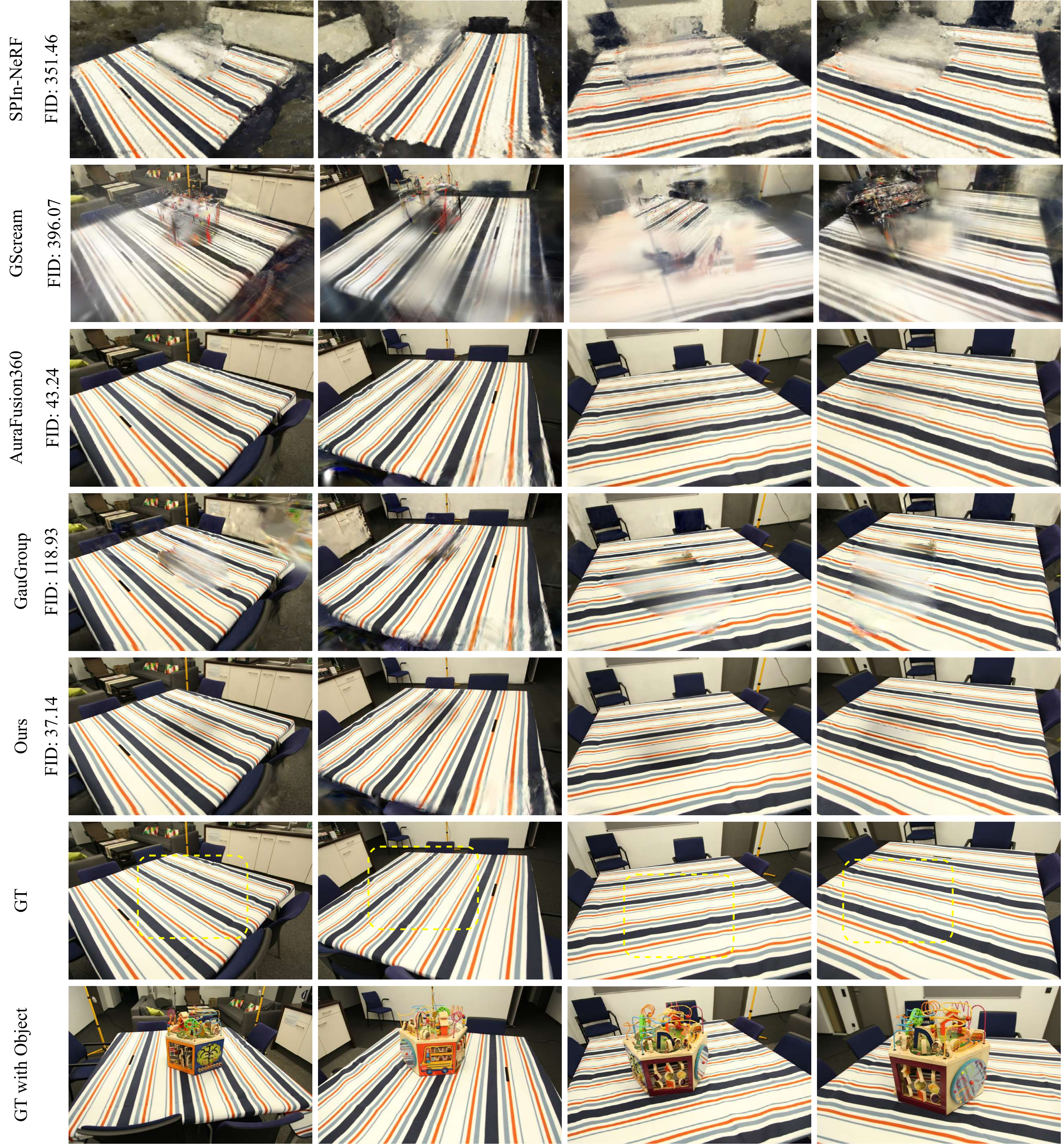}
  \caption{\textbf{Multi-view comparison on Inpaint360GS}~\texttt{cube}. We evaluate SPIn-NeRF~\cite{spinnerf}, GScream~\cite{gscream}, AuraFusion360~\cite{aurafusion360}, GauGroup~\cite{gaussiangrouping} and our method, with object-inclusive ground truth images provided for each corresponding view. In this scene, GScream~\cite{gscream} encounters significant issues due to inconsistencies between the depth provided by Marigold~\cite{marigold} and the depth scale of the COLMAP-initialized point cloud. The failure of depth alignment leads to degraded performance. While AuraFusion360 demonstrates competitive performance, it exhibits noticeable boundary ambiguity in the inpainted regions. In contrast, our method avoids this problem by directly defining depth using intrinsic properties of the Gaussian scene, thereby eliminating the need for external depth alignment. As a result, our pipeline achieves the best performance. }
  \label{fig:multiview_cube}
\vspace{-0mm}
\end{figure*}

\begin{figure*}[tb]
  \centering
  \includegraphics[height=18.7cm]{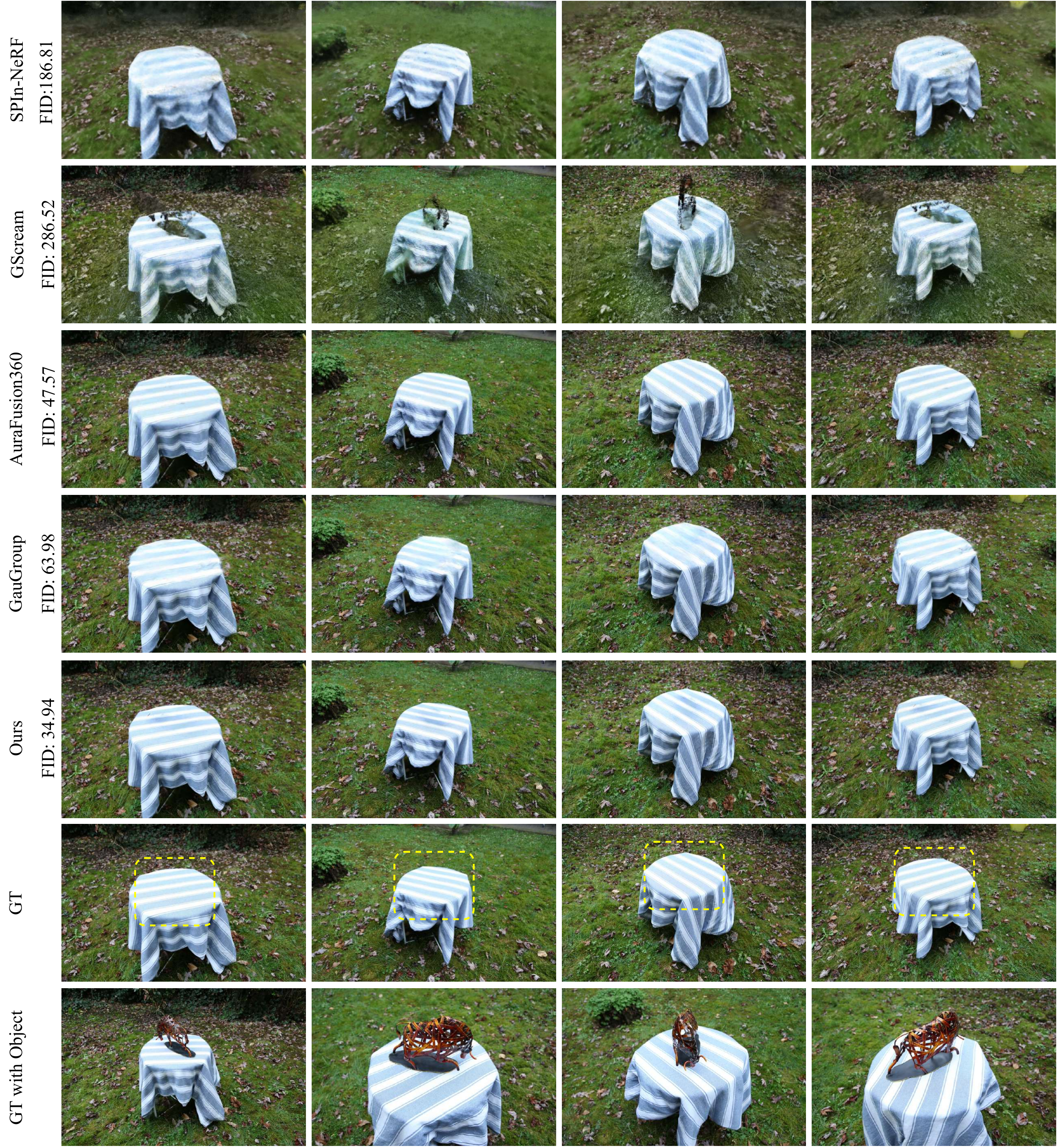}
  \caption{\textbf{Multi-view comparison on Inpaint360GS}~\texttt{redbull}. We evaluate SPIn-NeRF~\cite{spinnerf}, GScream~\cite{gscream}, AuraFusion360~\cite{aurafusion360}, GauGroup~\cite{gaussiangrouping} and our method, with object-inclusive ground truth images provided for each corresponding view. Although this is a single-object scene, the bull model contains fine-grained structures such as horns and a tail, posing challenges for accurate 3D Gaussian identity assignment. All methods except GScream produce visually reasonable results under this setting. Please zoom in for details.
 }
  \label{fig:multiview_redbull}
\vspace{-0mm}
\end{figure*}

\begin{figure*}[tb]
  \centering
  \includegraphics[height=18.7cm]{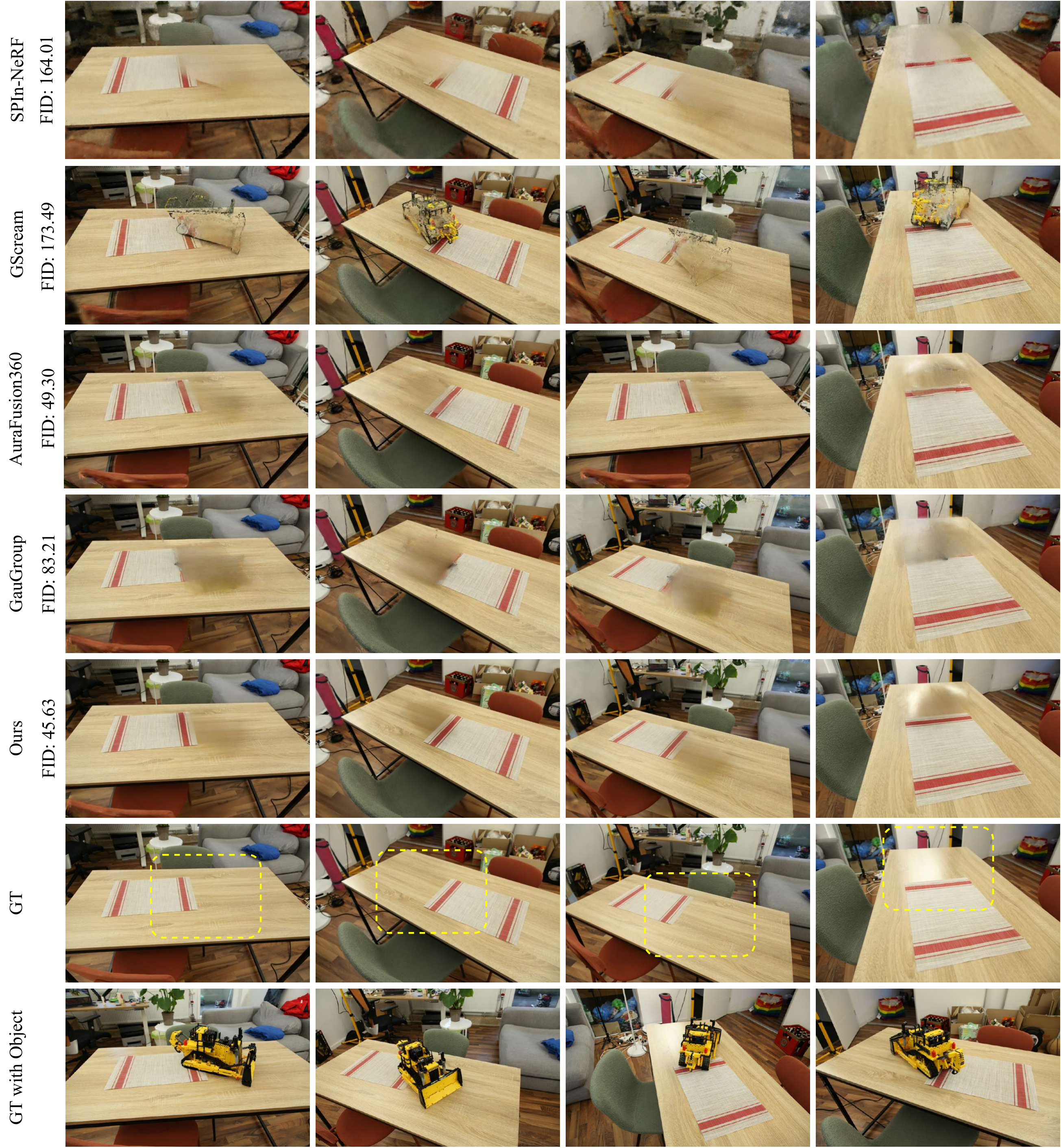}
  \caption{\textbf{Multi-view comparison on Inpaint360GS}~\texttt{truck}. We evaluate SPIn-NeRF~\cite{spinnerf}, GScream~\cite{gscream}, AuraFusion360~\cite{aurafusion360}, GauGroup~\cite{gaussiangrouping} and our method, with object-inclusive ground truth images provided for each corresponding view. Our method achieves the best FID score and is 20 $\times$ faster than AuraFusion~\cite{aurafusion360}, while requiring no additional parameter tuning. However, none of the evaluated methods, including ours, are yet capable of effectively handling complex lighting and shadow effects present in the scene, which remains an open challenge for future research.}
  \label{fig:multiview_truck}
\vspace{-0mm}
\end{figure*}

\begin{figure*}[tb]
  \centering
  \includegraphics[height=19cm]{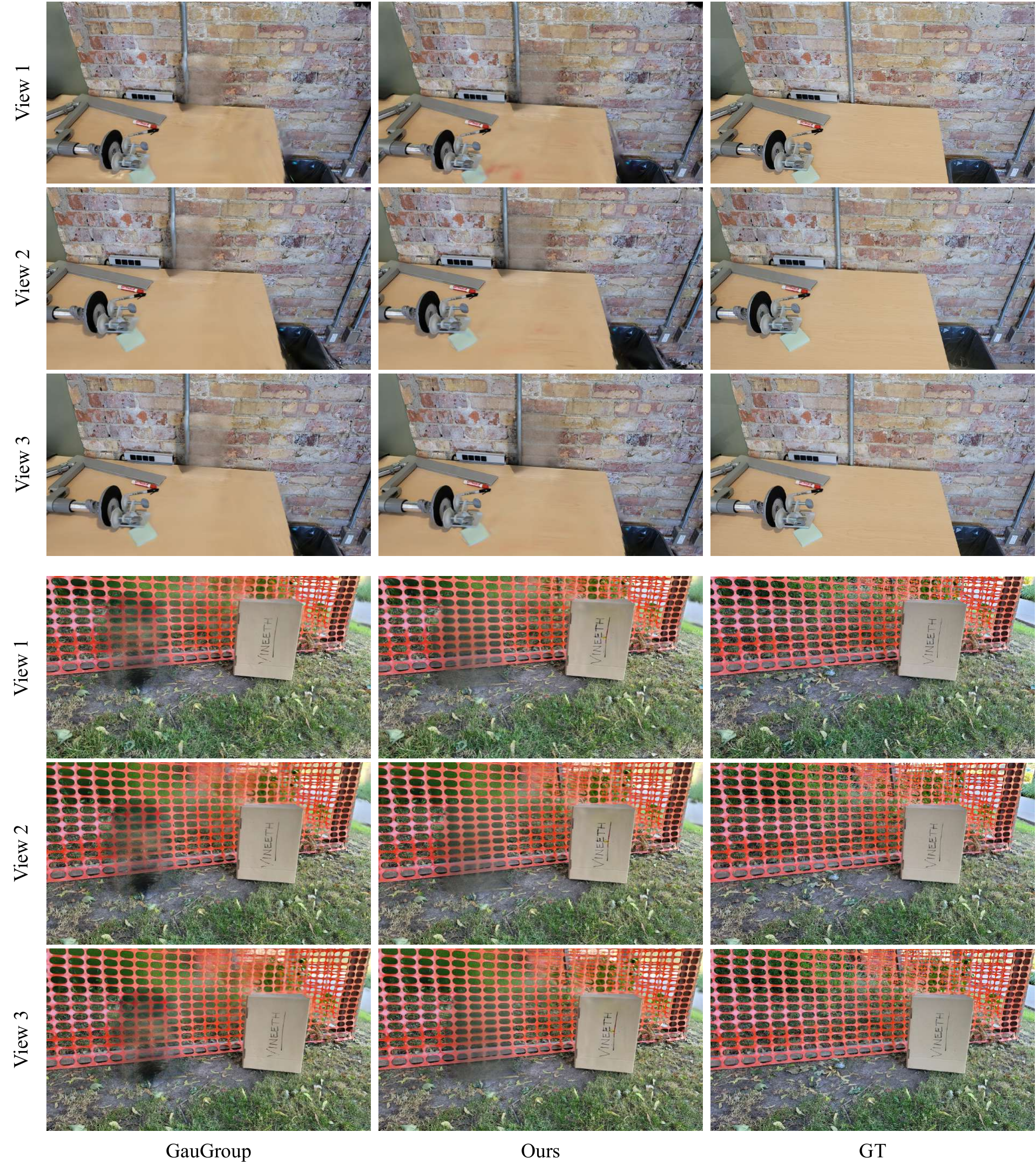}
  \caption{\textbf{Performance on SPIn-NeRF~\cite{spinnerf} Dataset.} We evaluate GauGroup~\cite{gaussiangrouping} and our method on front facing SPIn-NeRF~\cite{spinnerf} dataset. Our method remains robust on this dataset and consistently outperforms GauGroup, achieving a 0.6 dB improvement in PSNR and a notable 5 points gain in FID.
 }
  \label{fig:multiview_spinnerf}
\vspace{-0mm}
\end{figure*}

%%%%%%%%%%%%%%%%%%%%%%%%%%%%%%%%%%%%%%%%%%%%%%%%%%%%%%%%%%%%%%%%%%%%

%                     following for separate compiling              % 

%%%%%%%%%%%%%%%%%%%%%%%%%%%%%%%%%%%%%%%%%%%%%%%%%%%%%%%%%%%%%%%%%%%%

% %%%%%%%%% REFERENCES
% \clearpage

% \ifdefined\isMain
% \else
% \clearpage
% \small
% \bibliographystyle{ieeenat_fullname}
% \bibliography{main}

% \end{document}

\end{document}